\documentclass{article}

    \usepackage[preprint]{neurips_2024}

\usepackage[utf8]{inputenc} % allow utf-8 input
\usepackage[T1]{fontenc}    % use 8-bit T1 fonts
\usepackage{hyperref}       % hyperlinks
\usepackage{url}            % simple URL typesetting
\usepackage{booktabs}       % professional-quality tables
\usepackage{amsfonts}       % blackboard math symbols
\usepackage{nicefrac}       % compact symbols for 1/2, etc.
\usepackage{microtype}      % microtypography
\usepackage{xcolor}         % colors
\usepackage{amsmath}
\usepackage{graphicx}
\usepackage{subcaption}
\usepackage{multirow}
\usepackage{wrapfig}
\usepackage{alphalph}
\title{Exploring Activation Patterns of Parameters \\ in Language Models}
\author{%
  Yudong Wang, Damai Dai, Zhifang Sui\\
  MOE Key Lab of Computational Linguistics, School of Computer Science, Peking University\\
  \texttt{yudongwang@stu.pku.edu.cn},  \texttt{\{daidamai, szf\}@pku.edu.cn} \\
}

\begin{document}

\maketitle

\begin{abstract}

  Most work treats large language models as black boxes without in-depth understanding of their internal working mechanism.
  In order to explain the internal representations of LLMs, we propose a gradient-based metric to assess the activation level of model parameters.
  Based on this metric, we obtain three preliminary findings.
  (1) When the inputs are in the same domain, parameters in the shallow layers will be activated densely, which means a larger portion of parameters will have great impacts on the outputs. 
  In contrast, parameters in the deep layers are activated sparsely. 
  (2) When the inputs are across different domains, parameters in shallow layers exhibit higher similarity in the activation behavior than deep layers. 
  (3) In deep layers, the similarity of the distributions of activated parameters is positively correlated to the empirical data relevance. 
  Further, we develop three validation experiments to solidify these findings. 
  (1) Firstly, starting from the first finding, we attempt to configure different prune ratios for different layers, and find this method can benefit model pruning. 
  (2) Secondly, we find that a pruned model based on one calibration set can better handle tasks related to the calibration task than those not related, which validate the second finding.
  (3) Thirdly, Based on the STS-B and SICK benchmark, we find that two sentences with consistent semantics tend to share similar parameter activation patterns in deep layers, which aligns with our third finding. 
  Our work sheds light on the behavior of parameter activation in LLMs, and we hope these findings will have the potential to inspire more practical applications.
\end{abstract}

% 随着模型规模、数据规模的增大，模型得到了一定的通用能力。但是针对通用能力内部机理的研究很少，大部分工作还是将LLM作为一种黑盒模型。本工作提出了一种能衡量模型内部参数激活程度的评分机制，从而探究LLMs运行时的内部机理。本工作有四个主要发现。1） 针对不同数据源，模型内部参数被激活的程度不同，针对相同数据源，模型内部参数被激活的程度一致。2） 对相同数据源，如果模型内部参数被激活的程度越大，模型在相应数据源表现越差，数据源越丰富。3） 按经验主义，越相似的数据集之间，模型被激活的参数分布越一致，越不相似的数据集之间，模型被激活的参数分布越不同。4） 模型前层网络被激活分布在不同数据集之间表现较为一致，中间层分布更为不同，这可能反映了前端端网络是通用能力，后端网络则是具体任务的能力。为了验证这三个性质，我们设置了两个实验。通过对不同层剪枝比例的调整，我们提高了模型在单任务剪枝上的性能。通过对不同数据源距离的测量，我们设置的数据混合比例让模型预训练效果更好。通过

%%%%
%这是abstract，请将以下内容翻译为学术英语，避免使用偏僻词。
%随着模型规模、数据规模的增大，模型得到了一定的通用能力。但是针对通用能力内部机理的研究不够充分，大部分工作还是将LLM作为一种黑盒模型。本工作提出了一种能衡量模型内部参数激活程度的评分机制，从而探究LLMs运行时的内部机理。本工作有3个主要发现。1） 针对同源的输入，模型内部靠前的层中更多参数被激活，对结果有更大影响，而靠后的层中只有很少的参数对结果的影响较大。2） 对于非同源的输入，模型内部靠前层的参数被激活分布的相似度较高，靠后层的参数被激活分布的相似度较低。3）模型后层的激活程度的相似度与经验主义的数据相关性之间有显著的相关性。为了验证这三个性质，我们设置了两个模型剪枝实验，一个语意相似度实验。在剪枝实验中，我们通过调整不同层的剪枝比例提高了剪枝性能，这验证了第一个性质。而通过使用不同的calibration set作剪枝，剪枝后的模型在与calibration相关性强的测试上结果提升，而在其他方面结果下降，这验证了第二个性质。最后，我们验证了模型后层的激活程度的相似度在语意相似度benchmark STSB上的结果，验证了第三个性质。我们希望我们的工作能推动对LLM内部机理的理解，从而帮助LLM的发展。

\section{Introduction}\label{sec:intro}

% Since ChatGPT~\ref{}, Large Language Models (LLMs) have received a lot of attention. While the capabilities of LLMs become stronger, the explanation is still a lack of research. 

% In this work, we care about the following question: faced with different inputs, which parameters inside the LLMs decide the final result? In prior work, there are different opinions. On the one hand, researchers focused on network pruning believe that most of the parameters inside the network are useless. They try to analyze the weights in the network. With a given score, they prune those parameters they think lack of use to the final result. On the other side, those who focus on Mixture of Experts (MoE) trust that the scale of the parameters is the original reason enable LLMs to gain general capabilities, and network pruning will lead to the decrease of networks. They come up with sparse models with large amounts of parameters to explore the boundary of the LLMs. 

% In this study, we investigate a critical question that has been largely overlooked: What governs the universal capabilities of models? More specifically, when faced with diverse inputs, what weights within the LLM dictate the final output? The academic community is divided on this issue. Some researchers, particularly those who focus on network pruning, argue that a significant portion of the internal weights of the network are superfluous. Their approach is to analyze the network's weights and, based on a pre-established score, prune weights deemed irrelevant to the outcome.
Ever since the emergence of GPT-4~\cite{achiam2023gpt}, there has been a surge of interest in Large Language Models (LLMs). As these LLMs continue to advance and their capabilities strengthen, there remains a noticeable gap in research dedicated to their interpretability. 
%In this study, we attempt to investigate the following question: How do different capabilities coexist within the model? More specifically, when faced with inputs that require different capabilities, what are the variations in the internal representation of LLMs? 
In this study, we aim to investigate the coexistence of different capabilities within the model. More specifically, when faced with inputs that are across different domains, we observe variations in the internal representation of Large Language Models (LLMs).
There have been some explorations into the functions of specific layers and parameters in LLMs~\cite{azaria2023internal, geiger2024finding}. It is generally recognized that some significantly different capabilities in LLMs cannot fully coexist within a limited scale. However, there is still no targeted research that analyzes the operational patterns of different capabilities within large models in a more general sense.

%On the other hand, proponents of the Mixture of Experts (MoE) model maintain that the sheer scale of weights is the fundamental enabler of LLMs' general capabilities. They argue that network pruning inevitably leads to a reduction in network size, which in turn limits the model's capabilities. To counteract this, they propose sparse models with a large number of weights as a means to explore the limits of LLMs.

% \ref{} anaylzed the model capacities in different tasks after knowledge distillation. Although the distilled model performs well in the specific task, the original general model lost its capacities in other tasks it used to perform well. The phenomenon shows that limited to a scaling, the sum of capacities of a model is limited. \ref{} come up with the idea that LLMs will become an MoE within themselves, which showed that different parts of the network operate for different inputs. 

% Fu et al.~\cite{fu2023specializing} analyzed model capacities across various tasks following knowledge distillation. 
% Recent work analyzed model capacities across various tasks following knowledge distillation~\cite{fu2023specializing}. 
% The findings revealed that while the distilled model excels in the specific task it was designed for, the original, more general model experiences a decline in performance in other tasks it was previously proficient. 
Recent work has found that there may be some parameters within the model that exist for specific tasks.
Fu et al.~\cite{fu2023specializing} revealed that while the distilled model excels in the specific task it was designed for, the original, more general model experiences a decline in performance in other tasks it was previously proficient. 
This observation suggests that different tasks may tap into distinct capacities within a model, and these capacities seem to be mutually exclusive to some extent. 
In another study, Zhang et al.~\cite{zhang2021moefication} introduced the notion that LLMs inherently evolve into a Mixture of Experts (MoE) within themselves. 
This concept implies that different sections of the network are tasked with handling different inputs, further strengthening the idea of internal specialization within the model.

% Motivated by the two phonemenons above, this work attempts to answer the questions: which parameters decide the outputs of the network? (in the other word "activated") Is there any rules of the distribution of acivated parameters face with different inputs? 

Building on the insights gleaned from the aforementioned phenomena, this study seeks to unravel the following questions: 
%which parameters within the network are activated to determine the outputs? Does the distribution of these activated weights exhibit distinct patterns when faced with non-homogenous inputs? 
Which parameters within the network are activated to determine the outputs and does the distribution of these activated weights exhibit distinct patterns when faced with inputs across different domains? 
In essence, we aim to explore whether the degree of parameter activation varies in response to non-homogenous input scenarios, and if so, to what extent.

% According to the methods in network pruning~\ref{}, we evaluate the impact of a parameter with the difference between the original output of a model and a model with the parameter set to $0$. Specifically, we use the first-order term of the Taylor expansion of the model to measure the impact of a parameter on the outputs. Given two inputs $x, y$, we can gain two vectors $v_x, v_y$, which discribe the impact of the parameters inside the model. Observing the cosine similarity between these two vectors, we gain the following 4 phenomena:

% \begin{enumerate}

% \item For different data sources, the distributions of the front layers of the Llama are of high 
% cosine similarity, while the latter layers are of low. 

% \item For a same data source, the distributions of the Llama with different samples are of high cosine similarity.

% \item For different data sources, the distributions of the llama are of high cosine similarity with datasets which we think empirical of high similarity.

% \item For a same data source, the distributions of the llama are of low cosine similarity with harder datasets empirically. 

% \end{enumerate}

Drawing on methods from network pruning~\cite{ma2023llm}, we assess the influence of a parameter by comparing the original output of a model to that of a model in which the parameter is set to $0$. Specifically, we employ the first-order term of the Taylor expansion of the model to gauge the impact of a parameter on the outputs. Given two inputs $x, y$, whether homogenous or not, we derive two vectors $v_x, v_y$ that characterize the influence of the internal parameters of the model. By examining the \textbf{cos}ine similarity between these two vectors from \textbf{LLM} with different \textbf{d}ata (LLMDcos), we observe three phenomena:

\begin{itemize}

\item For inputs in the same domain, parameters in the shallow layers of the model are activated densely, while parameters in the deep layers are activated sparsely. 

\item For inputs from different domains, the similarity of the activation patterns of parameters in the shallow layers of the model is higher than deep layers. 

\item In deep layers, the similarity of the distributions of activated parameters is positively correlated to the empirical data relevance.

\end{itemize}

To validate our observed results, we designed three experiments, which include model pruning and semantic similarity tasks. The pruning method improved based on our analytical results outperformed the original pruning method. We validated our second finding by comparing the performance changes caused by the different calibration sets of the pruning method. The proposed LLMDcos was also validated to be related to semantic similarity. 

Our contributions are listed in the following:

\begin{itemize}

% \item When dealing with different data sources, the distributions of the initial layers of the LLMs exhibit high cosine similarity, whereas the later layers show low similarity.

% \item For the same data source, the distributions of activations of the LLMs across different samples demonstrate high cosine similarity.

% \item For different data sources, the distributions of activations of the LLMs exhibit high cosine similarity with datasets that are empirically deemed to be highly similar.

% \item For a single data source, the distributions of the LLMs show low cosine similarity with datasets that are empirically considered to be more challenging.

% \item For homogenous data, more parameters in the front layers of the model are activated and have a greater impact on the results, while fewer parameters in the latter layers have a significant impact on the results. 

% \item For non-homogenous data, the similarity of the activation distribution of parameters in the front layers of the model is higher, while the similarity of the activation distribution of parameters in the latter layers is lower. This may reflect that the parameters in the front layers have more general capabilities, while the parameters in the latter layers are discretely contributing to different capabilities. 

% \item The degree of activation in the latter layers of the model has a strong connection with semantic similarity, which may reflect the relationship between different capabilities and different sentences.

\item We employ a novel approach to analyze the internal capabilities of the model.

\item We observed the different capabilities of different layers in LLMs, summarized three phenomena, and designed experiments to validate each respectively.

\item We optimized the pruning method, providing a reference for other pruning methods.

\item We proposed a new method for calculating data similarity based on gradient information.

\end{itemize}

% We believe that this work contributes to a deeper understanding of the internal mechanisms of large-language models (LLMs). By shedding light on the distribution of activated weights within the network and introducing the LLMDcos metric, we provide novel insights into the functioning of these complex models. Furthermore, our findings offer valuable guidance for future research on model pruning and data similarity. Our study, therefore, represents a significant stride towards demystifying the inner workings of LLMs, ultimately paving the way for advancements in the field of explainable AI.

\section{Background and Motivation}

\subsection{Motivation}

Our motivation for this study stems from a phenomenon observed in distillation~\cite{fu2023specializing}: when one capability of a general model is enhanced through distillation, there tends to be a corresponding decline in other evaluations. This observation prompts us to investigate the nature of the relationship between different capabilities within a model - are they mutually reinforcing or mutually exclusive? And if both, under what circumstances does each scenario occur?

A study by ~\cite{zhang2021moefication} proposed the idea that a model internally generates a Mixture of Experts (MoE), which suggests that the model handling of different tasks could be attributed to the spontaneous formation of a sparse structure during training. This structure, in turn, might harbor distinct capabilities that are mutually exclusive to some extent.

Our work is primarily related to two areas: model pruning and data similarity. The parameter scoring method from model pruning serves as a valuable tool to explore which parameters within the model are most responsive to a given input. Meanwhile, the variation in the model's performance during evaluation due to different pruning settings can also validate our conclusions. On the other hand, we have utilized non-homologous data to investigate whether there would be different activation distributions within the model. From the results, the different activation distributions are related to data relevance. Through this lens, we aim to shed light on the internal dynamics of large language models and their response to varying inputs.

% \subsection{Causal Inference}
\subsection{Causal Inference}

%In the endeavor to construct reliable and interpretable AI, numerous techniques have been pruposed to scrutinize the causal reasoning process within Large Language Models (LLMs). 
In line with our intention, many techniques in causal inference are also aimed at exploring the mechanisms and patterns within the network.
These techniques include probing, attribution methods, and causal abstraction.

Probes are essentially models that are trained with the internal representations of a neural network as input, aiming to explore the inherent semantics within the model~\cite{hupkes2018visualisation, peters2018dissecting, tenney2019bert, clark2019does}. A plethora of studies employing probes have delved into internal information related to aspects such as time, space, and inferential variables. 
%However, the utility of probes is constrained by the training process, as they are typically aligned with pre-existing knowledge. Thus, their application is limited to determining whether a specific layer contains information that can be extracted from the input.

Attribution methods~\cite{shrikumar2016not, sundararajan2017axiomatic}, in line with our objectives, strive to quantify the degree to which a representation contributes to the output of the model for a specific sample or set of samples. Using gradient information, attribution methods inherently offer explanations, thereby demystifying the inner mechanics of the neural network. 
%However, in this study, our focus is on constant weights, as opposed to neural variations with different inputs. Furthermore, in contrast to previous work on Computer Vision and original networks, our emphasis is on LLMs.

Causal abstraction~\cite{geiger2021causal, geiger2024finding}, on the other hand, concentrates more on the specific implications within the network. It evaluates the effect of a weight or neuron by fixing, disturbing, or setting it to zero and observing the difference in the outputs. Like probes, causal abstraction requires extensive prior knowledge, which limits its ability to examine more general situations and information.

However, the aforementioned techniques, probe and causal abstraction, are aimed at verifying whether specific information from human reasoning processes exists within the network. Meanwhile, attribution methods concentrate on specific data within the dataset. In contrast, our work primarily focuses on explaining the internal mechanisms of the model by studying the differences in the model's internal state when facing inputs from different domains.

\subsection{Model Pruning}

The crux of model pruning lies in identifying the crucial parameters within the network. From the perspective of model pruning, we can derive insights into the significant role scoring of parameters.

Model pruning techniques~\cite{lecun1989optimal, hassibi1993optimal, han2015learning} for LLMs can be broadly categorized into two types~\cite{zhu2023survey}: structured pruning~\cite{frantar2023sparsegpt, zhang2023pruning, sun2023simple} and unstructured pruning~\cite{santacroce2023matters, ma2023llm}. Structured pruning aims to reduce the hidden state size by removing entire rows or columns from the weight matrix, which can lead to actual acceleration and pruning benefits. However, this method often results in a significant loss of performance. Unstructured pruning, on the other hand, involves eliminating individual connections, i.e., specific elements within the weight matrix. This approach can maintain model performance even at high pruning ratios but does not inherently lead to computational speedup unless a substantial proportion of connections is pruned within specific regions.

Regardless of the type of pruning, both methods focus on identifying which components of the network have the least impact on the output. Many studies have utilized Taylor expansion to define the rank of weights in terms of their influence on the network's structure, thereby guiding the pruning process by removing weights with minimal impact. 

In this work, we draw upon the Taylor expansion to define the degree to which internal weights are activated, thereby investigating the underlying mechanisms within the model.

% \section{Analysis}
\section{Preliminary Findings}

% \subsection{Preliminary}

In all the content of this paper, unless specifically marked, all model results are analysis results of Llama2-7b-hf~\cite{touvron2023llama}. This paper provides results from more models in subsequent sections.

\subsection{Definition of Activation and LLMDcos}\label{sec:define}

We begin with a standard deep learning problem in an empirical scenario. Given two data sources $X, Y$, our aim is to quantify the influence of the parameters $w$ within the model $D$. The activation of $w_i$ is defined as

% \begin{center}
    
\begin{equation}
    % \begin{equation}
     \mathcal{A}(X, w_i) = |D(X, w_i) - D(X, 0)|
    =|w_i\cdot\frac{\partial D(X, w_i)}{\partial w_i} + O(w_i^2)|
    \approx |w_i\cdot\frac{\partial D(X, w_i)}{\partial w_i}|
% \end{equation}
\end{equation}
% \end{center}

% Concat all the $\mathcal{A}(X, w_i)$ together we get $\mathcal{A}(X, w)\in R^{n}$, where $n$ refer to the number of parameters inside the model. We investigate the $\mathcal{A}(w)$ through out different sentence from different data source. We define the Language Model-based cosine metric of pair

By concatenating all the $\mathcal{A}(w_i)$, we derive $\mathcal{A}(w)\in R^{n}$, where $n$ denotes the number of parameters within the model. We examine $\mathcal{A}(w)$ across different sentences from various data sources. We define the \textbf{c}osine metric of a \textbf{D}ata pair based on the \textbf{L}arge \textbf{L}anguage \textbf{M}odel (LLMDcos):

\begin{equation}
    LLMDcos(X_1, X_2) = \frac{A(X_1, w)\cdot A(X_2, w)}{\sqrt{\|A(X_1, w)\|^2\cdot \|A(X_2, w)\|^2}}
\end{equation}

Where $X_1, X_2$ are two inputs from the same domain or different domains. With LLMDcos, our aim is to analyze different layers within LLMs, as well as the differences between different inputs.

\begin{figure}[t]
  \begin{subfigure}{0.15\textwidth}
    \includegraphics[width=\textwidth]{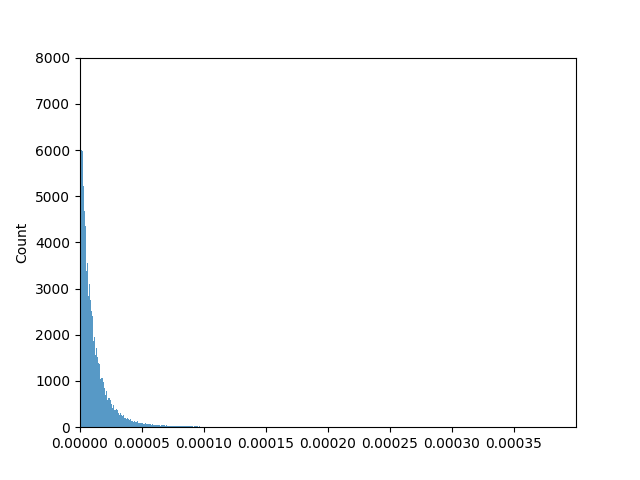}
    % \caption{layer 1}
  \end{subfigure}
  \hfill
  \begin{subfigure}{0.15\textwidth}
    \includegraphics[width=\textwidth]{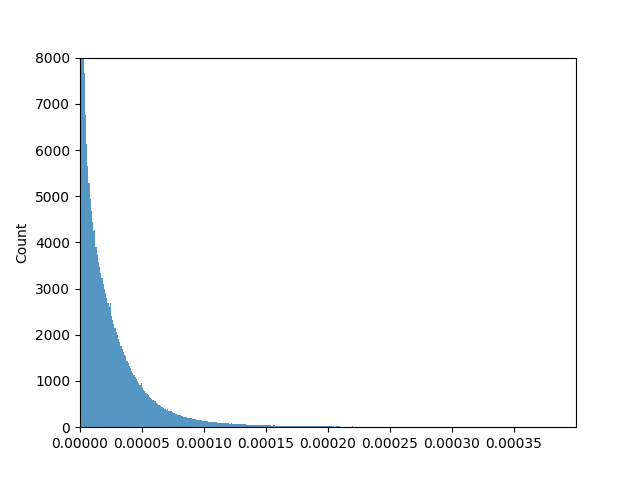}
    % \caption{layer 2}
  \end{subfigure}
  \hfill
  \begin{subfigure}{0.15\textwidth}
    \includegraphics[width=\textwidth]{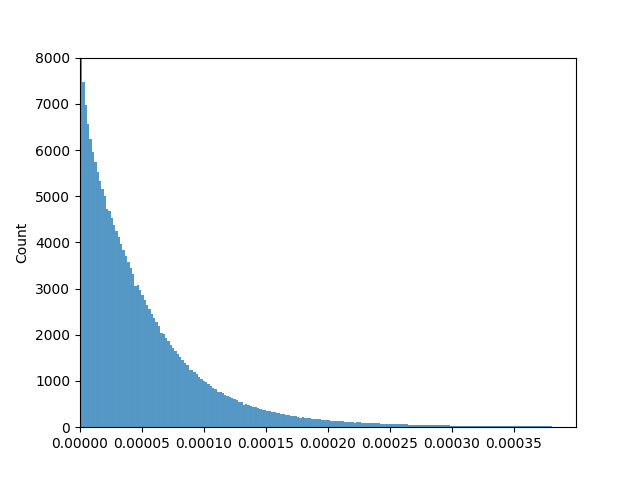}
    % \caption{layer 11}
  \end{subfigure}
  \hfill
  \begin{subfigure}{0.15\textwidth}
    \includegraphics[width=\textwidth]{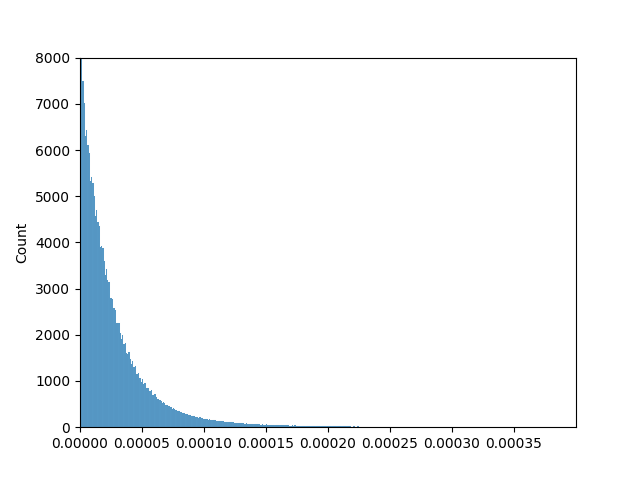}
    % \caption{layer 16}
  \end{subfigure}
  \hfill
  \begin{subfigure}{0.15\textwidth}
    \includegraphics[width=\textwidth]{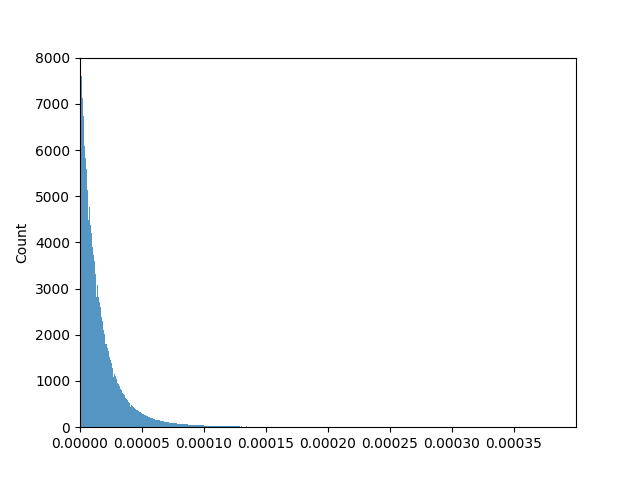}
    % \caption{layer 24}
  \end{subfigure}
  \hfill
  \begin{subfigure}{0.15\textwidth}
    \includegraphics[width=\textwidth]{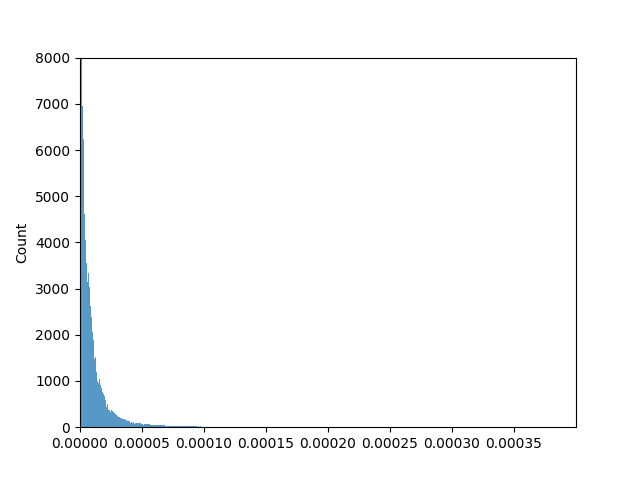}
    % \caption{layer 31}
  \end{subfigure}
  \hfill

  \begin{subfigure}{0.15\textwidth}
    \includegraphics[width=\textwidth]{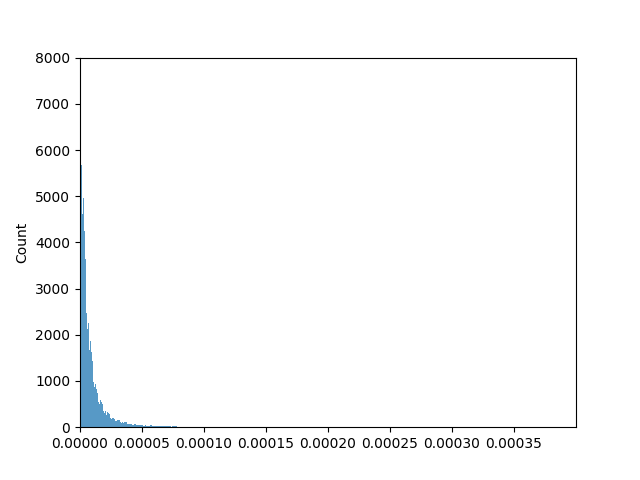}
    % \caption{layer 1}
  \end{subfigure}
  \hfill
  \begin{subfigure}{0.15\textwidth}
    \includegraphics[width=\textwidth]{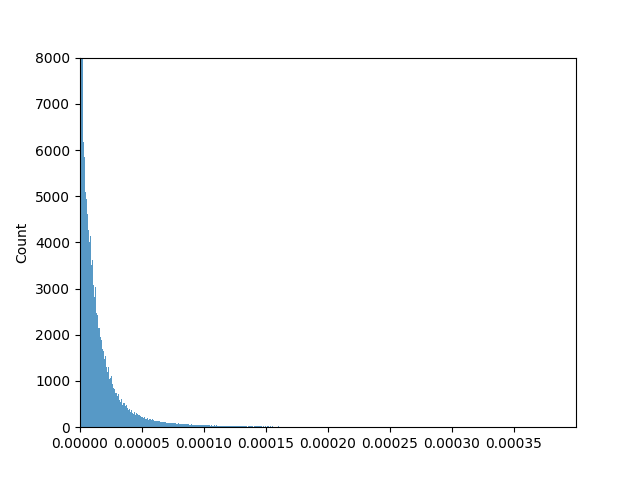}
    % \caption{layer 2}
  \end{subfigure}
  \hfill
  \begin{subfigure}{0.15\textwidth}
    \includegraphics[width=\textwidth]{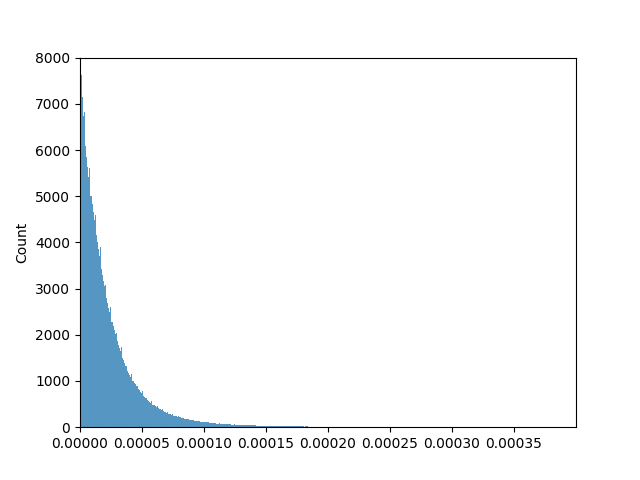}
    % \caption{layer 11}
  \end{subfigure}
  \hfill
  \begin{subfigure}{0.15\textwidth}
    \includegraphics[width=\textwidth]{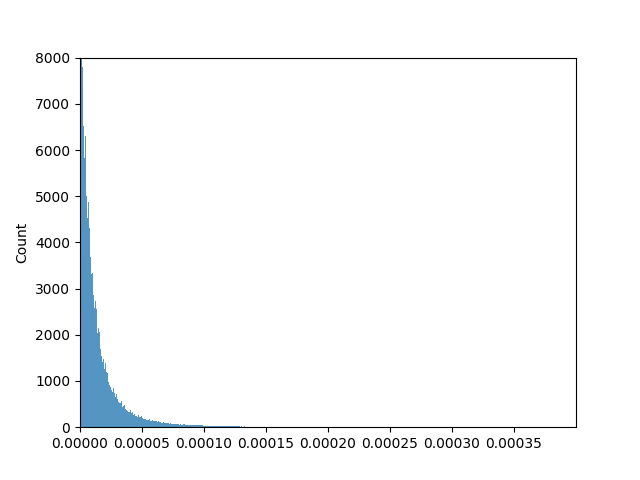}
    % \caption{layer 16}
  \end{subfigure}
  \hfill
  \begin{subfigure}{0.15\textwidth}
    \includegraphics[width=\textwidth]{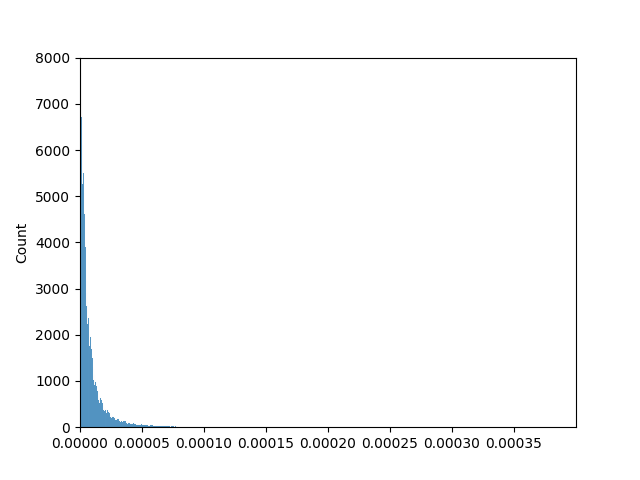}
    % \caption{layer 24}
  \end{subfigure}
  \hfill
  \begin{subfigure}{0.15\textwidth}
    \includegraphics[width=\textwidth]{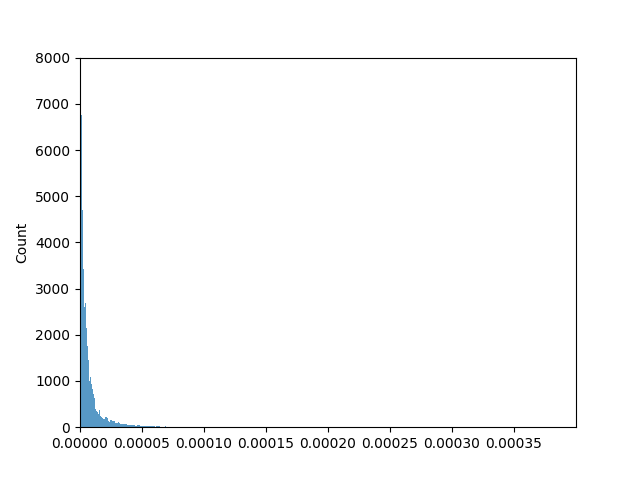}
    % \caption{layer 31}
  \end{subfigure}
  \hfill

  \begin{subfigure}{0.15\textwidth}
    \includegraphics[width=\textwidth]{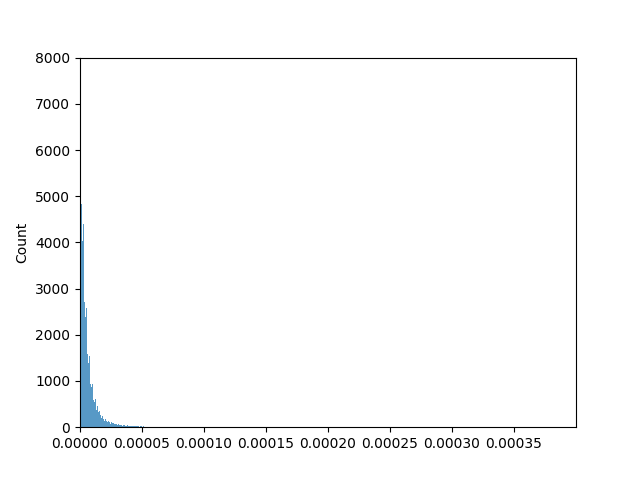}
    \caption{layer 1}
  \end{subfigure}
  \hfill
  \begin{subfigure}{0.15\textwidth}
    \includegraphics[width=\textwidth]{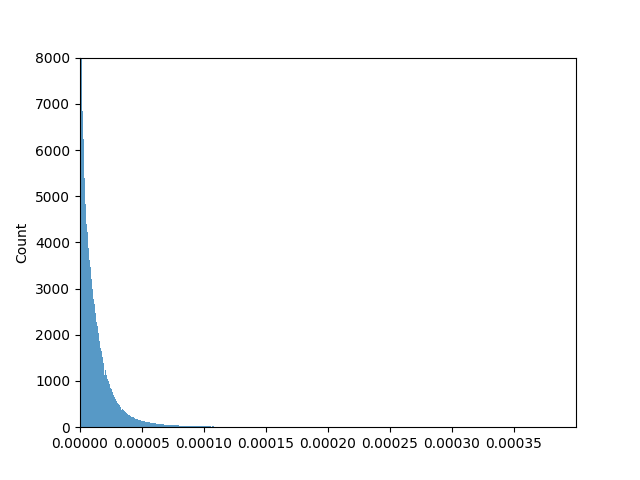}
    \caption{layer 2}
  \end{subfigure}
  \hfill
  \begin{subfigure}{0.15\textwidth}
    \includegraphics[width=\textwidth]{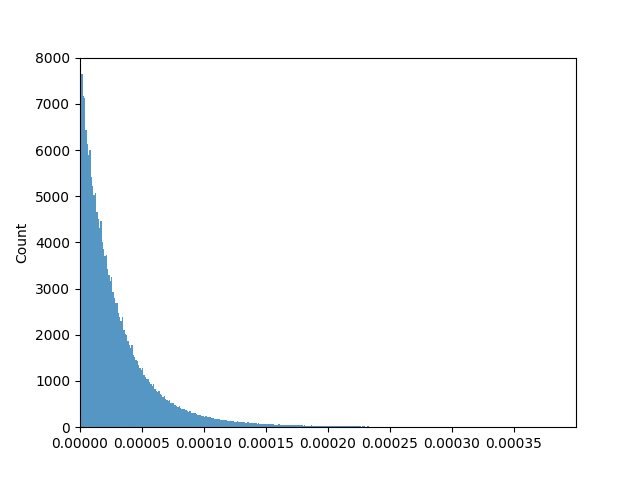}
    \caption{layer 9}
  \end{subfigure}
  \hfill
  \begin{subfigure}{0.15\textwidth}
    \includegraphics[width=\textwidth]{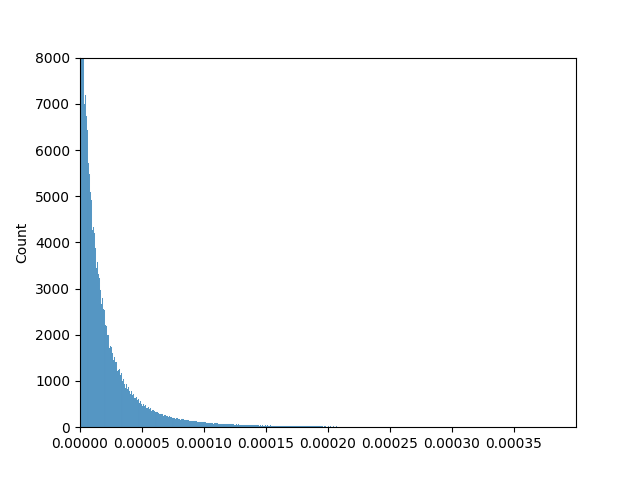}
    \caption{layer 16}
  \end{subfigure}
  \hfill
  \begin{subfigure}{0.15\textwidth}
    \includegraphics[width=\textwidth]{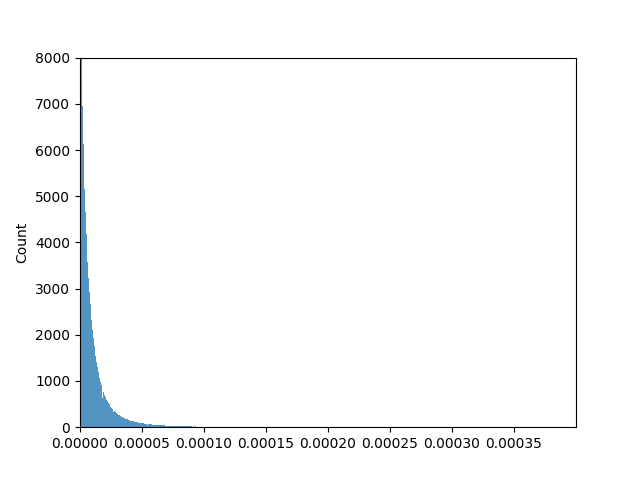}
    \caption{layer 24}
  \end{subfigure}
  \hfill
  \begin{subfigure}{0.15\textwidth}
    \includegraphics[width=\textwidth]{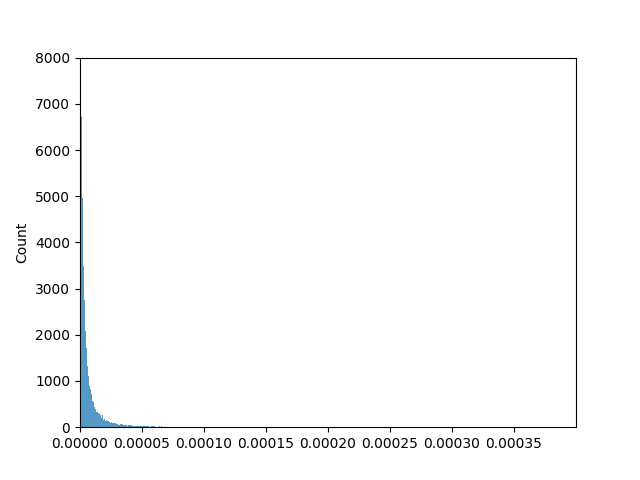}
    \caption{layer 31}
  \end{subfigure}
  \hfill

  \caption{Activated Parameter Statistics. The x-axis represents the value of $A(w_i)$, and the y-axis represents the quantity. For convenience of statistics, we have performed statistical processing, so the numerical value on the y-axis is an estimate of one-thousandth of the actual value. The first row is the statistical image for Boolq, the second row is for HumanEval, and the third row is the activation statistics for MMLU.}
  \label{fig:statistic}
\end{figure}

\subsection{Finding 1: Parameter Activation Patterns for Inputs in the Same Domain}\label{sec:ana1}

In this section, we attempt to analyze the distinct behaviors within Large Language Models (LLMs). Using our defined $\mathcal{A}(w_i)$, we analyze the distribution of activated parameters in different layers when faced with a single input.

In Figure~\ref{fig:statistic}, we present the statistical results from three data sources: Boolq~\cite{clark2019boolq}, HumanEval~\cite{chen2021codex}, and MMLU~\cite{hendrycks2020measuring}. For each dataset, we selected 64 samples and averaged the activation status of each parameter facing different samples. When calculating the activation status of data in different layers, we collectively consider all the tunable parameters within a layer. Specifically, this includes parameters from seven parts: the fully connected linear layers of Q, K, V, O, and the three fully connected linear layers of the MLP layer. To facilitate the statistics, we sort the $\mathcal{A}(w_i)$ values of the parameters, take the average every 1000 units, reducing the original 20,000 parameters to 20, and then perform the distribution statistics.

From the statistical results, we have selected the distribution characteristics of representative layers 1, 2, 9, 16, 24, and 31 for display. The full results can be referred to in the appendix. As shown in the figure, we can find that fewer parameters are activated in the first layer, meaning that only a small portion of parameters have a significant impact on the results. In layers 2-9, the parameters that have a greater impact on the results gradually increase. In the relatively deeper layers, the parameters that have a significant impact on the results decrease, concentrating on specific parts. This phenomenon is consistent across the three data sets representing different abilities. This leads us to speculate that for a single task, apart from the first layer, many parameters in the shallow layers are involved in the calculation of the results. Conversely, in the deep layers and the first layer, only a few parameters have a significant impact on the results.

\begin{figure}[t]
  \begin{subfigure}{0.32\textwidth}
    \includegraphics[width=\textwidth]{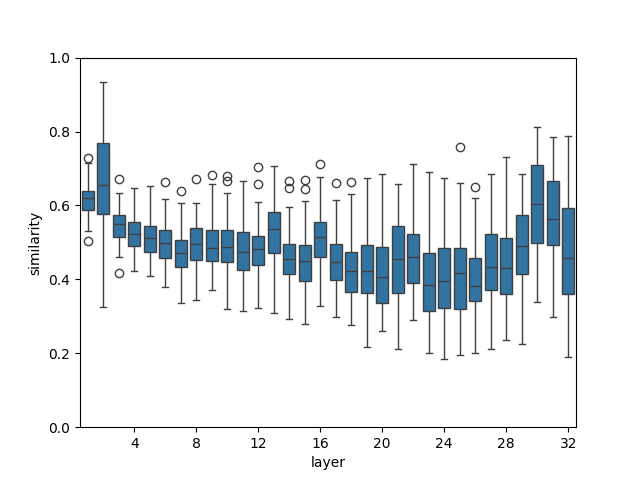}
    \caption{Boolq-Boolq}
  \end{subfigure}
  \hfill
  \begin{subfigure}{0.32\textwidth}
    \includegraphics[width=\textwidth]{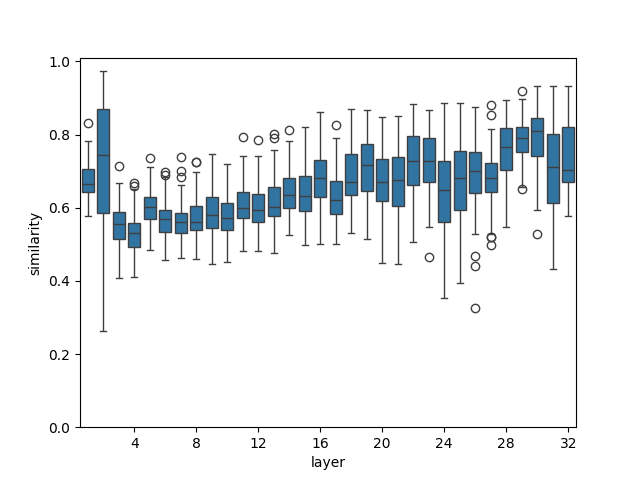}
    \caption{MMLU-MMLU}
  \end{subfigure}
  \hfill
  \begin{subfigure}{0.32\textwidth}
    \includegraphics[width=\textwidth]{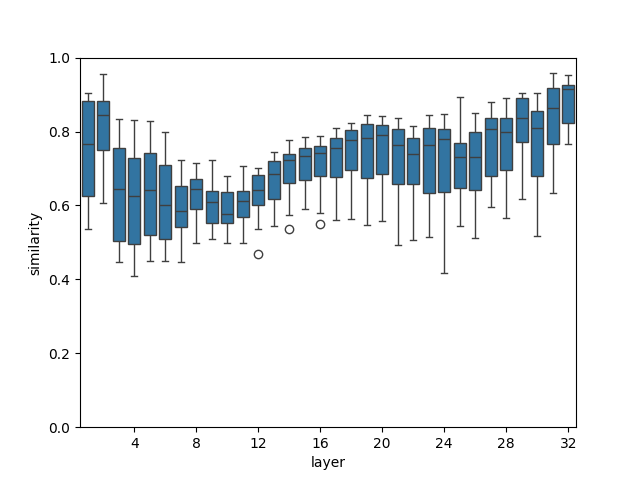}
    \caption{HumanEval-HumanEval}
  \end{subfigure}
  \begin{subfigure}{0.32\textwidth}
    \includegraphics[width=\textwidth]{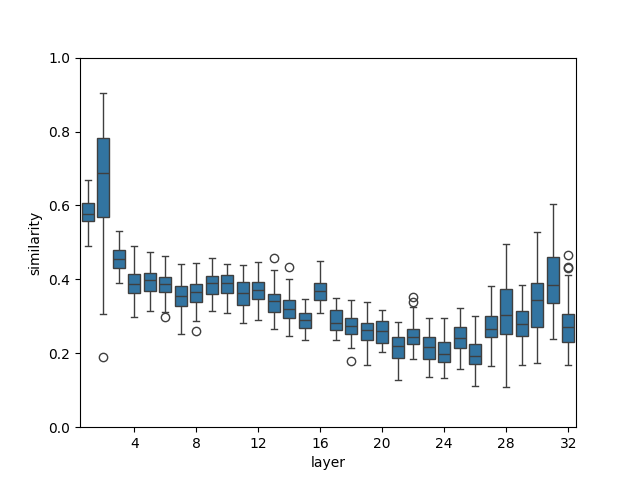}
    \caption{Boolq-MMLU}
  \end{subfigure}
  \hfill
  \begin{subfigure}{0.32\textwidth}
    \includegraphics[width=\textwidth]{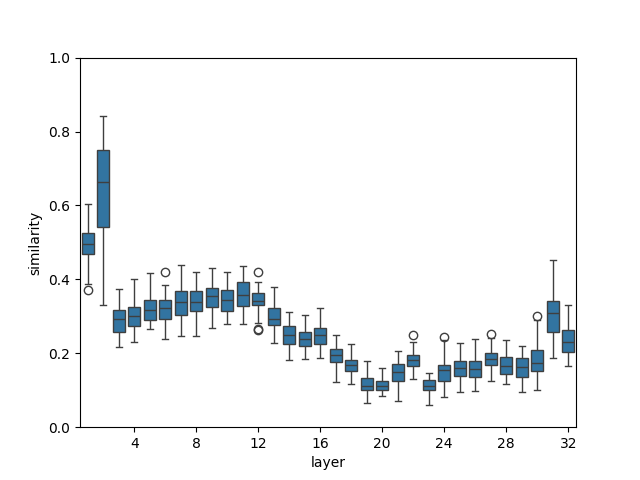}
    \caption{Boolq-HumanEval}
  \end{subfigure}
  \hfill
  \begin{subfigure}{0.32\textwidth}
    \includegraphics[width=\textwidth]{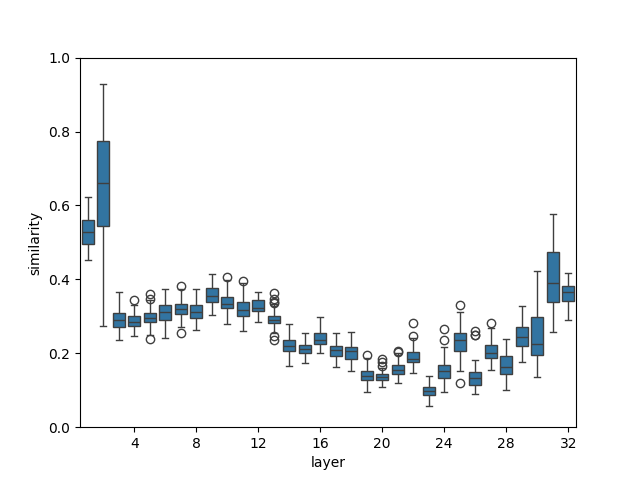}
    \caption{MMLU-HumanEval}
  \end{subfigure}
  \caption{Statistics of LLMDcos values for different layers. The x-axis represents the layer number, and the y-axis represents the LLMDcos values of 64 samples. The two data sources are indicated below the image. All images are results from Llama2-7b.}
  \label{fig:main}
\end{figure}
\subsection{Finding 2: Parameter Activation Patterns for Inputs in the Different Domains}\label{sec:ana2}

In order to observe the functionality of different layers within Large Language Models (LLMs), we have documented the activation scenarios of various layers when faced with input from different data domains. In response to inputs from two data domains, we calculated the LLMDcos for each layer each time. 

As depicted in Figure~\ref{fig:main}, we conducted an analysis on data from three data domains: Boolq, HumanEval, and MMLU. Apart from HumanEval-HumanEval where we only experimented with 16 sample groups, we statistically analyzed 64 sample groups in all other experiments. Consistent with the previous section, within the same layer, we included the parameters from seven parts in our statistics.

From the results, we observed that when faced with the same data source, the activation levels of different layers were similar. When faced with different data sources, the first 12 layers of the network had higher LLMDcos, while the later layers had lower similarity. Notably, the second layer had a high degree of similarity in any analysis between two data sources. Moreover, when faced with empirically similar data sources, the similarity in activation levels of the later layers was relatively high, that is, MMLU and Boolq showed relatively high similarity, while the activation distribution of both datasets had a significant difference from the HumanEval dataset. Combining the conclusions from the previous section, we can speculate that for different tasks, the shallow end of the network has a more general understanding ability. When faced with different tasks, similar parameters are activated to understand the problem, especially in the second layer. In the deeper layers, the network has relatively dispersed parameters, that is, some parameters are activated for specific tasks.

\subsection{Finding 3: Observing Data through Activated Distribution}\label{sec:data}

Through the aforementioned experiments, we observed that for different datasets, the similarity of the deep layers in the model significantly decreases, while for the same dataset, the similarity between the shallow and deep layers of the model remains consistent. This leads us to hypothesize that the similarity in the later layers may be related to semantic similarity. To validate this hypothesis, we test the similarity of more datasets in the Section~\ref{sec2:val3}, as well as conduct tests on benchmarks for semantic similarity.

\subsection{Generality Across Different Models}

\begin{figure}[t]
  \begin{subfigure}{0.32\textwidth}
    \includegraphics[width=\textwidth]{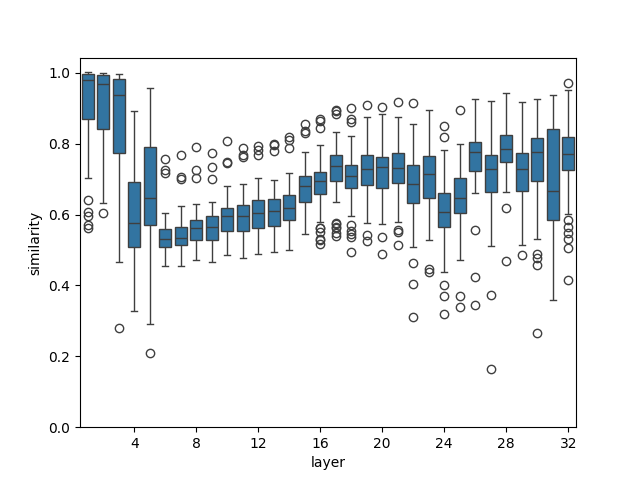}
    \caption{Llama-7B \\ MMLU-MMLU}
  \end{subfigure}
  \hfill
  \begin{subfigure}{0.32\textwidth}
    \includegraphics[width=\textwidth]{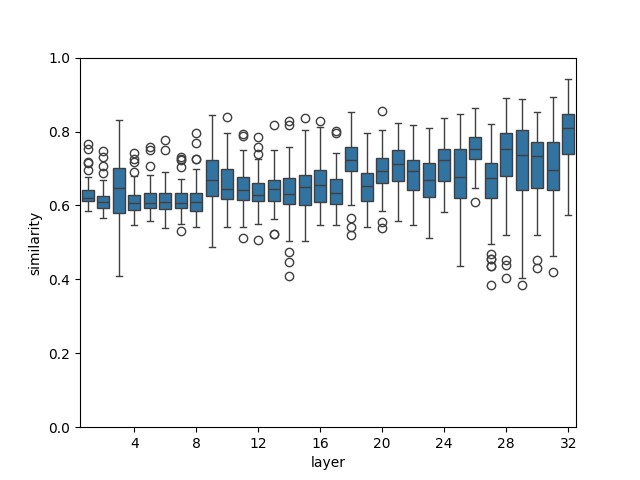}
    \caption{Qwen-7B \\ MMLU-MMLU}
  \end{subfigure}
  \hfill
  \begin{subfigure}{0.32\textwidth}
    \includegraphics[width=\textwidth]{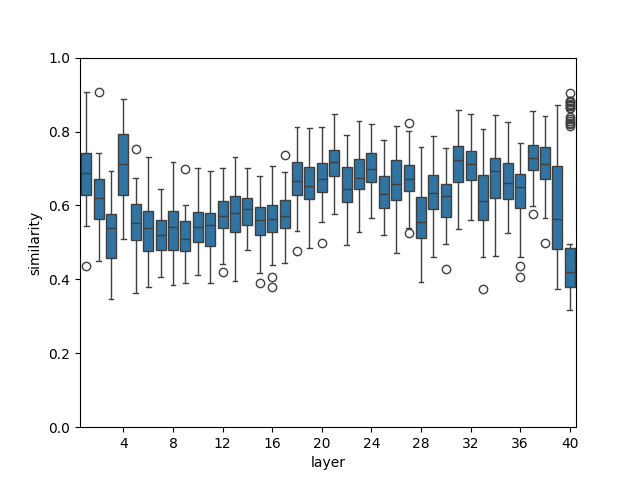}
    \caption{Llama2-13B \\ MMLU-MMLU}
  \end{subfigure}
  \begin{subfigure}{0.32\textwidth}
    \includegraphics[width=\textwidth]{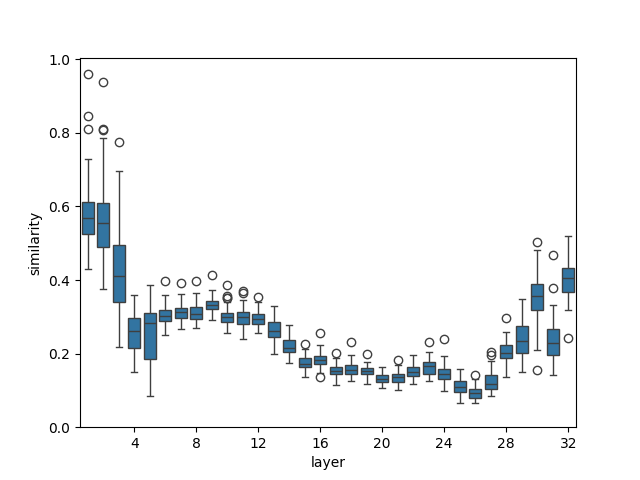}
    \caption{Llama-7B \\ MMLU-HumanEval}
  \end{subfigure}
  \hfill
  \begin{subfigure}{0.32\textwidth}
    \includegraphics[width=\textwidth]{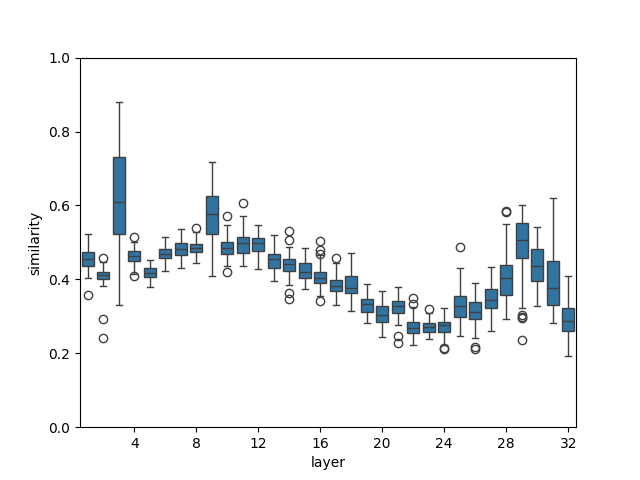}
    \caption{Qwen-7B \\ MMLU-HumanEval}
  \end{subfigure}
  \hfill
  \begin{subfigure}{0.32\textwidth}
    \includegraphics[width=\textwidth]{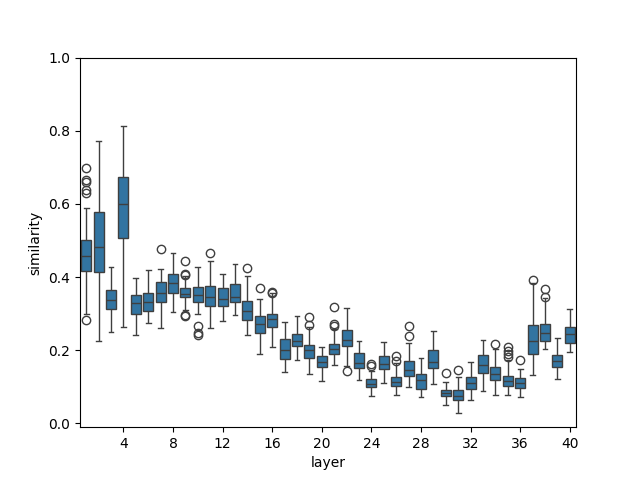}
    \caption{Llama2-13B \\ MMLU-HumanEval}
  \end{subfigure}
  \caption{Statistics of LLMDcos values for different models. The x-axis represents the layer number, and the y-axis represents the LLMDcos values of 64 samples. The two data sources and the model are indicated below the image. The images are consistent with the results from Llama2-7B.}
  \label{fig:diff_model}
\end{figure}

To determine the universality of our results, rather than their specificity to the Llama series or the 7B models, we performed experiments analogous to those in Section 3.2 on Qwen-7b~\cite{bai2023qwen}, Llama2-13b and Llama-7b~\cite{touvron2023llama1}. The results are illustrated in the Figure~\ref{fig:diff_model}.

It is noteworthy that the results from llama-7b are remarkably similar to those from llama2-7b. Although the outcomes from qwen-7b and llama2-13b deviate somewhat from those of llama2-7b, the overall trend remains consistent, i.e., there is an extremely high degree of similarity in certain layers at the shallow part, with the shallow end generally exhibiting a higher degree of similarity and the deep half showing a lower degree of similarity.

% 规划一下 前3页差不多就这样了
% 4-5页写分析：
% 逐层单数据源分布，逐层双数据源统计，这两个内容可以占1整页，
% 同数据统计，考虑6种：阅读理解、代码、数学、常识QA、世界知识
% 这里我们只考虑阅读理解（boolq）、常识QA（siqa， hellaSwag）、代码（humaneval）、数学（gsm8k）、世界知识（wikipedia2）、综合（mmlu、c4） 8*8图像可以画2个，这个分析可以写一页半
% 6-7页写实验
% 剪枝：两种减法：1、换角度剪。 2、换方向剪。
% 数据相关性度（包的，stsb0.5， 1e-9，非常完美）
% 8页写discussion
% 不同模型：qwen、13B。 不同语言：中文、英文、法文
% 
% 9页写相关工作、limitation和总结
% 

% Avoid using obscure vocabulary, strive for concise sentence structures, and employ academic English to rewrite and polish the aforementioned content.
% 请将以下内容翻译为学术英语，避免使用偏僻词。

\section{Validation Experiments}

Given that the assessments in Section~\ref{sec:define} are entirely based on our definition of parameter activation levels. This paper, drawing upon the preceding analysis, suggests several application approaches. The success of these applications substantiates the validity of our analytical results.

% \subsection{pruning LLMs with different sparasity of different layers}
\subsection{Validation 1: Pruning LLMs with Different Sparasity According to the Activation Level}\label{val1}

Based on the analysis results from Section~\ref{sec:ana1}, we can observe the following phenomena: For a 32-layer Llama-2-7b-hf network, most of the parameters in the first layer and the deep half of the network have little impact on the results. However, in the parameters of the 2-17 layers of the network, there are relatively more parameters that have a significant impact on the results. To validate this conclusion, we will prune the model, making the 2-17 layers of the network more sparse, while the 1st and 18-32 layers of the network have a smaller degree of sparsity.

We employed the unstructured pruning method proposed by \cite{sun2023simple}, and while keeping all other settings unchanged, we set the 2-17 layers (2-21 for Llama2-13B) to be pruned by only 45\%, while the 1st and 18-32 (1st and 22-40 for Llama2-13B) layers were pruned by 55\% in the setting where 50\% of the whole network was pruned. To compare the test results of the network, we conducted tests on two different metrics on six datasets based on the original settings. All the calibtration dataset is C4~\cite{raffel2020exploring}. The evaluation datasets include Wikitext2~\cite{merity2016pointer}, Boolq, SIQA~\cite{sap2019socialiqa}, PIQA~\cite{Bisk2020}, Hellaswag~\cite{zellers2019hellaswag}, and MMLU.

As can be seen from the results in the Table~\ref{tab:01_prune_main}, we observe that our method consistently improves the Perplexity (PPL) value on Wikitext2 across all model results, suggesting that our approach can generally enhance the language modeling capability of the models. In the zero-shot results, it is worth noting that the improvements in Hellaswag are universal, while Boolq generally experiences a decline. In conjunction with the results from the Figure~\ref{fig:enter-label}, we find that Hellaswag has the highest correlation with C4, while Boolq is relatively lower. Therefore, the results of using C4 as the calibration set are less satisfactory in Boolq and other evaluations.

% \subsection{pruning LLMs with different sparasity of different layers}
\subsection{Validation 2: Pruning LLMs with Different Calibration Set}

% \begin{table}[t!]
\begin{table*}[t]
% \begin{minipage}[b]{0.45\textwidth}
\centering

     \footnotesize
 % \resizebox{1\linewidth}{!}{%
\begin{tabular}{lccccccc}
			\toprule
\multirow{2}{*}{\textbf{Models}}  & \multirow{2}{*}{\textbf{pruning method}} & \multirow{2}{*}{wikitext2} & \multicolumn{4}{c}{zero-shot} & \multirow{2}{*}{MMLU} \\
\cline{4-7}

&& &Boolq & SIQA & PIQA & hellaswag &    \\

\midrule
\multirow{2}{*}{Llama-7B} & Wanda & 7.26 & 66.41 & 35.94 & 52.73 & 29.30 & 20.70 \\
                    & Wanda(ours) & 7.19 & 62.50 & 35.94 & 52.34 & 29.69 & 25.78 \\
\midrule
\multirow{2}{*}{Llama2-7B} & Wanda & 6.46 & 77.73 & 40.23 & 51.56 & 29.30 & 37.11 \\
                     & Wanda(ours) & 6.38 & 73.05 & 41.41 & 51.17 & 34.38 & 35.55 \\
\midrule
\multirow{2}{*}{Llama2-13B} & Wanda & 5.58 & 81.64 & 55.47 & 53.91 & 50.00 & 42.58 \\
                      & Wanda(ours) & 5.52 & 80.08 & 55.47 & 55.86 & 55.08 & 41.41 \\

\bottomrule
\end{tabular}
% }

\caption{
    Pruning experiment results. All models have an overall pruning ratio of 50\%, with c4 as the calibration set. The evaluation metric for wikitext2 is perplexity, and for MMLU it is 5-shot. Wanda (ours) refers to our Wanda model after layer-by-layer adjustment of the pruning ratio, ensuring the overall pruning ratio remains the same.
}
%	\vspace{-0.5ex}
    % \tabnote{$^{\rm a}$All the training accuracy is higher than $95\%$ except  {VGG-16} 92.20 in ImageNet sampled by \methodRandom.}
	\label{tab:01_prune_main}
\end{table*}
% \end{table}
% \end{minipage}

% \begin{table}[t!]
\begin{table*}[t]
% \begin{minipage}[b]{0.45\textwidth}
\centering

     \footnotesize
 % \resizebox{1\linewidth}{!}{%
\begin{tabular}{lccccccc}
			\toprule
\multirow{2}{*}{\textbf{Models}}  & \multirow{2}{*}{\textbf{Calibration set}} & \multirow{2}{*}{wikitext2} & \multicolumn{4}{c}{zero-shot} & \multirow{2}{*}{MMLU} \\
\cline{4-7}

&& &Boolq & SIQA & PIQA & hellaswag &    \\

\midrule
\multirow{2}{*}{Llama-7B} & Boolq & 7.22 & 63.67 & 39.84 & 52.73 & 24.61 & 25.78 \\
                           & SIQA & 7.44 & 60.94 & 34.38 & 52.34 & 25.78 & 24.61 \\
\midrule
\multirow{2}{*}{Llama2-7B} & Boolq & 6.44 & 73.83 & 42.97 & 51.56 & 32.42 & 35.16 \\
                            & SIQA & 6.67 & 71.88 & 46.48 & 53.13 & 32.03 & 36.33 \\
\midrule
\multirow{2}{*}{Llama2-13B} & Boolq & 5.57 & 80.08 & 57.03 & 53.91 & 53.91 & 43.36 \\
                             & SIQA & 5.70 & 75.39 & 57.03 & 57.81 & 50.39 & 47.27 \\

\bottomrule
\end{tabular}
% }

\caption{
    Pruning experiment results. In the figure, all models have an overall pruning ratio of 50\%, and the pruning method used is Wanda (ours). The evaluation metric for wikitext2 is perplexity, and for MMLU it is 5-shot.
}
%	\vspace{-0.5ex}
    % \tabnote{$^{\rm a}$All the training accuracy is higher than $95\%$ except  {VGG-16} 92.20 in ImageNet sampled by \methodRandom.}
	\label{tab:02_prune_calib}
\end{table*}
% \end{table}
% \end{minipage}

Based on the results from Section~\ref{sec:ana2}, we observe the following: For a 32-layer Llama-2-7b-hf network, when faced with different data domains, the majority of parameters in the deep part of the network exhibit a lower degree of activation similarity. In contrast, in the shallow layers of the network, there is a relatively higher degree of similarity in the distribution of parameter activation. This leads us to hypothesize that the shallow layers of the network consist of more generic parameters, while the deep layers are discretely composed of parameters that address different problems. To validate this conclusion, we will modify the calibration set to specifically prune for specialized tasks, under the premise of pruning different layers of the model at varying proportions as outlined in the previous section. We will then verify whether this results in a decrease in performance on other test results.

We adopted the pruning method from Section~\ref{val1}: Wanda (ours), ensuring all other settings remained unchanged. We adjusted the calibration set to be the same length as Boolq and SIQA. To ensure the sentence length of the calibration set is the same, we concatenated different Boolq and SIQA data, including answers, into long sentences and cut them to a specific length (2048 for the 7b model, 4096 for the 13b model).

The results are shown in the Table~\ref{tab:02_prune_calib}. We observe that compared to the C4 dataset, the language modeling capability of the pruning results using Boolq as the calibration set continues to decline, while the PPL value of SIQA is even lower. This might be due to Boolq serving as reading material, which encompasses a wider range of knowledge, while the content of SIQA is relatively singular by comparison. For Llama-7b, the results on most zero-shot tasks declined due to the substantial loss of language modeling capability in SIQA. However, the results for Llama-2 align with our expectations. The pruning results using Boolq as the calibration set consistently outperform those based on C4 and SIQA on Boolq. Moreover, on PIQA, MMLU, and SIQA, three evaluations with stronger correlation with SIQA, the pruning model results using SIQA as the calibration set always outperform those of C4 and Boolq. This further validates our hypothesis.

\subsection{Validation 3: Semantic Similarity with LLMDcos}\label{sec2:val3}

In the analysis presented in Section~\ref{sec:data}, we observed that the similarity of the deep layers of the network decreases for different data domains. 

To validate the relationship between LLMDcos and data relevance, we tested its performance on a semantic similarity benchmark STS-B~\cite{cer2017semeval} and SICK~\cite{marelli-etal-2014-sick}. 

\begin{figure}[t]
    % \centering

\begin{minipage}{0.45\textwidth}  
\centering
% \centering
\begin{tabular}{l|cc}
\toprule
Model       & STS-B & SICK \\
\midrule
Llama-7B   & 0.30 & 0.52\\
Llama2-7B  & 0.66 & 0.51\\
Llama2-13B & 0.43 & 0.52\\
\bottomrule
\end{tabular}
\captionof{table}{Spearman correlation between LLMDcos and semantic similarity. We sampled 256 examples on each dataset, and all p-values were far less than 0.001. We use the LLMDcos of 20-30 layers (20-38 for Llama2-13B)}
\label{fig:sim}
\end{minipage}
\hfill
\begin{minipage}{0.45\textwidth}  
    \includegraphics[width=1\textwidth]{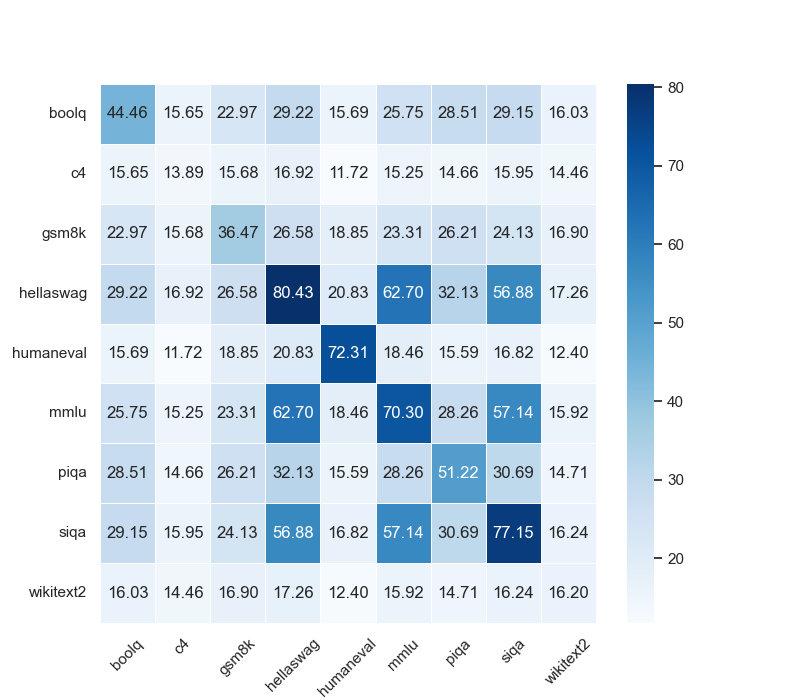}
    \caption{Dataset relevance. The figure shows the similarity calculated based on the mean LLMDcos of layers 16-29 in Llama-2-7b.}
    \label{fig:enter-label}
    
\end{minipage}
\end{figure}

As shown in the Table~\ref{fig:sim}, the LLMDcos calculated by Llama2-7B yielded the best results. As evident from the results, LLama2-7b achieved the best performance, while the results of Llama2-13B were relatively inferior. We believe that, compared to semantic similarity, LLMDcos assesses more of the similarity in the capabilities of the Large Language Models (LLMs) required by the inputs. For 13B models, these capabilities may be more densely represented in the parameters, leading to these deviations.

In addition to this, we calculated the similarity relationships across nine datasets, including Boolq, C4, GSM8K~\cite{cobbe2021gsm8k}, Hellaswag, HumanEval, MMLU, PIQA, SIQA, and Wikitext2. The results are shown in Figure~\ref{fig:enter-label}

% 规划一下 前3页差不多就这样了
% 4-5页写分析： done
% 逐层单数据源分布，逐层双数据源统计，这两个内容可以占1整页，
% 同数据统计，考虑6种：阅读理解、代码、数学、常识QA、世界知识
% 这里我们只考虑阅读理解（boolq）、常识QA（siqa， hellaSwag）、代码（humaneval）、数学（gsm8k）、世界知识（wikipedia2）、综合（mmlu、C4） 8*8图像可以画2个，这个分析可以写一页半
% 6-7页写实验
% 剪枝：两种减法：1、换角度剪。 2、换方向剪。
% 数据相关性度（包的，stsb0.5， 1e-9，非常完美）
% 8页写discussion
% 不同模型：qwen、13B。 不同语言：中文、英文、法文
% 
% 9页写相关工作、limitation和总结
% 

% Avoid using obscure vocabulary, strive for concise sentence structures, and employ academic English to rewrite and polish the aforementioned content.
% 请将以下内容翻译为学术英语，避免使用偏僻词。

% \section{discussion}

\section{related work}

\subsection{data similarity}

Our work reflects data relevance. In the field of NLP, data similarity mainly refers to text similarity. Text similarity in early machine learning was primarily based on statistical methods, such as word frequency and sentence length. However, in the era of deep learning, semantic similarity has become of greater interest, leading to the proposal of a series of models. Recently, as data training efficiency has been increasingly recognized, many data selection works that employ clustering ideas are based on data similarity.

Most of the relatively recent work has directly calculated relevance using the hidden state after the embedding layer. \cite{laskar2020contextualized} proposed two methods for calculating data similarity, one based on the average pooling of the token dimension of BERT's hidden state, and the other through training with the CLS token. Both methods assisted in the main task of answer selection. \cite{li2020sentence} improved the effect of similarity calculation by mapping data similarity to a Gaussian distribution based on a kernel distribution.

In addition, there has been little new development in evaluation datasets. The STS-B dataset has not been updated since 2017, and it consists only of similar short sentences. In the era of large language models, we may need a more generalized text similarity. 

\subsection{Role of Each Layer in LLMs}

In many causality inference papers, the functionality of different layers of the model is discussed. These studies employ probes, causal alignment, and some even more drastic measures, such as directly skipping certain layers to observe the difference in results. However, most of the work focuses on a specific function, rather than a macroscopic discussion of the functions of different layers.

For example, Zhao et al.~\cite{zhao2023causality} found that the third layer in Llama may have a significant impact on whether the model outputs toxic information. Azaria et al.~\cite{azaria2023internal} verified that the later layers of the model are more aware of whether they contain false information. Mcgrath et al.~\cite{mcgrath2023hydra} indicated that different layers in the model may have different information about numbers.

%In contrast, our work focuses more on the macroscopic differences between different layers, and our conjectures have been confirmed in the pruning process.

\section{limitation and future work}\label{sec:lim}

\subsection{limitations}

% The authors identify two main limitations of this study. On one hand, there are constraints due to GPU limitations, and on the other hand, there is a lack of theoretical proof.
We identify two main limitations of this study. On one hand, there are constraints due to GPU limitations, and on the other hand, there is a lack of theoretical proof.

Due to GPU memory limitations, we only have eight A40s for experimentation. These memory constraints prevent us from verifying the results of Llama-70b. 

In terms of theory, our activation degree algorithm is a very rough calculation. However, more refined calculations are constrained by time complexity and space complexity, and likewise cannot be performed on our machine. With more effective mathematical reasoning, we believe we can obtain more refined results.

\subsection{Future Work}

% We plan future work mainly includes two aspects: a better understanding of data distribution in the era of large models, and a better understanding of network weights.

In this paper, we find that the degree of activation may be related to data similarity, which inspires us to think about a new dimension of data relevance: What kind of capability is used to model the current sentence? Data similarity based on capability may play different roles in the pre-training of large models and SFT.

% The understanding of the capability of network weights is also something we will not give up. In the future, we will conduct more detailed function statistics for different layers, such as exploring why the similarity degree of the second layer of Qwen is not high. Is this related to the fact that English is not its only main language?

\section{Conclusion}

% This paper employ the methods motivated from model pruning to measure the activation level of internal parameters of the model when facing specific inputs. In the statistics of activation degree, this work finds that different layers of LLM reflect different activation distributions. 
% %We propose two hypotheses: 1) The front layers of the network have more activated weights, and more parameters influence the final calculation results, while the latter layers of the network have fewer activated weights, and fewer parameters influence the output of results. 2) The parameters of the front layers of the network have more generic capabilities, while the parameters of the latter layers are more discrete, and different parameters are only activated when facing specific tasks. 
% The paper conducted two experiments in pruning, involving three different models from the Llama series and seven different datasets. At the same time, we observed that the similarity of network activation has a strong correlation with data similarity. We calculated its performance on the STSB dataset, and computed the data similarity results for nine datasets, which also mirrored the pruning experiments. We hope that our work can advance researchers' understanding of LLM.
This paper proposes a metric for measuring the activation patterns of internal parameters in language models, and subsequently introduces LLMDcos to calculate the similarity of internal network activations when facing inputs from two different data domains. Based on this metric, we made three discoveries, each reflecting that the shallow layers of the network are more generic, the deep layers possess more specific capabilities, and the deep layer's LLMDcos is related to data similarity. To validate our findings, we designed three corresponding verification experiments. Two pruning experiments respectively verified the differences in the distribution of internal parameter activations in different model layers. Furthermore, the results on the semantic similarity benchmark also reflected that the deep layer's LLMDcos can represent data similarity. We hope that our work can advance researchers' understanding of LLM.

% \bibliographystyle{plain}
% \bibliography{neurips_2024}

%%%%%%%%%%%%%%%%%%%%%%%%%%%%%%%%%%%%%%%%%%%%%%%%%%%%%%%%%%%%

\appendix
\newpage
\section{Appendix}

\subsection{Experiments Setting}

We use the implementation of Wanda from \url{https://github.com/locuslab/wanda}. We only use unstructed pruning with 50\% sparasity.

All models and datasets used in this paper are sourced from \url{huggingface.co}.

All experiments in this paper were conducted on a single machine equipped with eight 50G A40 GPU. All the single experiment can be finished with in half an hour.

All the model are loaded with torch.float16 except for the validation experiments 3, which we use float 32 instead.

\subsection{Statistics for Finding 1 on Boolq}

\begin{figure}[htbp]
  \begin{subfigure}{0.15\textwidth}
    \includegraphics[width=\textwidth]{images/boolq/0.png}
    \caption{layer 1}
  \end{subfigure}
  \hfill
  \begin{subfigure}{0.15\textwidth}
    \includegraphics[width=\textwidth]{images/boolq/1.png}
    \caption{layer 2}
  \end{subfigure}
  \hfill
  \begin{subfigure}{0.15\textwidth}
    \includegraphics[width=\textwidth]{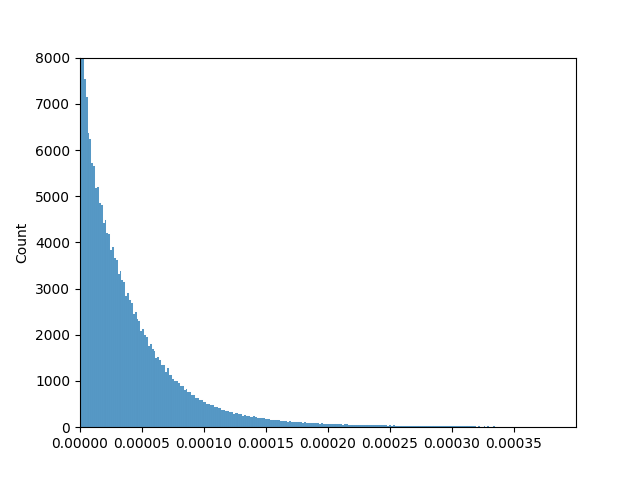}
    \caption{layer 3}
  \end{subfigure}
  \hfill
  \begin{subfigure}{0.15\textwidth}
    \includegraphics[width=\textwidth]{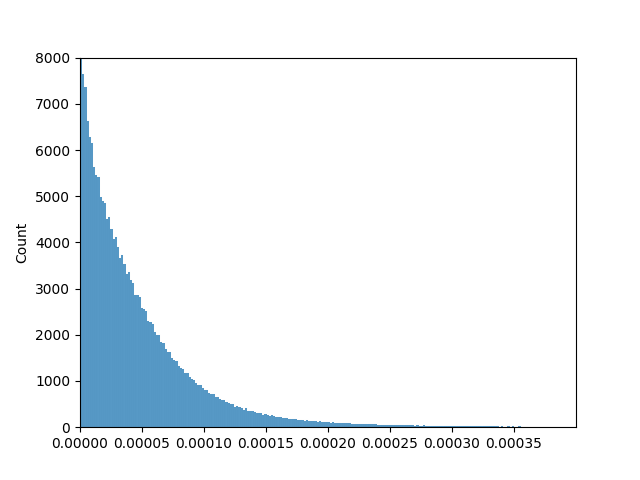}
    \caption{layer 4}
  \end{subfigure}
  \hfill
  \begin{subfigure}{0.15\textwidth}
    \includegraphics[width=\textwidth]{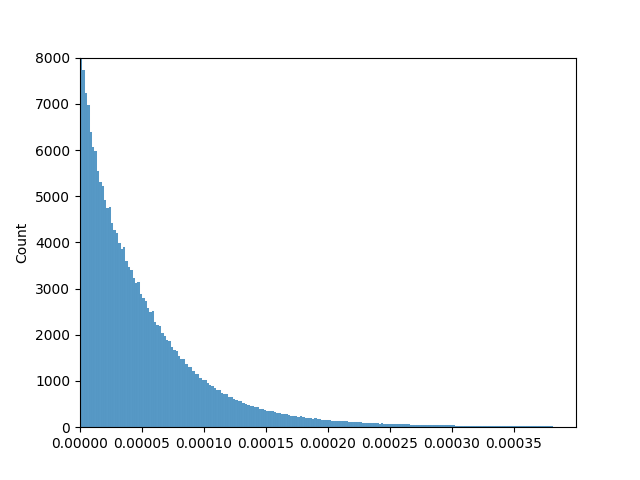}
    \caption{layer 5}
  \end{subfigure}
  \hfill
  \begin{subfigure}{0.15\textwidth}
    \includegraphics[width=\textwidth]{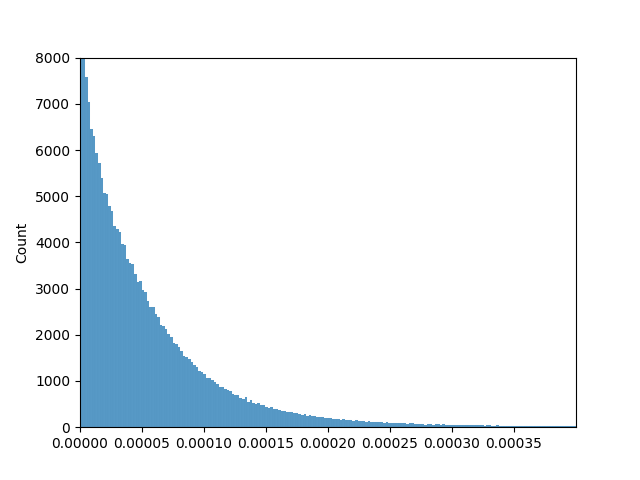}
    \caption{layer 6}
  \end{subfigure}
  \hfill
  \begin{subfigure}{0.15\textwidth}
    \includegraphics[width=\textwidth]{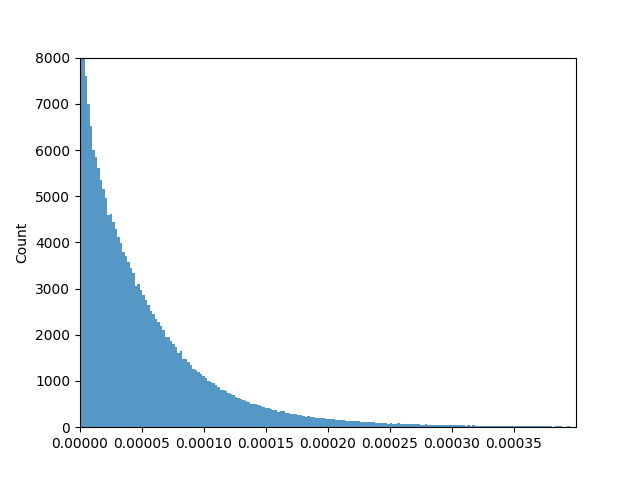}
    \caption{layer 7}
  \end{subfigure}
  \hfill
  \begin{subfigure}{0.15\textwidth}
    \includegraphics[width=\textwidth]{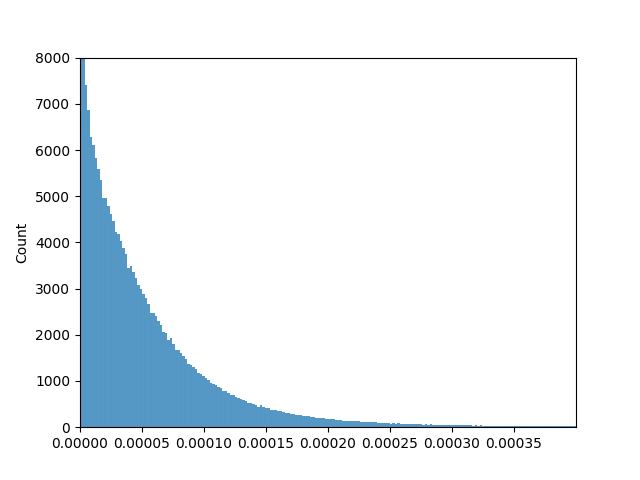}
    \caption{layer 8}
  \end{subfigure}
  \hfill
  \begin{subfigure}{0.15\textwidth}
    \includegraphics[width=\textwidth]{images/boolq/8.png}
    \caption{layer 9}
  \end{subfigure}
  \hfill
  \begin{subfigure}{0.15\textwidth}
    \includegraphics[width=\textwidth]{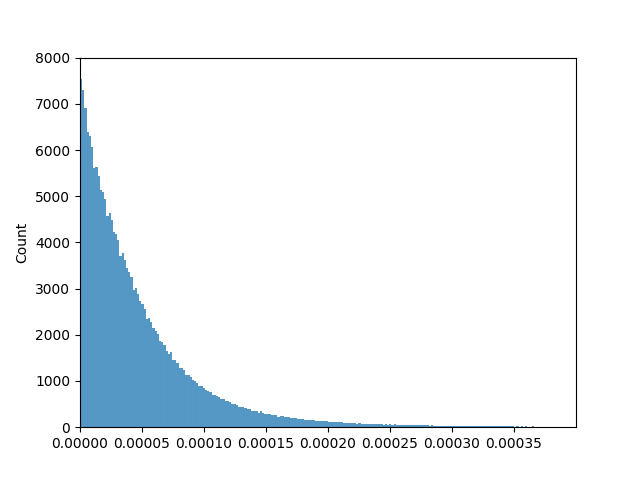}
    \caption{layer 10}
  \end{subfigure}
  \hfill
  \begin{subfigure}{0.15\textwidth}
    \includegraphics[width=\textwidth]{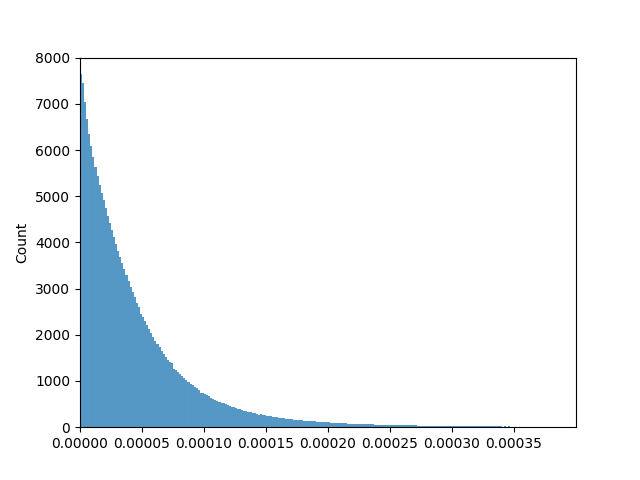}
    \caption{layer 11}
  \end{subfigure}
  \hfill
  \begin{subfigure}{0.15\textwidth}
    \includegraphics[width=\textwidth]{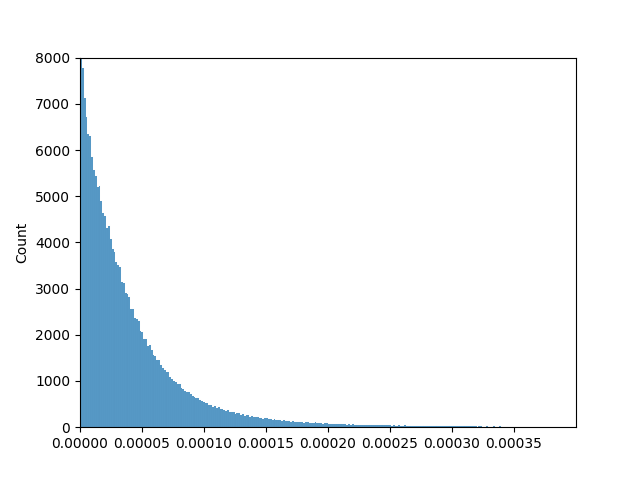}
    \caption{layer 12}
  \end{subfigure}
  \hfill
  \begin{subfigure}{0.15\textwidth}
    \includegraphics[width=\textwidth]{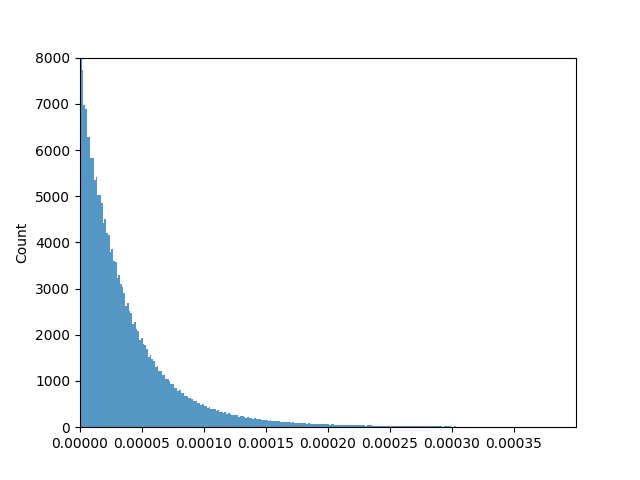}
    \caption{layer 13}
  \end{subfigure}
  \hfill
  \begin{subfigure}{0.15\textwidth}
    \includegraphics[width=\textwidth]{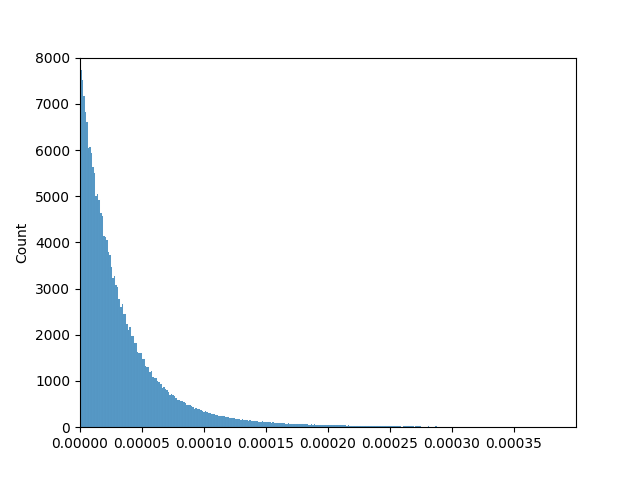}
    \caption{layer 14}
  \end{subfigure}
  \hfill
  \begin{subfigure}{0.15\textwidth}
    \includegraphics[width=\textwidth]{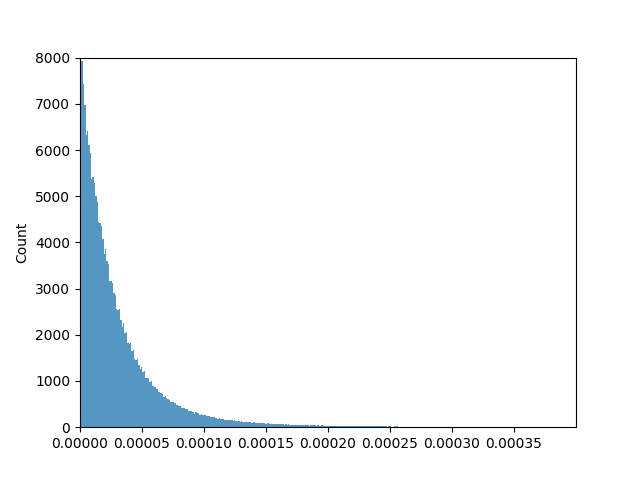}
    \caption{layer 15}
  \end{subfigure}
  \hfill
  \begin{subfigure}{0.15\textwidth}
    \includegraphics[width=\textwidth]{images/boolq/15.png}
    \caption{layer 16}
  \end{subfigure}
  \hfill
  \begin{subfigure}{0.15\textwidth}
    \includegraphics[width=\textwidth]{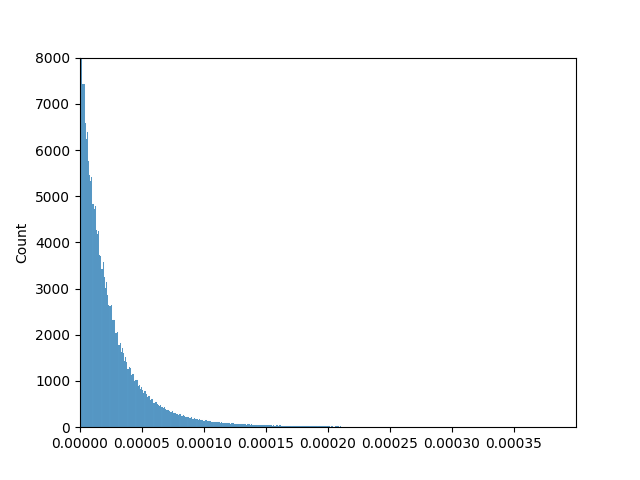}
    \caption{layer 17}
  \end{subfigure}
  \hfill
  \begin{subfigure}{0.15\textwidth}
    \includegraphics[width=\textwidth]{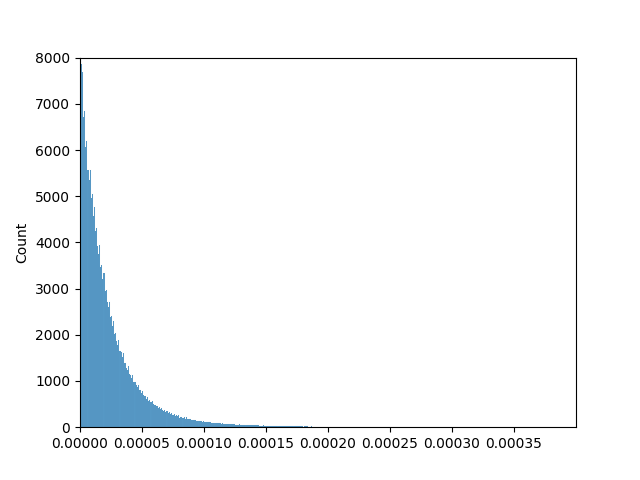}
    \caption{layer 18}
  \end{subfigure}
  \hfill
  \begin{subfigure}{0.15\textwidth}
    \includegraphics[width=\textwidth]{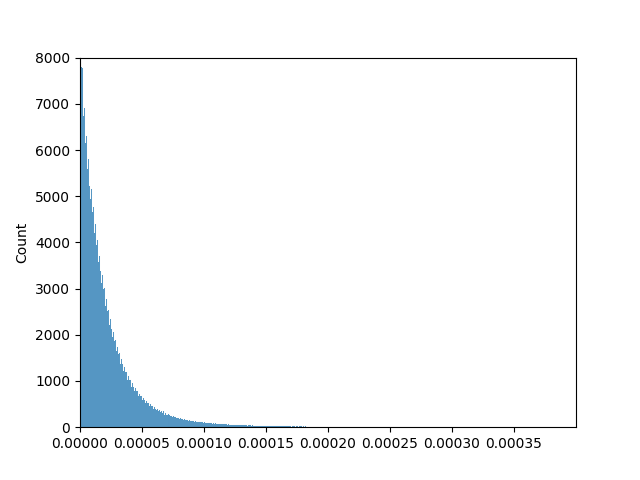}
    \caption{layer 19}
  \end{subfigure}
  \hfill
  \begin{subfigure}{0.15\textwidth}
    \includegraphics[width=\textwidth]{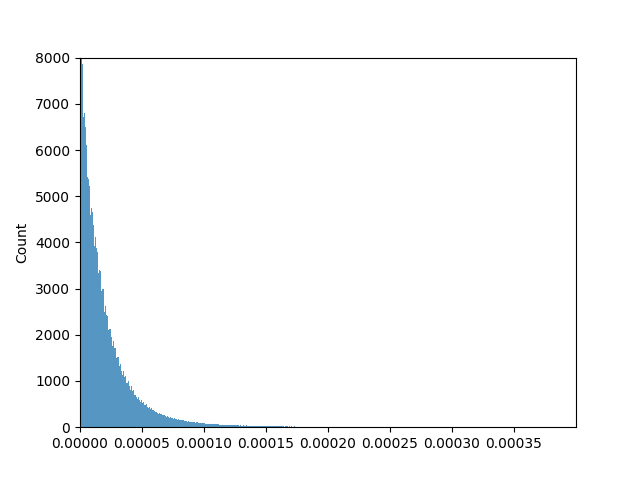}
    \caption{layer 20}
  \end{subfigure}
  \hfill
  \begin{subfigure}{0.15\textwidth}
    \includegraphics[width=\textwidth]{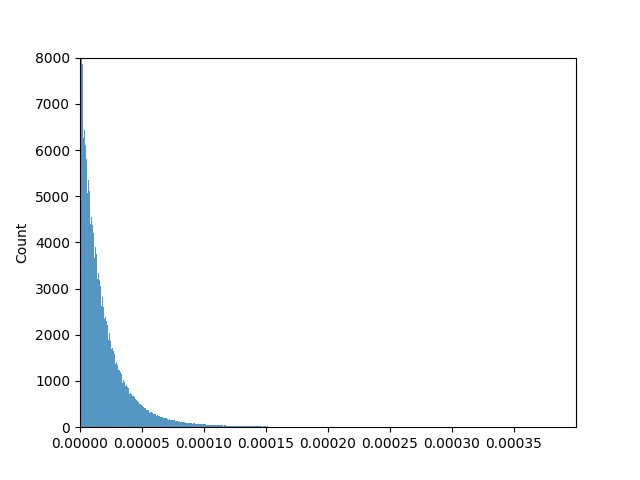}
    \caption{layer 21}
  \end{subfigure}
  \hfill
  \begin{subfigure}{0.15\textwidth}
    \includegraphics[width=\textwidth]{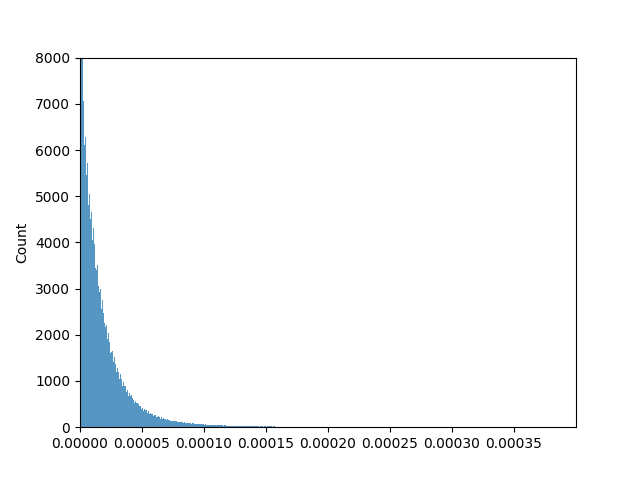}
    \caption{layer 22}
  \end{subfigure}
  \hfill
  \begin{subfigure}{0.15\textwidth}
    \includegraphics[width=\textwidth]{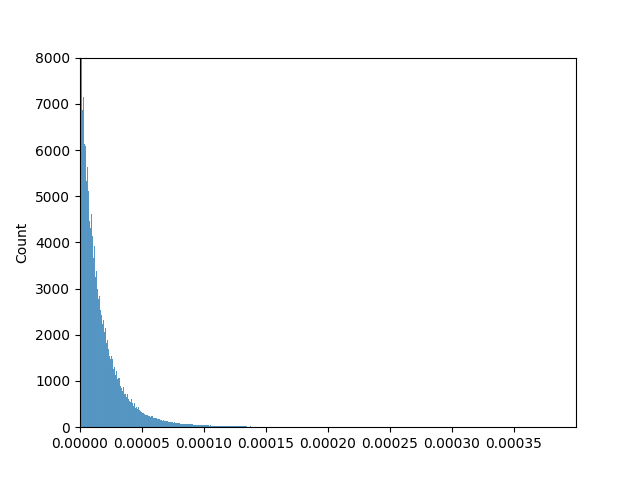}
    \caption{layer 23}
  \end{subfigure}
  \hfill
  \begin{subfigure}{0.15\textwidth}
    \includegraphics[width=\textwidth]{images/boolq/23.png}
    \caption{layer 24}
  \end{subfigure}
  \hfill
  \begin{subfigure}{0.15\textwidth}
    \includegraphics[width=\textwidth]{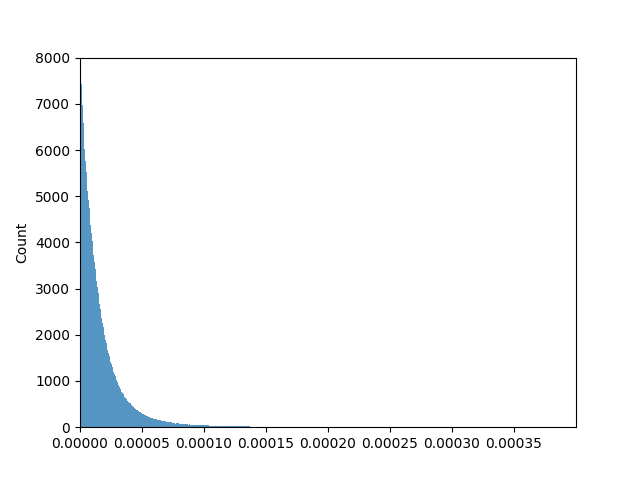}
    \caption{layer 25}
  \end{subfigure}
  \hfill
  \begin{subfigure}{0.15\textwidth}
    \includegraphics[width=\textwidth]{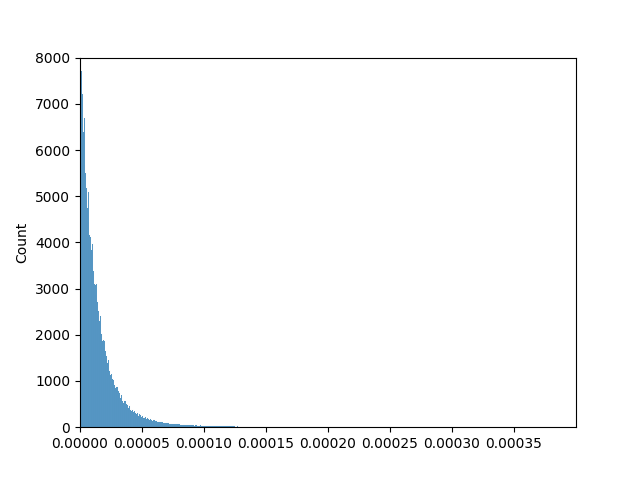}
    \caption{layer 26}
  \end{subfigure}
  \hfill
  \begin{subfigure}{0.15\textwidth}
    \includegraphics[width=\textwidth]{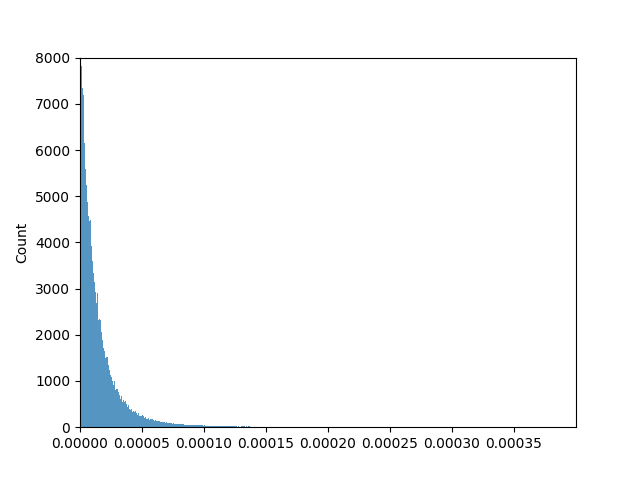}
    \caption{layer 27}
  \end{subfigure}
  \hfill
  \begin{subfigure}{0.15\textwidth}
    \includegraphics[width=\textwidth]{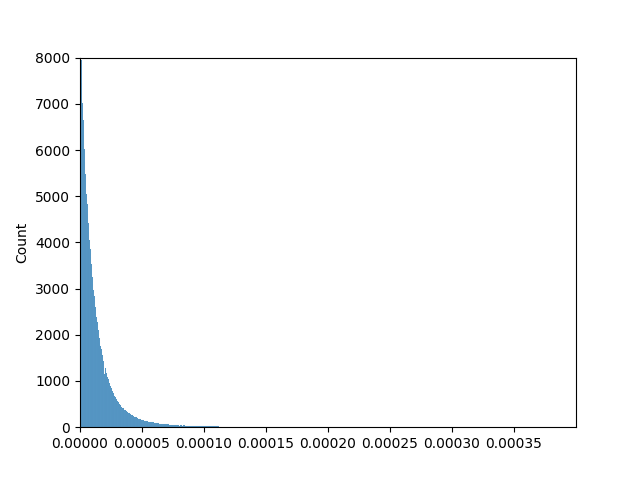}
    \caption{layer 28}
  \end{subfigure}
  \hfill
  \begin{subfigure}{0.15\textwidth}
    \includegraphics[width=\textwidth]{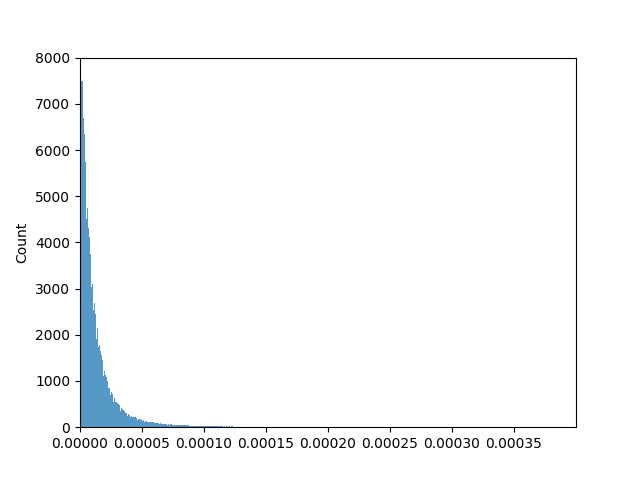}
    \caption{layer 29}
  \end{subfigure}
  \hfill
  \begin{subfigure}{0.15\textwidth}
    \includegraphics[width=\textwidth]{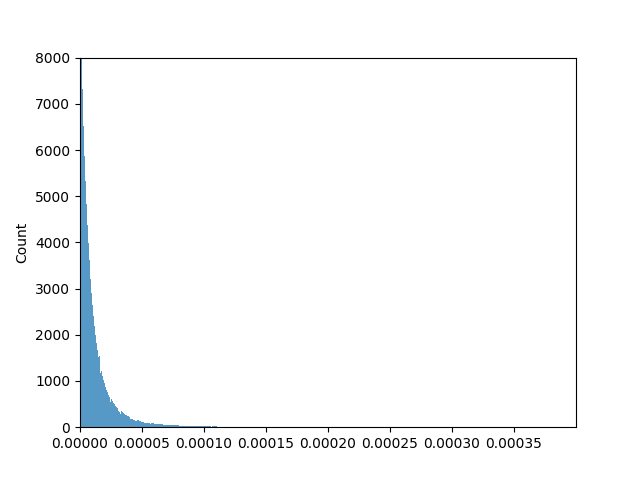}
    \caption{layer 30}
  \end{subfigure}
  \hfill
  \begin{subfigure}{0.15\textwidth}
    \includegraphics[width=\textwidth]{images/boolq/30.png}
    \caption{layer 31}
  \end{subfigure}
  \hfill
  \begin{subfigure}{0.15\textwidth}
    \includegraphics[width=\textwidth]{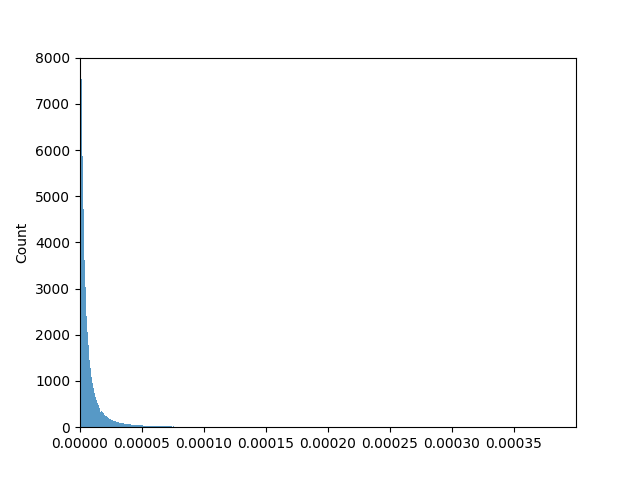}
    \caption{layer 32}
  \end{subfigure}
  \hfill

  \caption{Activated Parameter Statistics. The x-axis represents the value of $A(w_i)$, and the y-axis represents the quantity. For convenience of statistics, we have performed statistical processing, so the numerical value on the y-axis is an estimate of one-thousandth of the actual value. }
\end{figure}
\newpage
\subsection{Statistics for Finding 1 on HumanEval}

\begin{figure}[htbp]
  \begin{subfigure}{0.15\textwidth}
    \includegraphics[width=\textwidth]{images/he/0.png}
    \caption{layer 1}
  \end{subfigure}
  \hfill
  \begin{subfigure}{0.15\textwidth}
    \includegraphics[width=\textwidth]{images/he/1.png}
    \caption{layer 2}
  \end{subfigure}
  \hfill
  \begin{subfigure}{0.15\textwidth}
    \includegraphics[width=\textwidth]{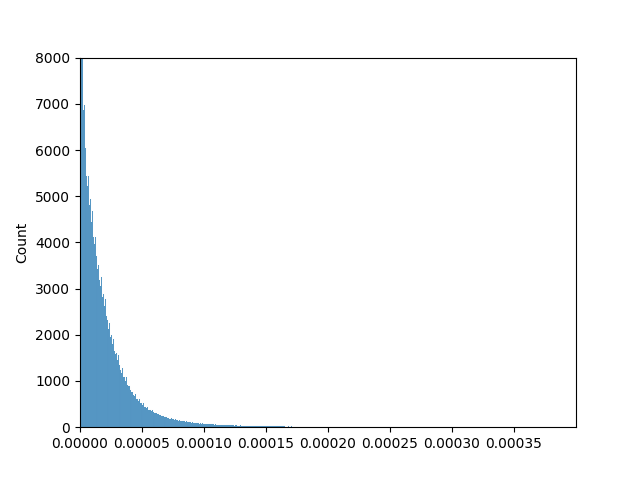}
    \caption{layer 3}
  \end{subfigure}
  \hfill
  \begin{subfigure}{0.15\textwidth}
    \includegraphics[width=\textwidth]{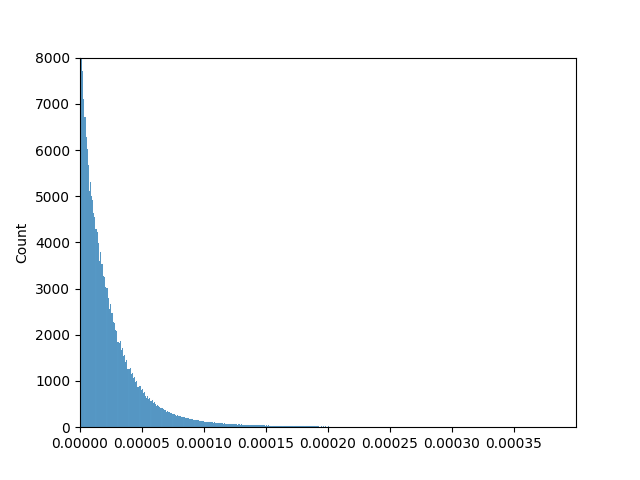}
    \caption{layer 4}
  \end{subfigure}
  \hfill
  \begin{subfigure}{0.15\textwidth}
    \includegraphics[width=\textwidth]{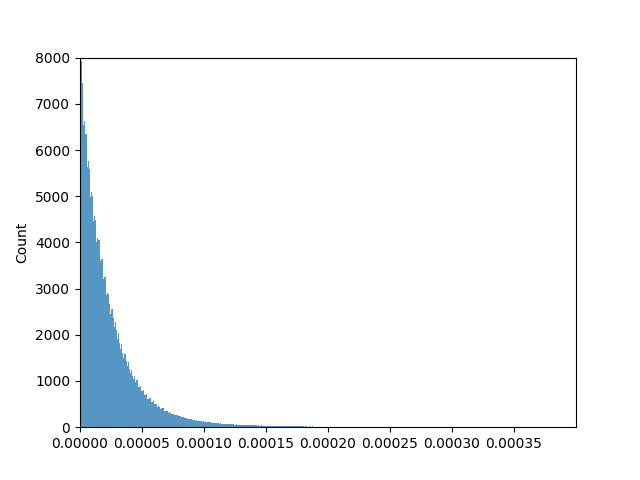}
    \caption{layer 5}
  \end{subfigure}
  \hfill
  \begin{subfigure}{0.15\textwidth}
    \includegraphics[width=\textwidth]{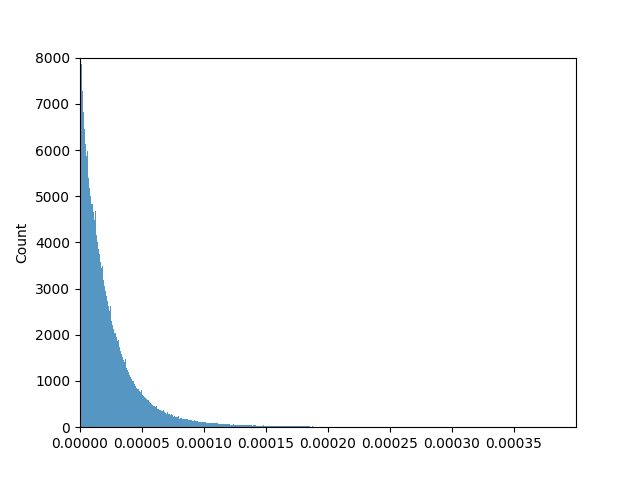}
    \caption{layer 6}
  \end{subfigure}
  \hfill
  \begin{subfigure}{0.15\textwidth}
    \includegraphics[width=\textwidth]{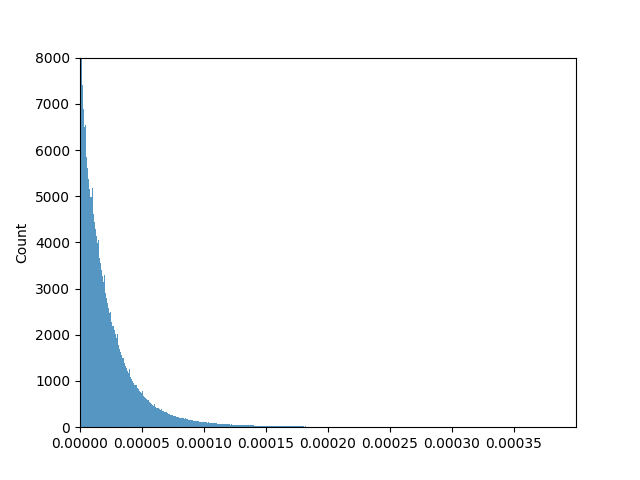}
    \caption{layer 7}
  \end{subfigure}
  \hfill
  \begin{subfigure}{0.15\textwidth}
    \includegraphics[width=\textwidth]{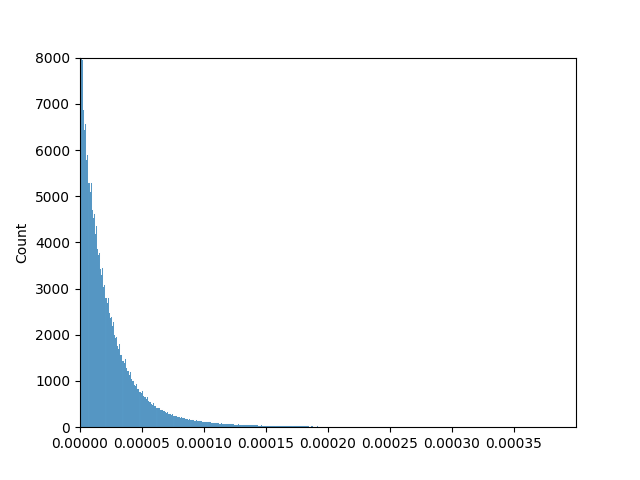}
    \caption{layer 8}
  \end{subfigure}
  \hfill
  \begin{subfigure}{0.15\textwidth}
    \includegraphics[width=\textwidth]{images/he/8.png}
    \caption{layer 9}
  \end{subfigure}
  \hfill
  \begin{subfigure}{0.15\textwidth}
    \includegraphics[width=\textwidth]{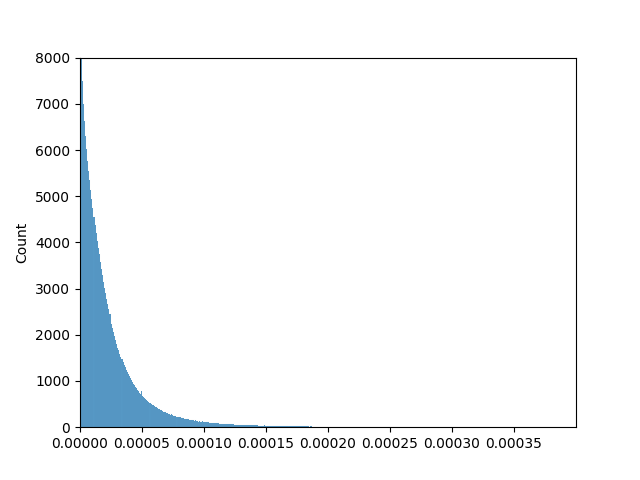}
    \caption{layer 10}
  \end{subfigure}
  \hfill
  \begin{subfigure}{0.15\textwidth}
    \includegraphics[width=\textwidth]{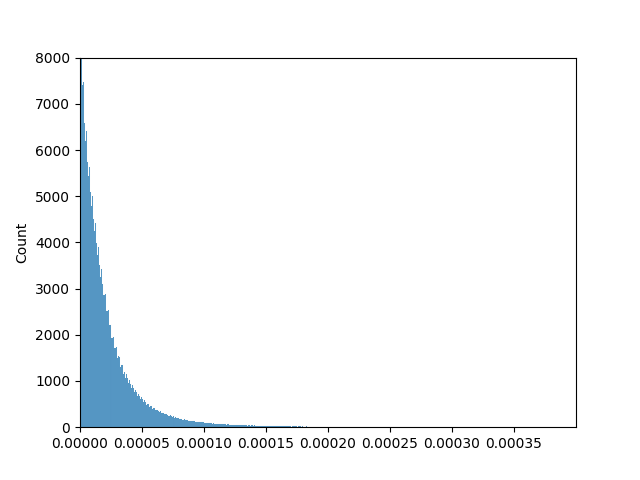}
    \caption{layer 11}
  \end{subfigure}
  \hfill
  \begin{subfigure}{0.15\textwidth}
    \includegraphics[width=\textwidth]{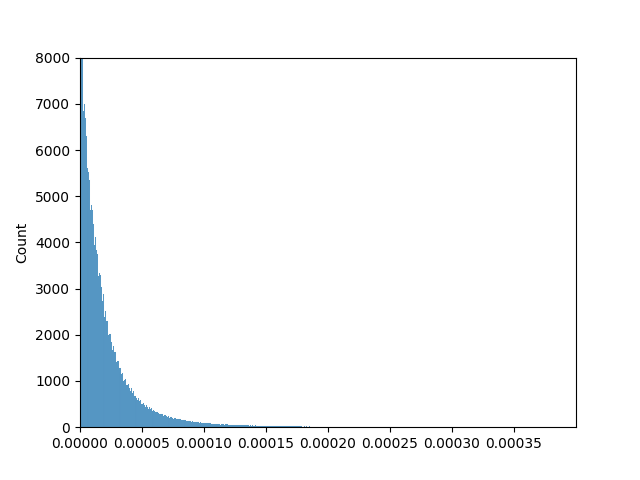}
    \caption{layer 12}
  \end{subfigure}
  \hfill
  \begin{subfigure}{0.15\textwidth}
    \includegraphics[width=\textwidth]{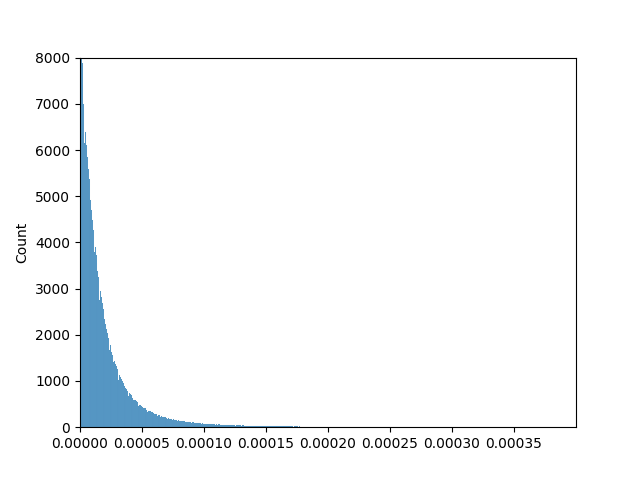}
    \caption{layer 13}
  \end{subfigure}
  \hfill
  \begin{subfigure}{0.15\textwidth}
    \includegraphics[width=\textwidth]{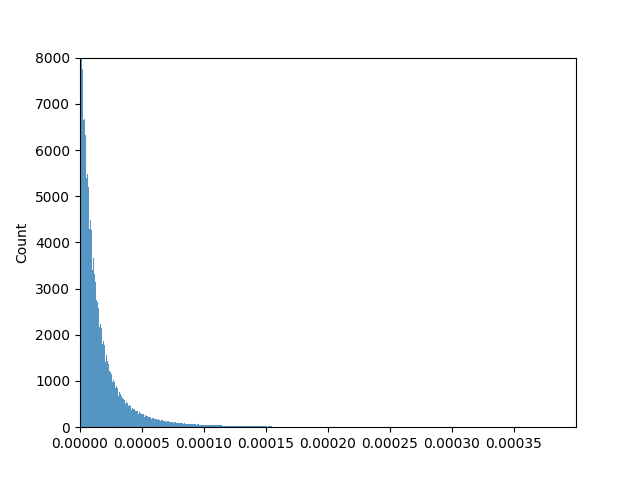}
    \caption{layer 14}
  \end{subfigure}
  \hfill
  \begin{subfigure}{0.15\textwidth}
    \includegraphics[width=\textwidth]{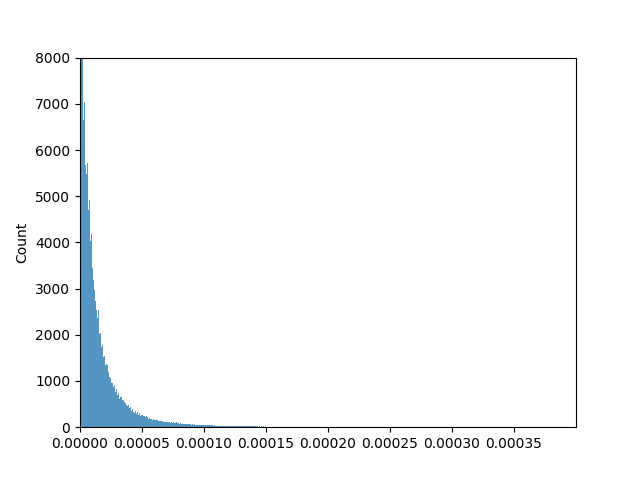}
    \caption{layer 15}
  \end{subfigure}
  \hfill
  \begin{subfigure}{0.15\textwidth}
    \includegraphics[width=\textwidth]{images/he/15.png}
    \caption{layer 16}
  \end{subfigure}
  \hfill
  \begin{subfigure}{0.15\textwidth}
    \includegraphics[width=\textwidth]{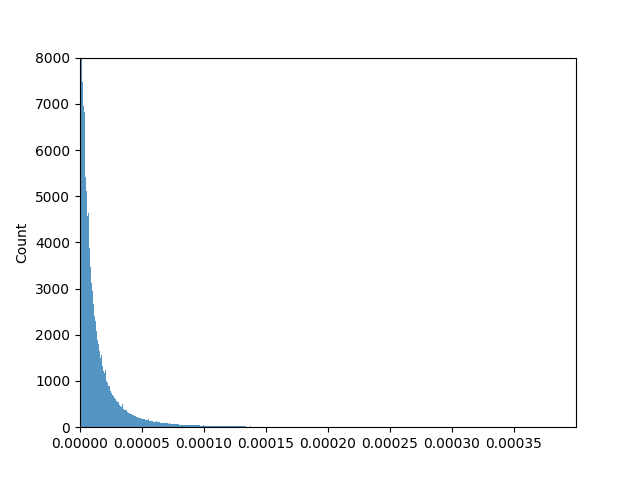}
    \caption{layer 17}
  \end{subfigure}
  \hfill
  \begin{subfigure}{0.15\textwidth}
    \includegraphics[width=\textwidth]{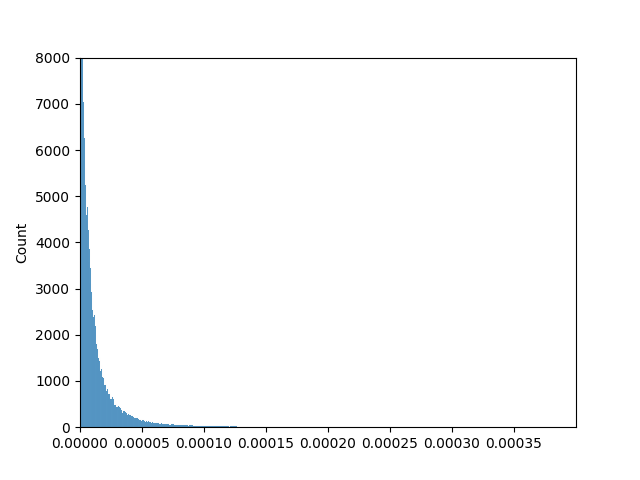}
    \caption{layer 18}
  \end{subfigure}
  \hfill
  \begin{subfigure}{0.15\textwidth}
    \includegraphics[width=\textwidth]{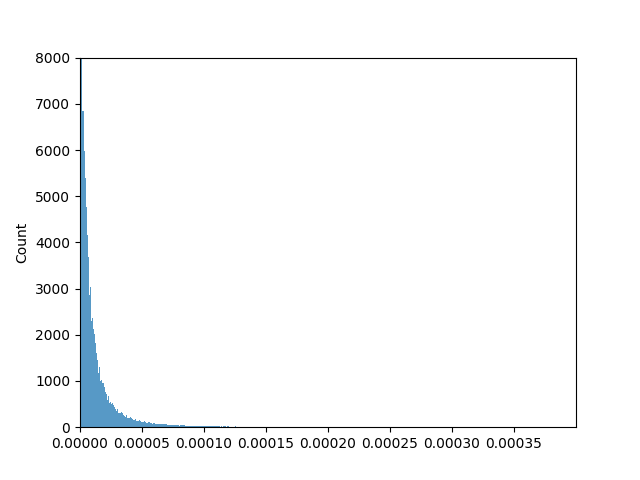}
    \caption{layer 19}
  \end{subfigure}
  \hfill
  \begin{subfigure}{0.15\textwidth}
    \includegraphics[width=\textwidth]{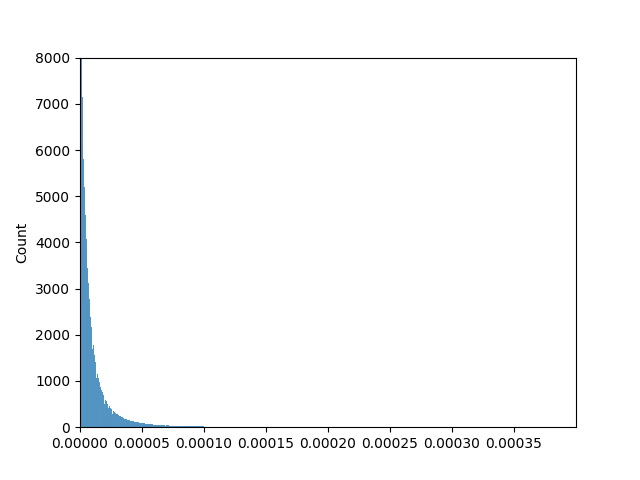}
    \caption{layer 20}
  \end{subfigure}
  \hfill
  \begin{subfigure}{0.15\textwidth}
    \includegraphics[width=\textwidth]{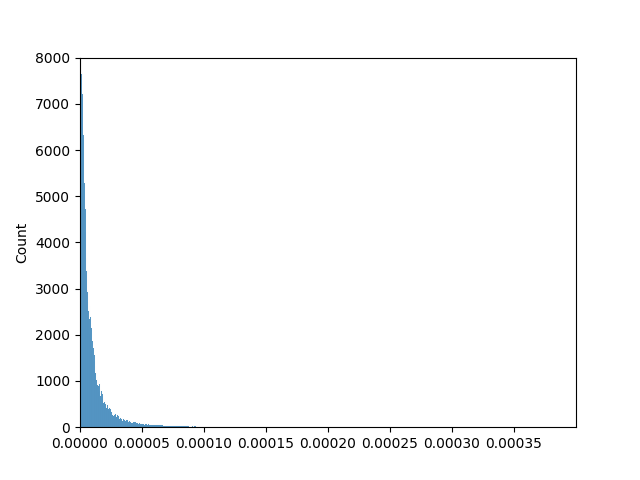}
    \caption{layer 21}
  \end{subfigure}
  \hfill
  \begin{subfigure}{0.15\textwidth}
    \includegraphics[width=\textwidth]{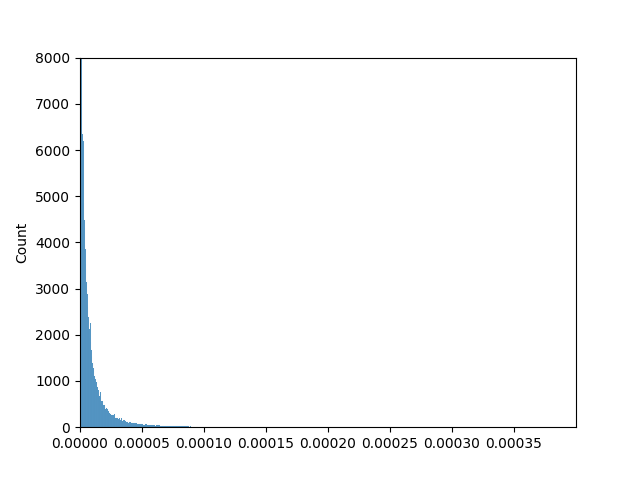}
    \caption{layer 22}
  \end{subfigure}
  \hfill
  \begin{subfigure}{0.15\textwidth}
    \includegraphics[width=\textwidth]{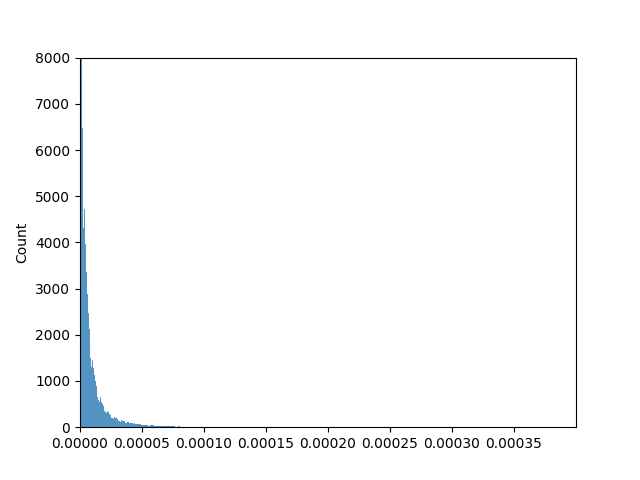}
    \caption{layer 23}
  \end{subfigure}
  \hfill
  \begin{subfigure}{0.15\textwidth}
    \includegraphics[width=\textwidth]{images/he/23.png}
    \caption{layer 24}
  \end{subfigure}
  \hfill
  \begin{subfigure}{0.15\textwidth}
    \includegraphics[width=\textwidth]{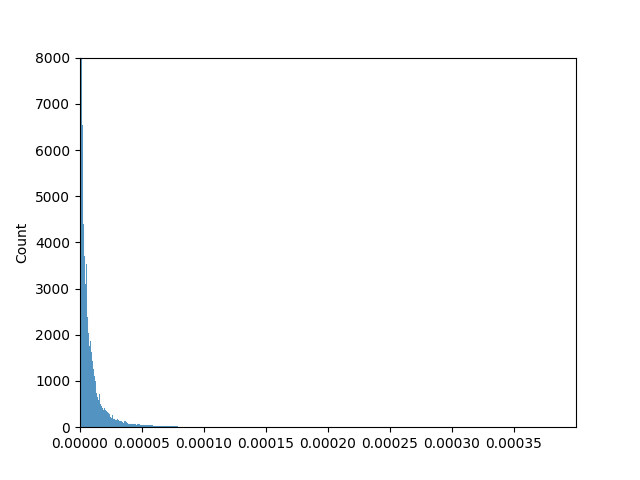}
    \caption{layer 25}
  \end{subfigure}
  \hfill
  \begin{subfigure}{0.15\textwidth}
    \includegraphics[width=\textwidth]{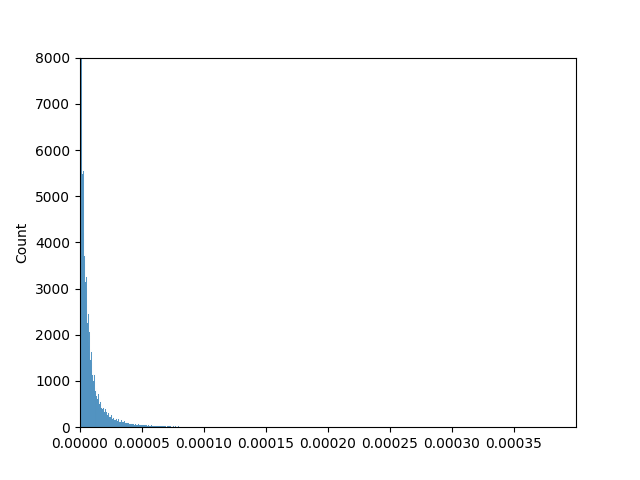}
    \caption{layer 26}
  \end{subfigure}
  \hfill
  \begin{subfigure}{0.15\textwidth}
    \includegraphics[width=\textwidth]{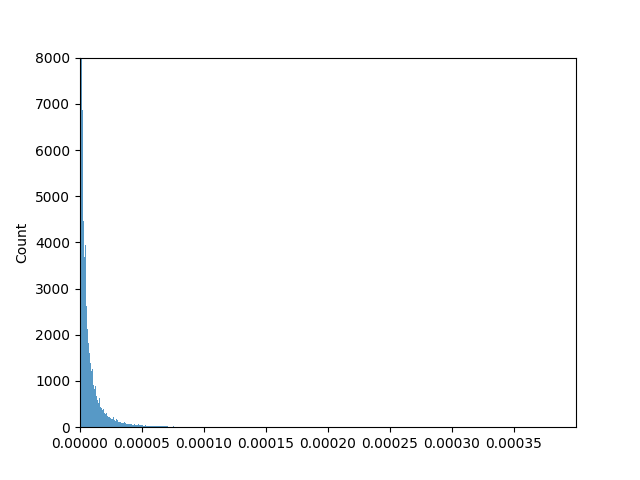}
    \caption{layer 27}
  \end{subfigure}
  \hfill
  \begin{subfigure}{0.15\textwidth}
    \includegraphics[width=\textwidth]{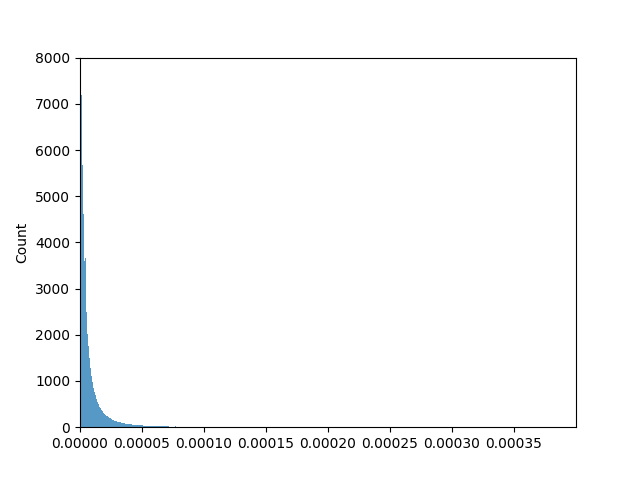}
    \caption{layer 28}
  \end{subfigure}
  \hfill
  \begin{subfigure}{0.15\textwidth}
    \includegraphics[width=\textwidth]{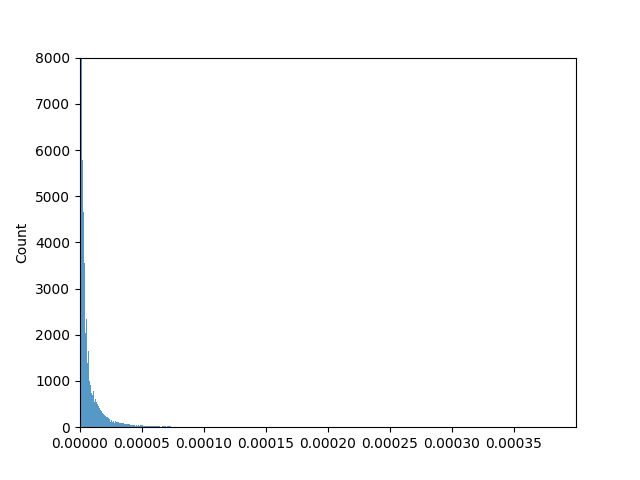}
    \caption{layer 29}
  \end{subfigure}
  \hfill
  \begin{subfigure}{0.15\textwidth}
    \includegraphics[width=\textwidth]{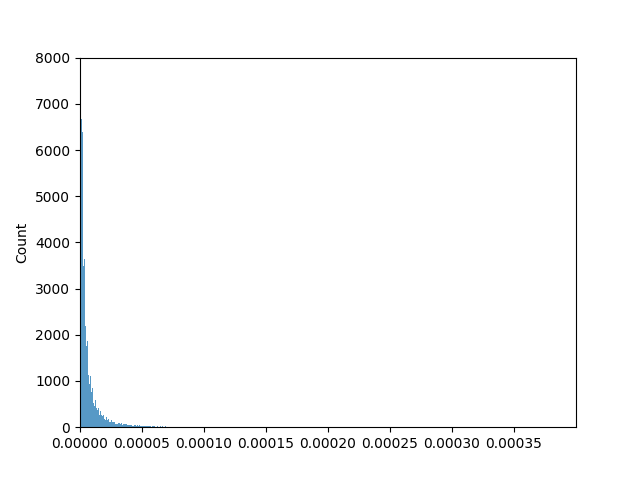}
    \caption{layer 30}
  \end{subfigure}
  \hfill
  \begin{subfigure}{0.15\textwidth}
    \includegraphics[width=\textwidth]{images/he/30.png}
    \caption{layer 31}
  \end{subfigure}
  \hfill
  \begin{subfigure}{0.15\textwidth}
    \includegraphics[width=\textwidth]{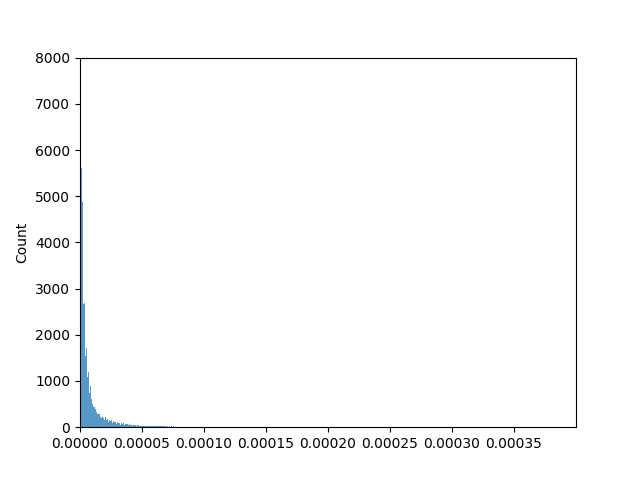}
    \caption{layer 32}
  \end{subfigure}
  \hfill

  \caption{Activated Parameter Statistics. The x-axis represents the value of $A(w_i)$, and the y-axis represents the quantity. For convenience of statistics, we have performed statistical processing, so the numerical value on the y-axis is an estimate of one-thousandth of the actual value. }
\end{figure}
\newpage
\subsection{Statistics for Finding 1 on MMLU}

\begin{figure}[htbp]
  \begin{subfigure}{0.15\textwidth}
    \includegraphics[width=\textwidth]{images/mmlu/0.png}
    \caption{layer 1}
  \end{subfigure}
  \hfill
  \begin{subfigure}{0.15\textwidth}
    \includegraphics[width=\textwidth]{images/mmlu/1.png}
    \caption{layer 2}
  \end{subfigure}
  \hfill
  \begin{subfigure}{0.15\textwidth}
    \includegraphics[width=\textwidth]{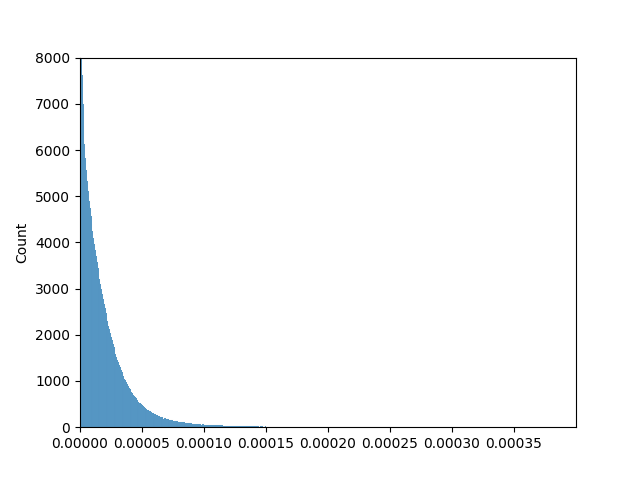}
    \caption{layer 3}
  \end{subfigure}
  \hfill
  \begin{subfigure}{0.15\textwidth}
    \includegraphics[width=\textwidth]{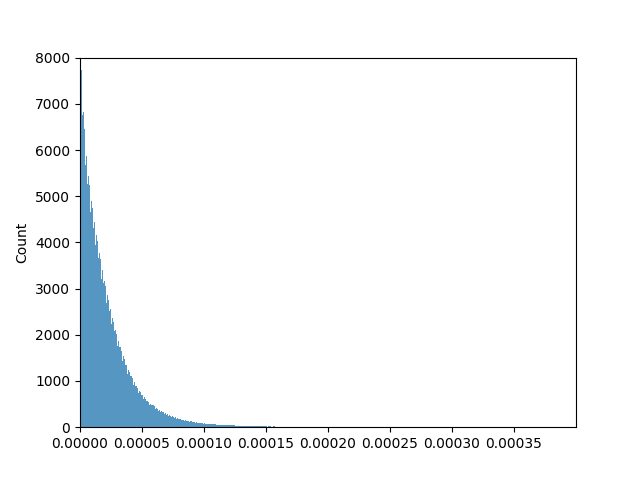}
    \caption{layer 4}
  \end{subfigure}
  \hfill
  \begin{subfigure}{0.15\textwidth}
    \includegraphics[width=\textwidth]{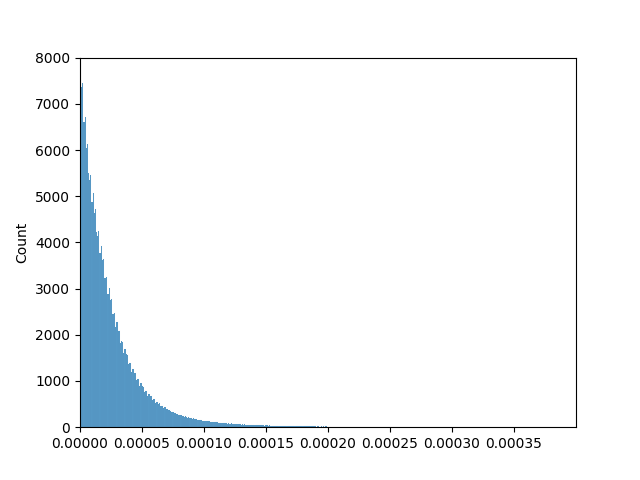}
    \caption{layer 5}
  \end{subfigure}
  \hfill
  \begin{subfigure}{0.15\textwidth}
    \includegraphics[width=\textwidth]{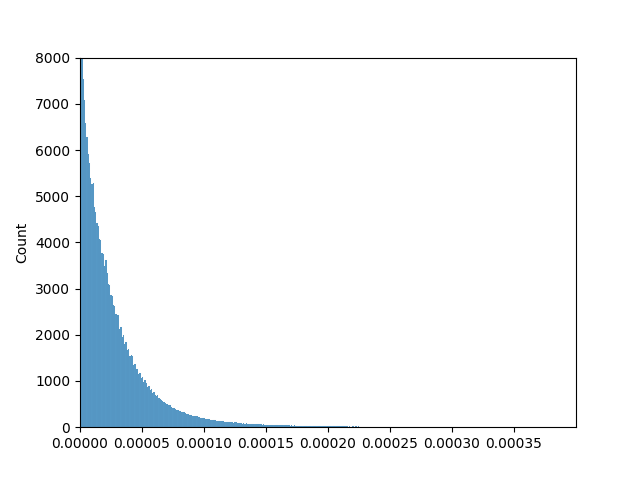}
    \caption{layer 6}
  \end{subfigure}
  \hfill
  \begin{subfigure}{0.15\textwidth}
    \includegraphics[width=\textwidth]{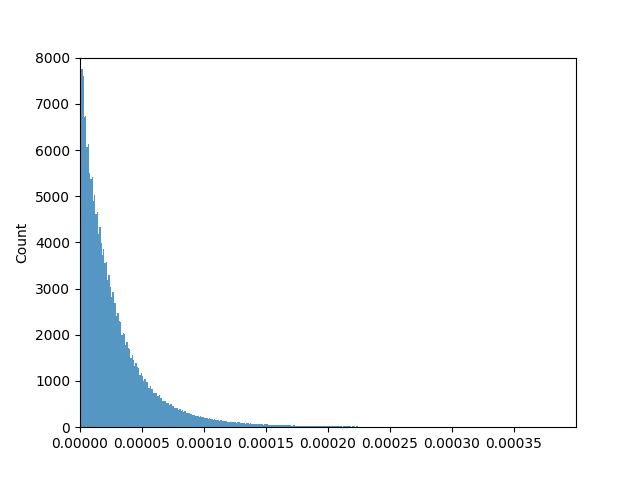}
    \caption{layer 7}
  \end{subfigure}
  \hfill
  \begin{subfigure}{0.15\textwidth}
    \includegraphics[width=\textwidth]{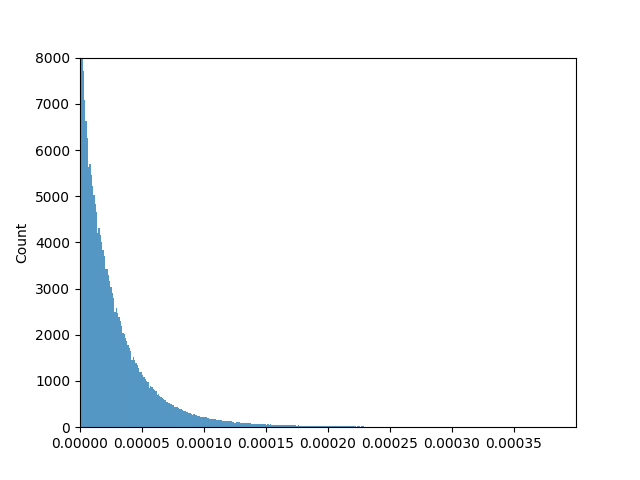}
    \caption{layer 8}
  \end{subfigure}
  \hfill
  \begin{subfigure}{0.15\textwidth}
    \includegraphics[width=\textwidth]{images/mmlu/8.png}
    \caption{layer 9}
  \end{subfigure}
  \hfill
  \begin{subfigure}{0.15\textwidth}
    \includegraphics[width=\textwidth]{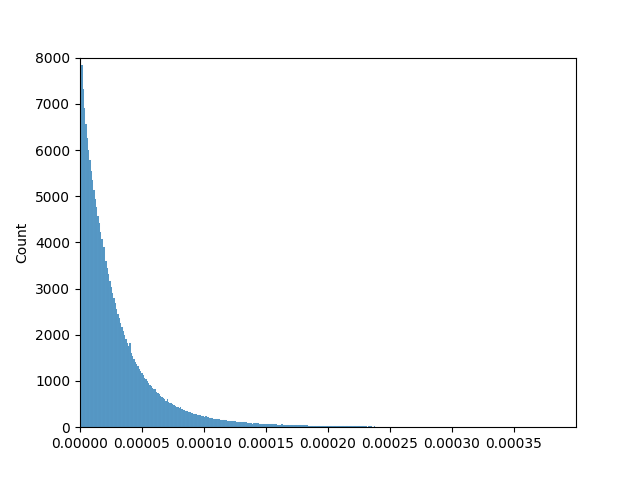}
    \caption{layer 10}
  \end{subfigure}
  \hfill
  \begin{subfigure}{0.15\textwidth}
    \includegraphics[width=\textwidth]{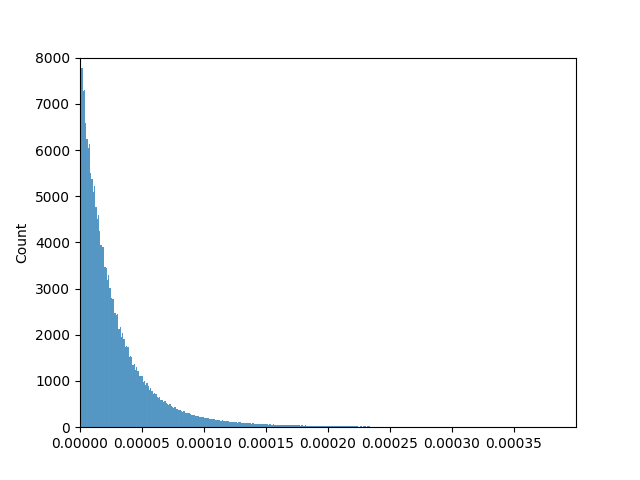}
    \caption{layer 11}
  \end{subfigure}
  \hfill
  \begin{subfigure}{0.15\textwidth}
    \includegraphics[width=\textwidth]{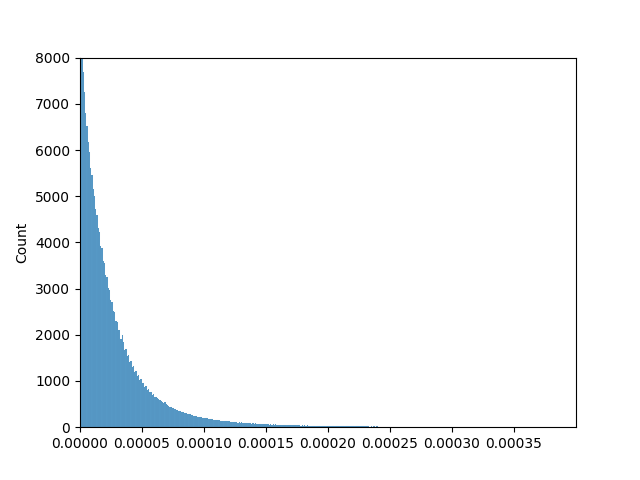}
    \caption{layer 12}
  \end{subfigure}
  \hfill
  \begin{subfigure}{0.15\textwidth}
    \includegraphics[width=\textwidth]{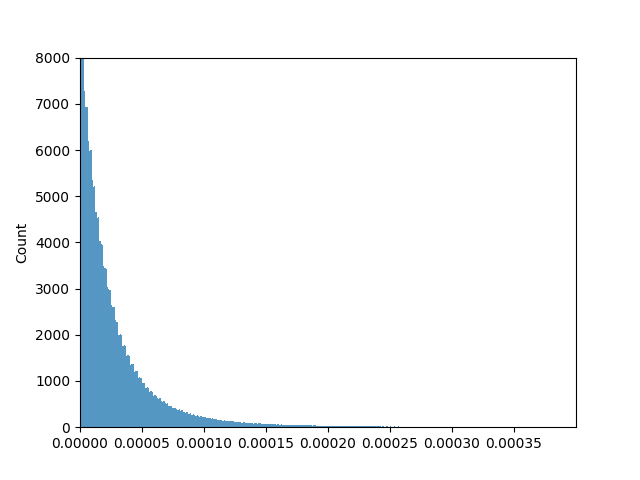}
    \caption{layer 13}
  \end{subfigure}
  \hfill
  \begin{subfigure}{0.15\textwidth}
    \includegraphics[width=\textwidth]{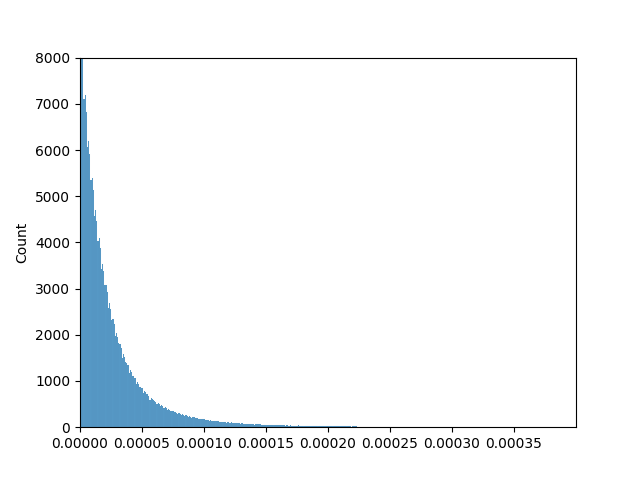}
    \caption{layer 14}
  \end{subfigure}
  \hfill
  \begin{subfigure}{0.15\textwidth}
    \includegraphics[width=\textwidth]{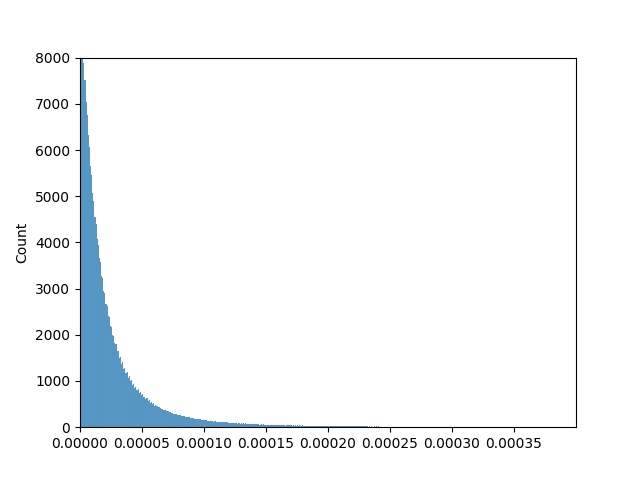}
    \caption{layer 15}
  \end{subfigure}
  \hfill
  \begin{subfigure}{0.15\textwidth}
    \includegraphics[width=\textwidth]{images/mmlu/15.png}
    \caption{layer 16}
  \end{subfigure}
  \hfill
  \begin{subfigure}{0.15\textwidth}
    \includegraphics[width=\textwidth]{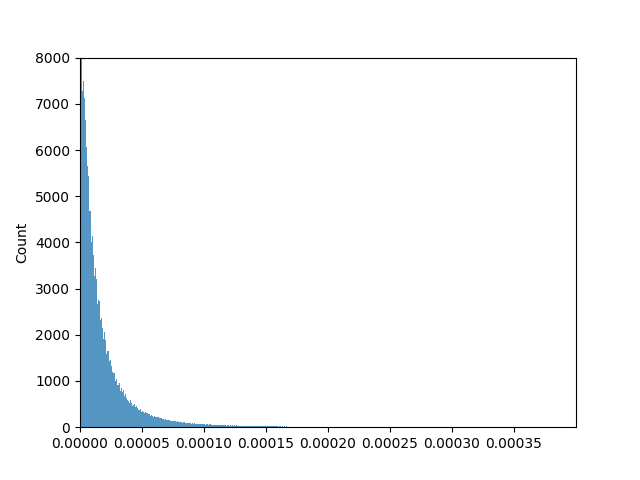}
    \caption{layer 17}
  \end{subfigure}
  \hfill
  \begin{subfigure}{0.15\textwidth}
    \includegraphics[width=\textwidth]{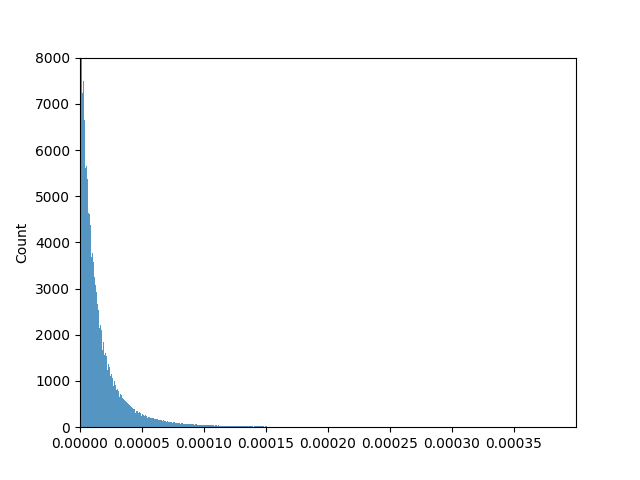}
    \caption{layer 18}
  \end{subfigure}
  \hfill
  \begin{subfigure}{0.15\textwidth}
    \includegraphics[width=\textwidth]{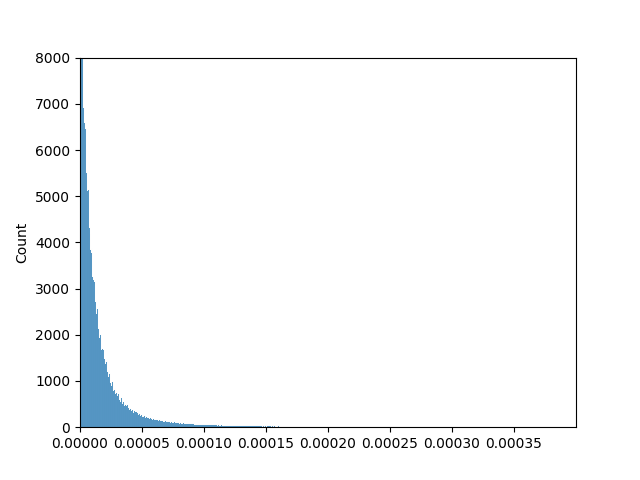}
    \caption{layer 19}
  \end{subfigure}
  \hfill
  \begin{subfigure}{0.15\textwidth}
    \includegraphics[width=\textwidth]{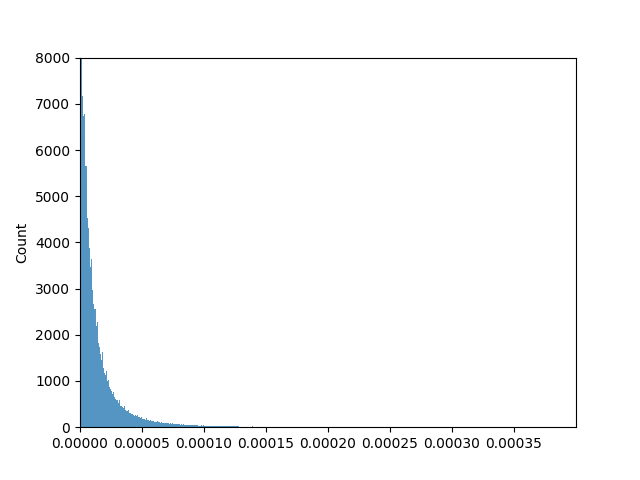}
    \caption{layer 20}
  \end{subfigure}
  \hfill
  \begin{subfigure}{0.15\textwidth}
    \includegraphics[width=\textwidth]{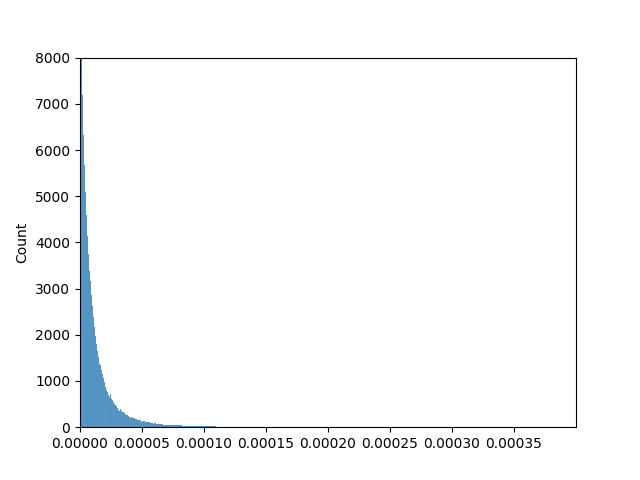}
    \caption{layer 21}
  \end{subfigure}
  \hfill
  \begin{subfigure}{0.15\textwidth}
    \includegraphics[width=\textwidth]{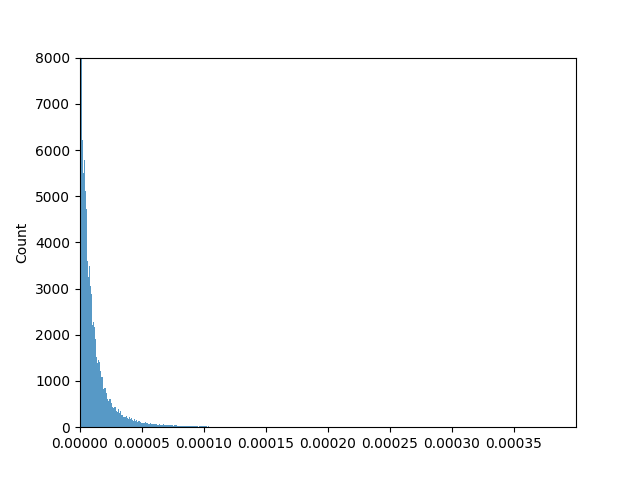}
    \caption{layer 22}
  \end{subfigure}
  \hfill
  \begin{subfigure}{0.15\textwidth}
    \includegraphics[width=\textwidth]{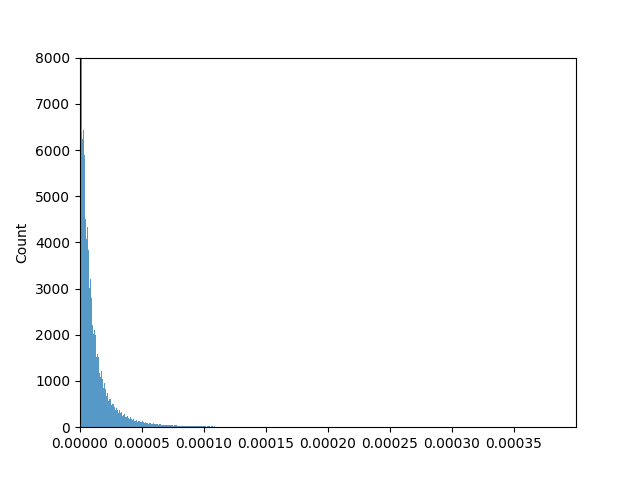}
    \caption{layer 23}
  \end{subfigure}
  \hfill
  \begin{subfigure}{0.15\textwidth}
    \includegraphics[width=\textwidth]{images/mmlu/23.png}
    \caption{layer 24}
  \end{subfigure}
  \hfill
  \begin{subfigure}{0.15\textwidth}
    \includegraphics[width=\textwidth]{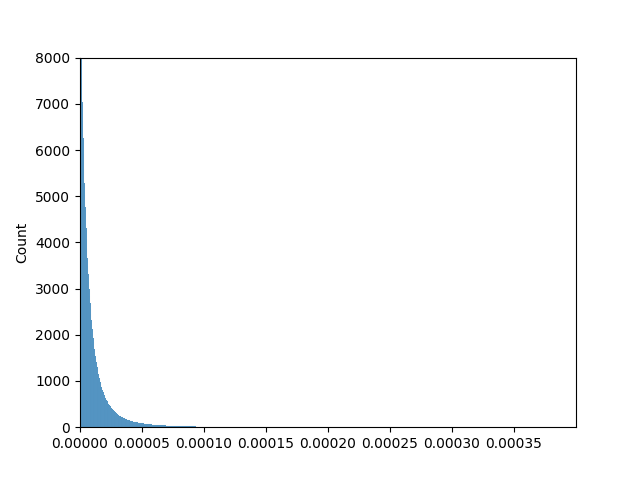}
    \caption{layer 25}
  \end{subfigure}
  \hfill
  \begin{subfigure}{0.15\textwidth}
    \includegraphics[width=\textwidth]{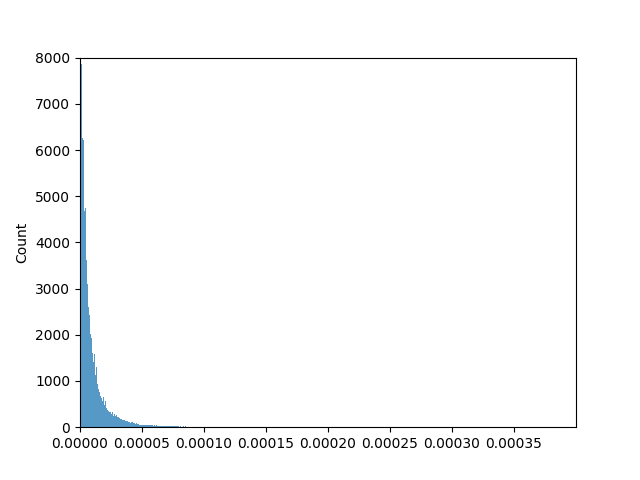}
    \caption{layer 26}
  \end{subfigure}
  \hfill
  \begin{subfigure}{0.15\textwidth}
    \includegraphics[width=\textwidth]{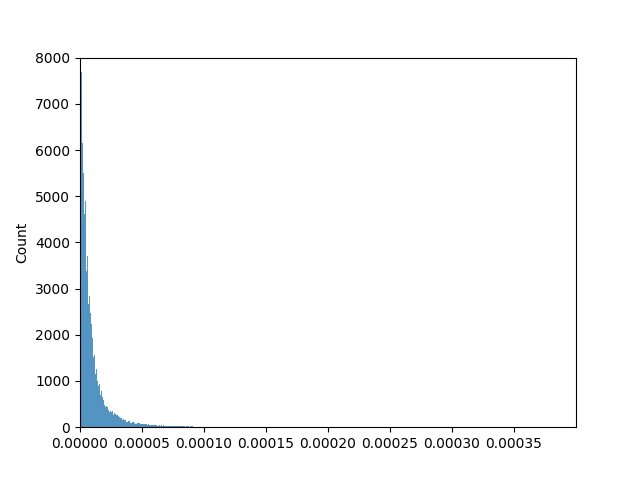}
    \caption{layer 27}
  \end{subfigure}
  \hfill
  \begin{subfigure}{0.15\textwidth}
    \includegraphics[width=\textwidth]{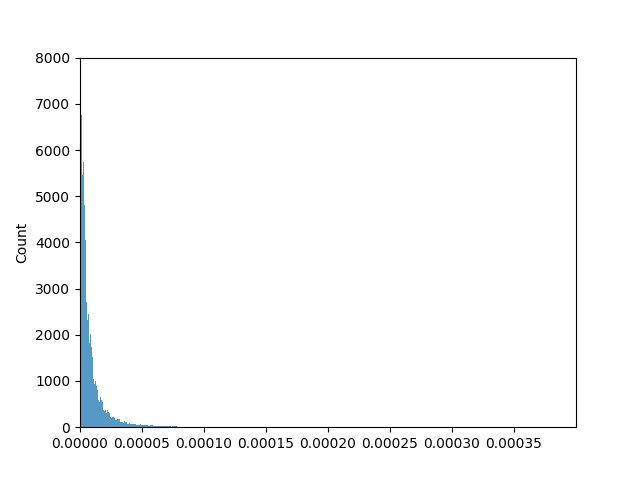}
    \caption{layer 28}
  \end{subfigure}
  \hfill
  \begin{subfigure}{0.15\textwidth}
    \includegraphics[width=\textwidth]{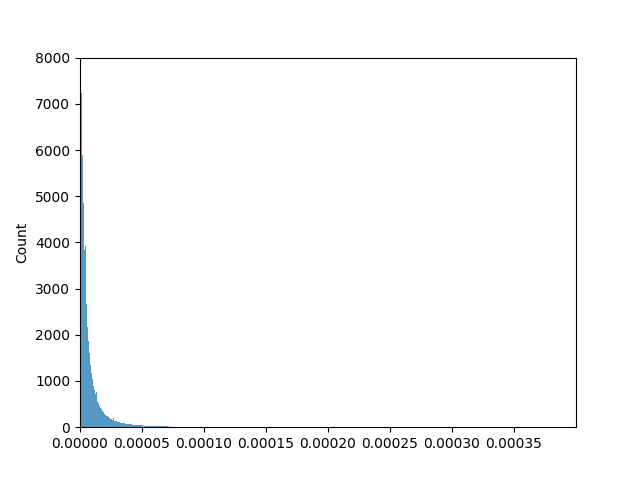}
    \caption{layer 29}
  \end{subfigure}
  \hfill
  \begin{subfigure}{0.15\textwidth}
    \includegraphics[width=\textwidth]{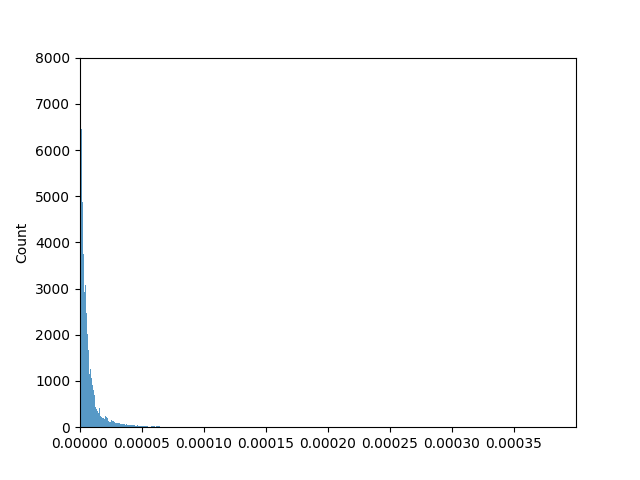}
    \caption{layer 30}
  \end{subfigure}
  \hfill
  \begin{subfigure}{0.15\textwidth}
    \includegraphics[width=\textwidth]{images/mmlu/30.png}
    \caption{layer 31}
  \end{subfigure}
  \hfill
  \begin{subfigure}{0.15\textwidth}
    \includegraphics[width=\textwidth]{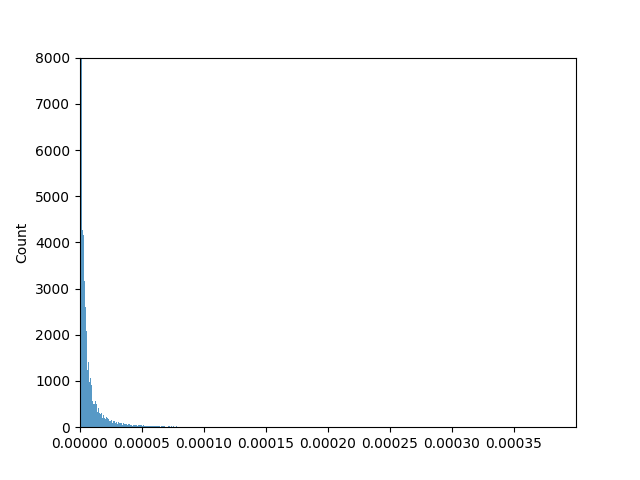}
    \caption{layer 32}
  \end{subfigure}
  \hfill

  \caption{Activated Parameter Statistics. The x-axis represents the value of $A(w_i)$, and the y-axis represents the quantity. For convenience of statistics, we have performed statistical processing, so the numerical value on the y-axis is an estimate of one-thousandth of the actual value. }
\end{figure}
\newpage
\subsection{Statistics for Finding 2}

\begin{figure}[htbp]
\begin{subfigure}{0.15\textwidth}
    \includegraphics[width=\textwidth]{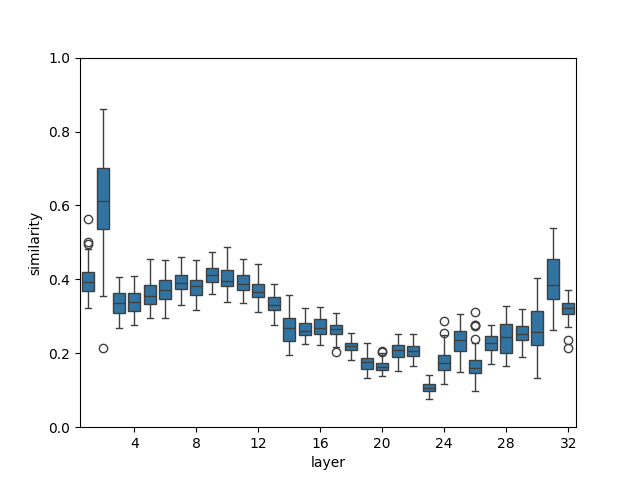}
\caption{humaneval-hellaswag}
  \end{subfigure}
  \hfill
\begin{subfigure}{0.15\textwidth}
    \includegraphics[width=\textwidth]{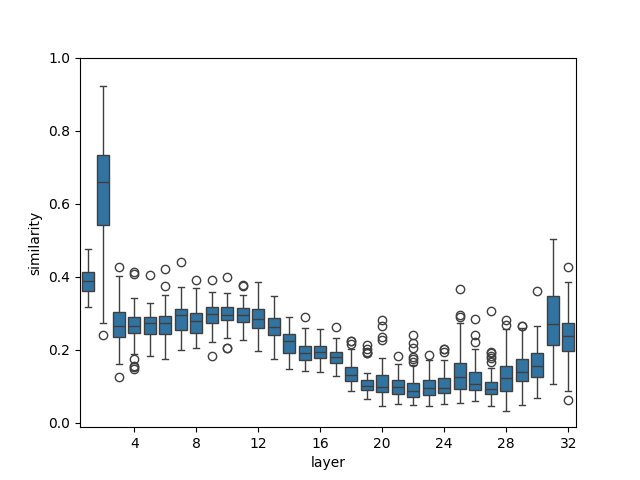}
\caption{wikitext2-humaneval}
  \end{subfigure}
  \hfill
\begin{subfigure}{0.15\textwidth}
    \includegraphics[width=\textwidth]{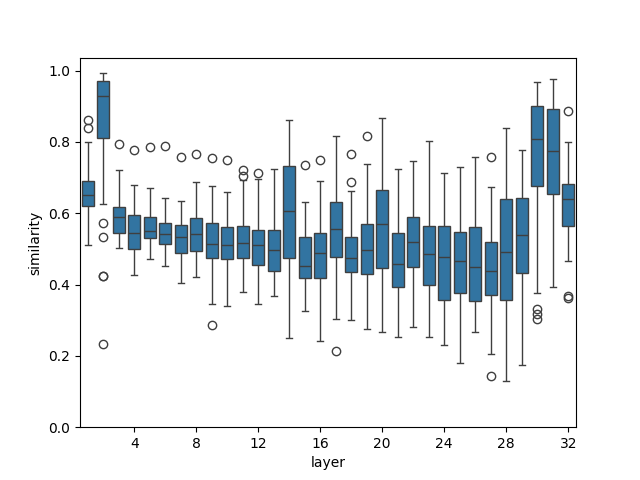}
\caption{piqa-piqa}
  \end{subfigure}
  \hfill
\begin{subfigure}{0.15\textwidth}
    \includegraphics[width=\textwidth]{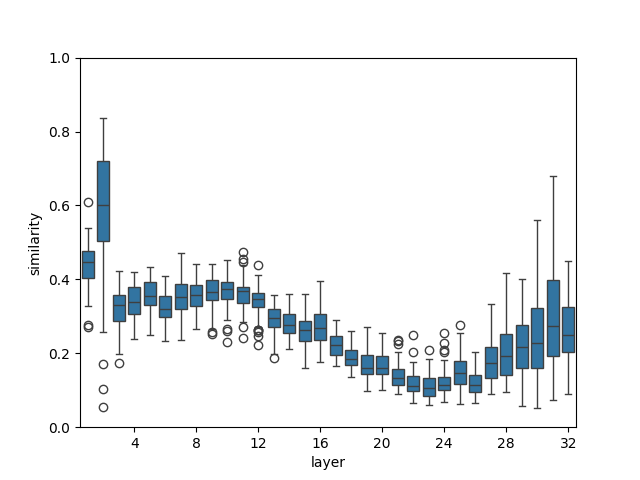}
\caption{c4-hellaswag}
  \end{subfigure}
  \hfill
\begin{subfigure}{0.15\textwidth}
    \includegraphics[width=\textwidth]{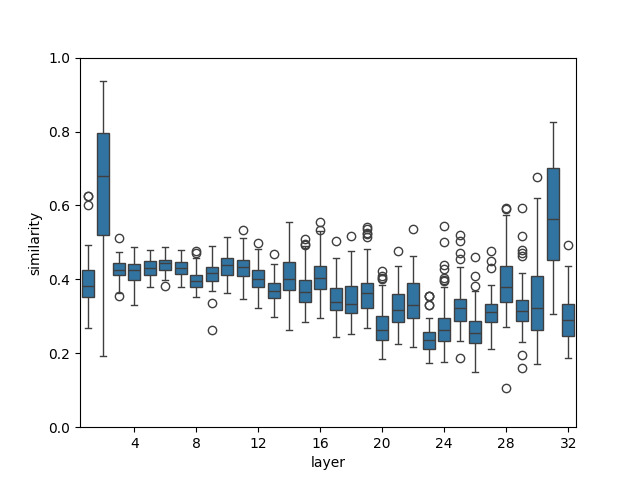}
\caption{piqa-hellaswag}
  \end{subfigure}
  \hfill
\begin{subfigure}{0.15\textwidth}
    \includegraphics[width=\textwidth]{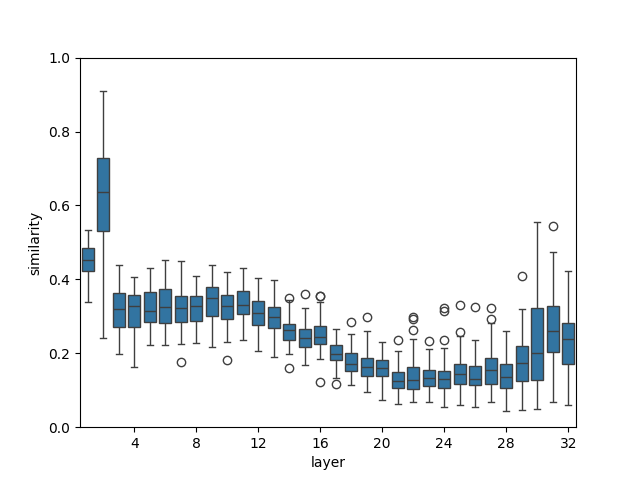}
\caption{wikitext2-boolq}
  \end{subfigure}
  \hfill
\begin{subfigure}{0.15\textwidth}
    \includegraphics[width=\textwidth]{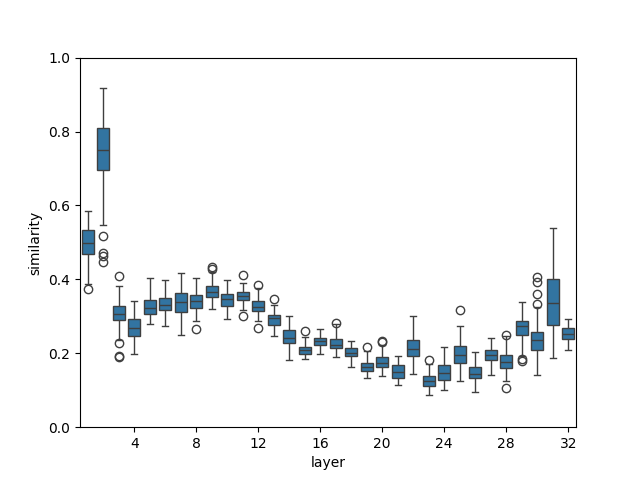}
\caption{humaneval-gsm8k}
  \end{subfigure}
  \hfill
\begin{subfigure}{0.15\textwidth}
    \includegraphics[width=\textwidth]{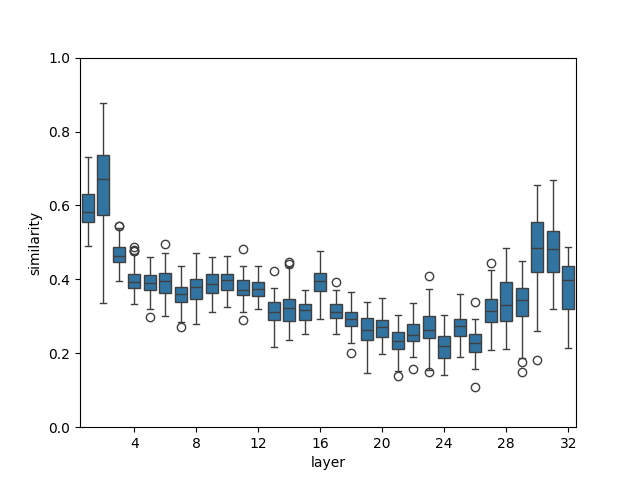}
\caption{siqa-boolq}
  \end{subfigure}
  \hfill
\begin{subfigure}{0.15\textwidth}
    \includegraphics[width=\textwidth]{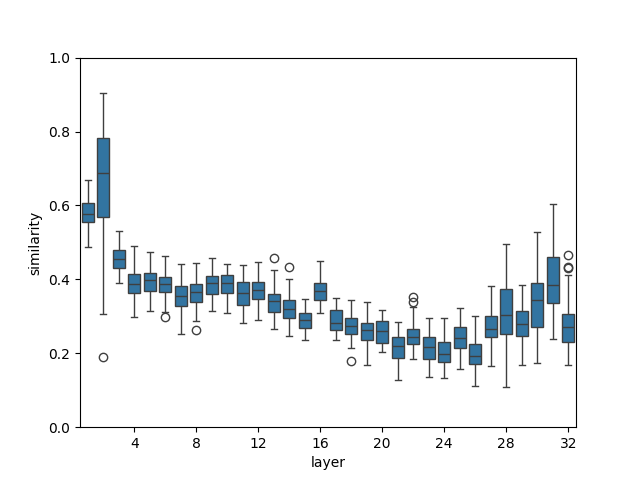}
\caption{boolq-mmlu}
  \end{subfigure}
  \hfill
\begin{subfigure}{0.15\textwidth}
    \includegraphics[width=\textwidth]{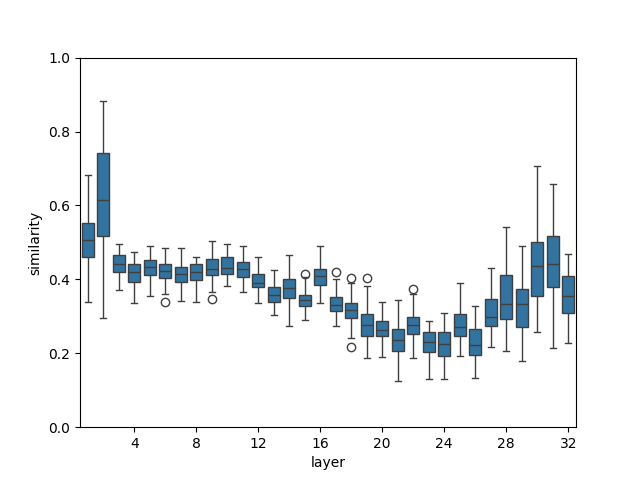}
\caption{boolq-hellaswag}
  \end{subfigure}
  \hfill
\begin{subfigure}{0.15\textwidth}
    \includegraphics[width=\textwidth]{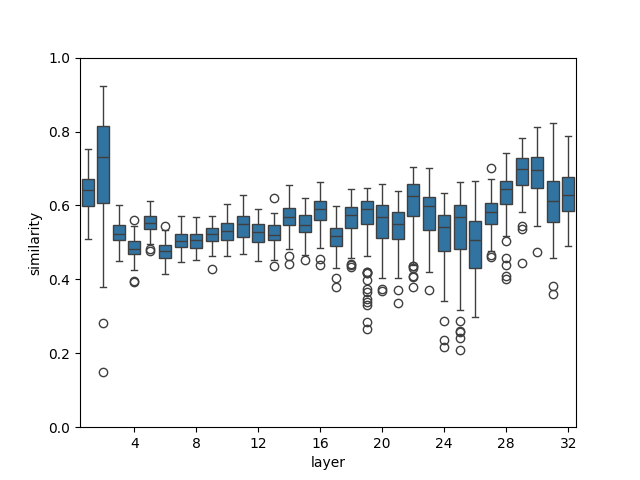}
\caption{siqa-mmlu}
  \end{subfigure}
  \hfill
\begin{subfigure}{0.15\textwidth}
    \includegraphics[width=\textwidth]{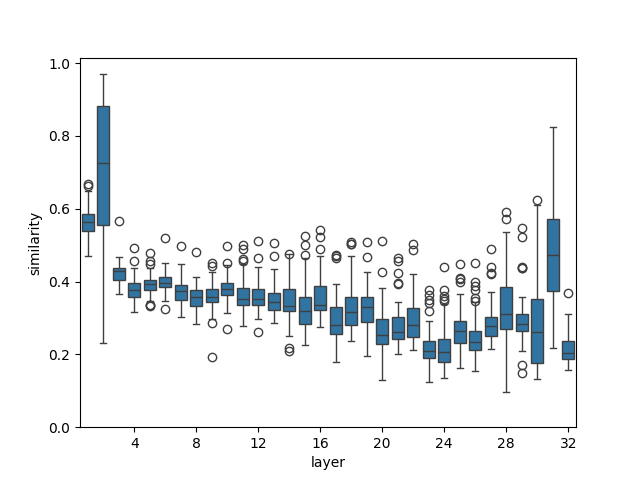}
\caption{piqa-mmlu}
  \end{subfigure}
  \hfill
\begin{subfigure}{0.15\textwidth}
    \includegraphics[width=\textwidth]{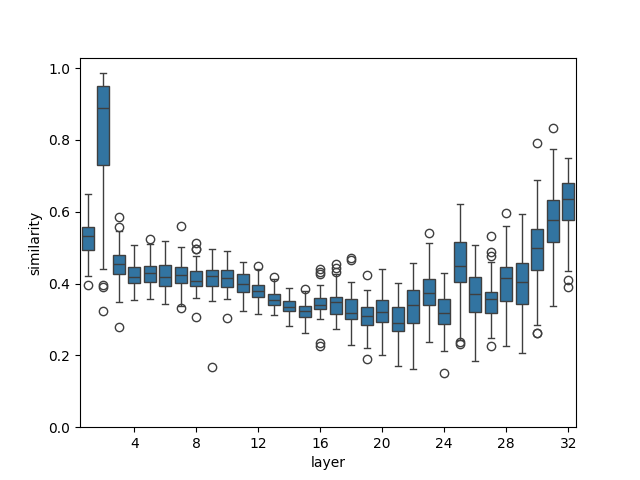}
\caption{gsm8k-gsm8k}
  \end{subfigure}
  \hfill
\begin{subfigure}{0.15\textwidth}
    \includegraphics[width=\textwidth]{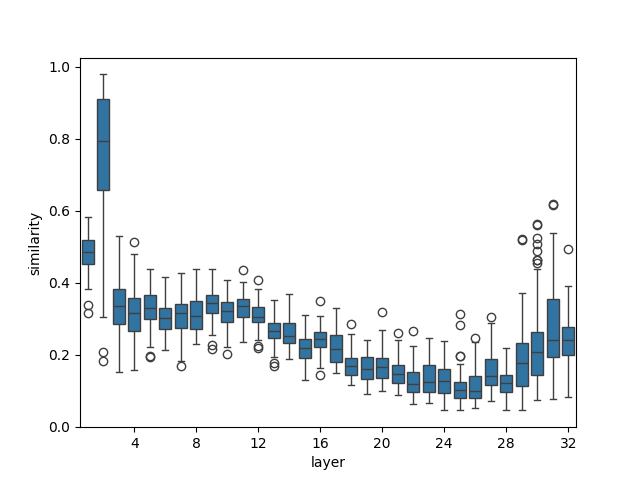}
\caption{c4-gsm8k}
  \end{subfigure}
  \hfill
\begin{subfigure}{0.15\textwidth}
    \includegraphics[width=\textwidth]{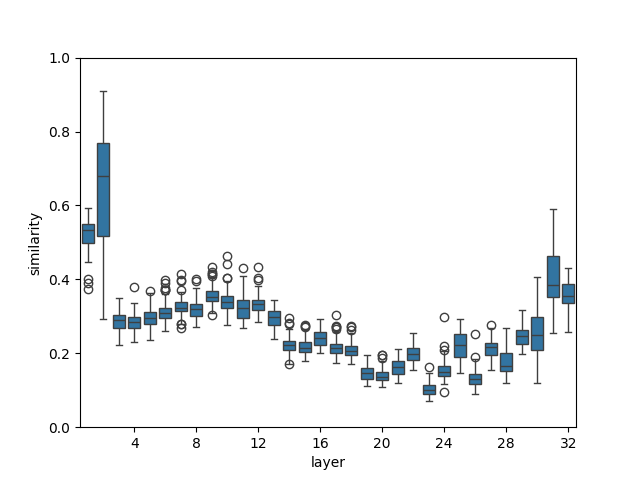}
\caption{humaneval-mmlu}
  \end{subfigure}
  \hfill
\begin{subfigure}{0.15\textwidth}
    \includegraphics[width=\textwidth]{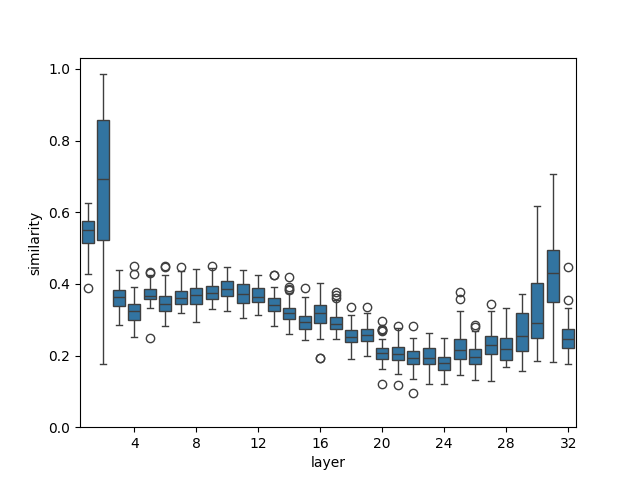}
\caption{gsm8k-mmlu}
  \end{subfigure}
  \hfill
\begin{subfigure}{0.15\textwidth}
    \includegraphics[width=\textwidth]{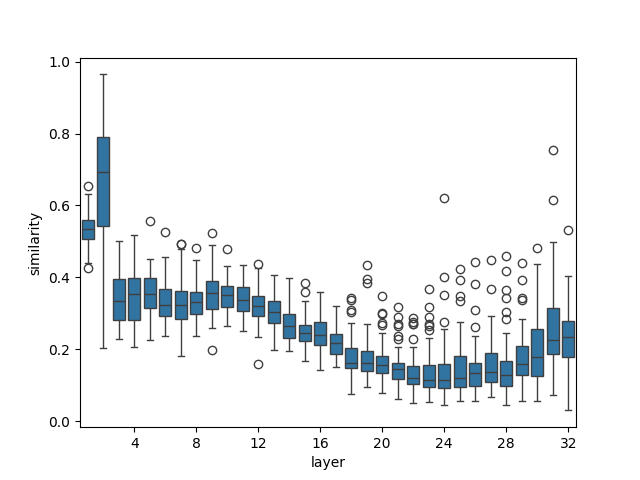}
\caption{wikitext2-wikitext2}
  \end{subfigure}
  \hfill
\begin{subfigure}{0.15\textwidth}
    \includegraphics[width=\textwidth]{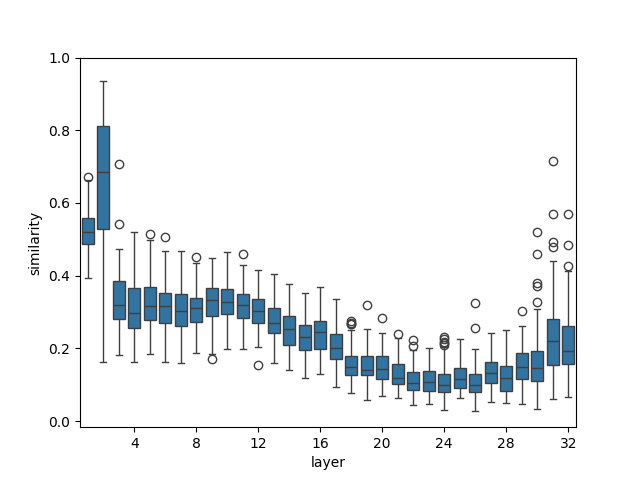}
\caption{c4-c4}
  \end{subfigure}
  \hfill
\begin{subfigure}{0.15\textwidth}
    \includegraphics[width=\textwidth]{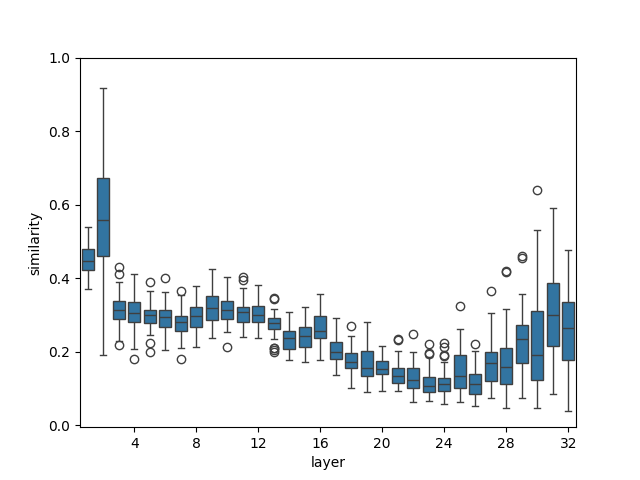}
\caption{wikitext2-siqa}
  \end{subfigure}
  \hfill
\begin{subfigure}{0.15\textwidth}
    \includegraphics[width=\textwidth]{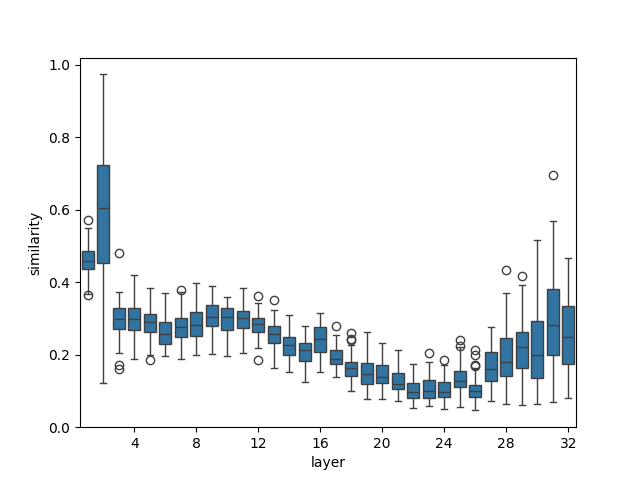}
\caption{c4-mmlu}
  \end{subfigure}
  \hfill
\begin{subfigure}{0.15\textwidth}
    \includegraphics[width=\textwidth]{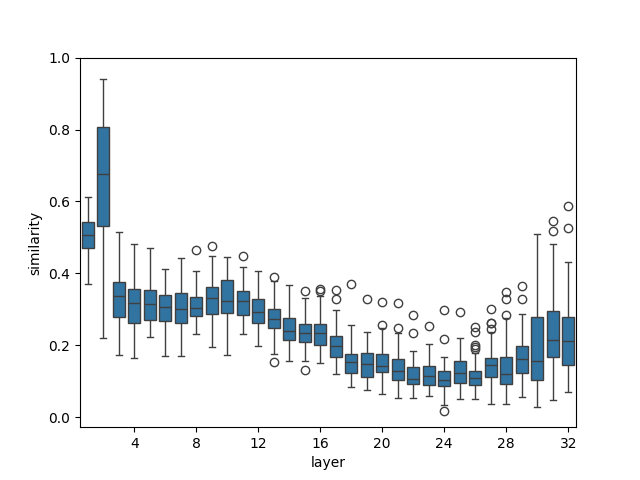}
\caption{wikitext2-c4}
  \end{subfigure}
  \hfill
\begin{subfigure}{0.15\textwidth}
    \includegraphics[width=\textwidth]{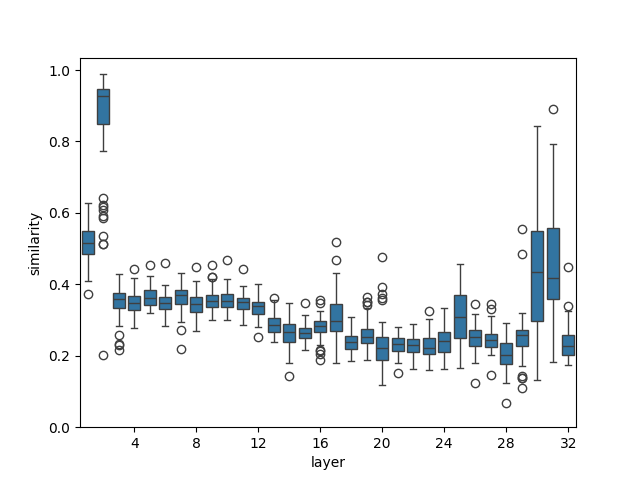}
\caption{piqa-gsm8k}
  \end{subfigure}
  \hfill
\begin{subfigure}{0.15\textwidth}
    \includegraphics[width=\textwidth]{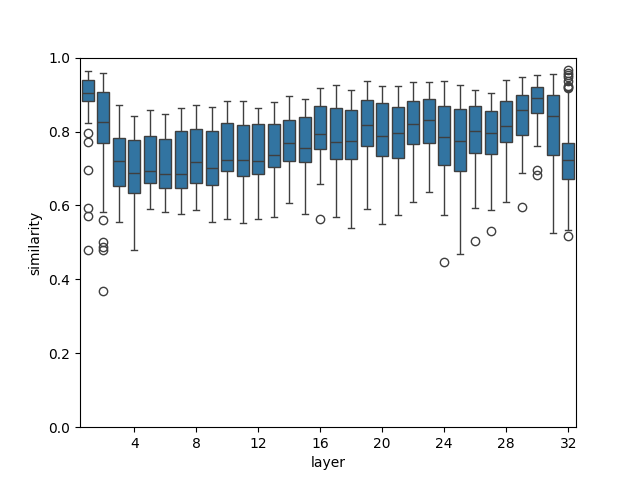}
\caption{hellaswag-hellaswag}
  \end{subfigure}
  \hfill
\begin{subfigure}{0.15\textwidth}
    \includegraphics[width=\textwidth]{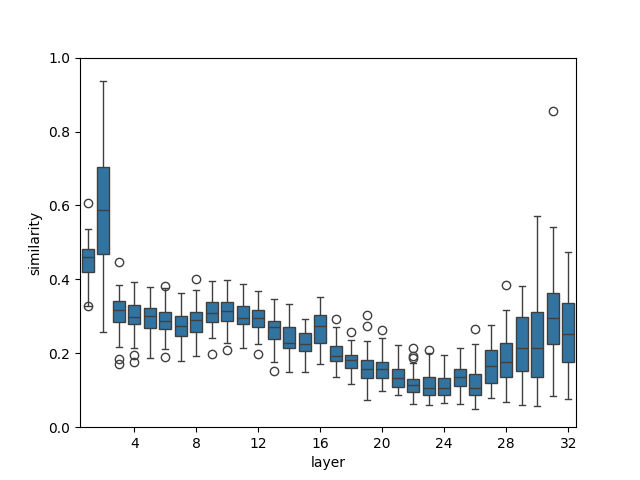}
\caption{c4-siqa}
  \end{subfigure}
  \hfill
\begin{subfigure}{0.15\textwidth}
    \includegraphics[width=\textwidth]{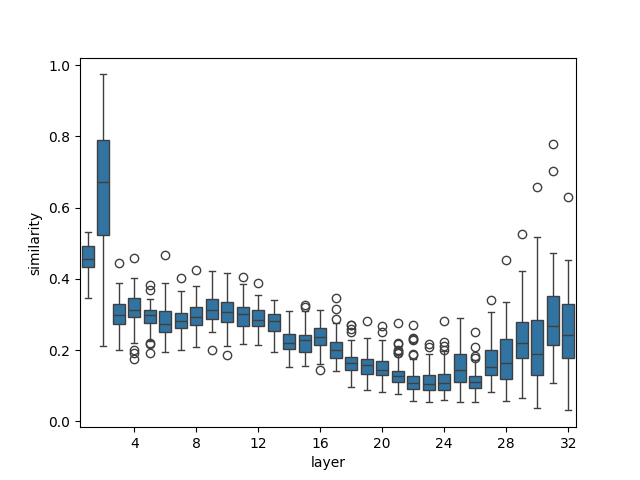}
\caption{wikitext2-mmlu}
  \end{subfigure}
  \hfill
\begin{subfigure}{0.15\textwidth}
    \includegraphics[width=\textwidth]{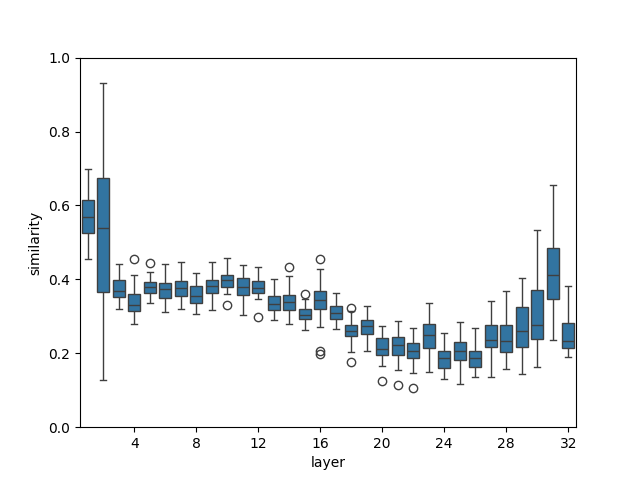}
\caption{siqa-gsm8k}
  \end{subfigure}
  \hfill
\begin{subfigure}{0.15\textwidth}
    \includegraphics[width=\textwidth]{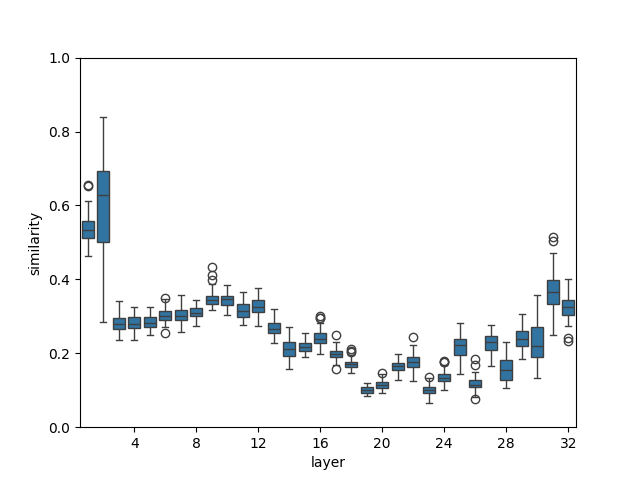}
\caption{siqa-humaneval}
  \end{subfigure}
  \hfill
\begin{subfigure}{0.15\textwidth}
    \includegraphics[width=\textwidth]{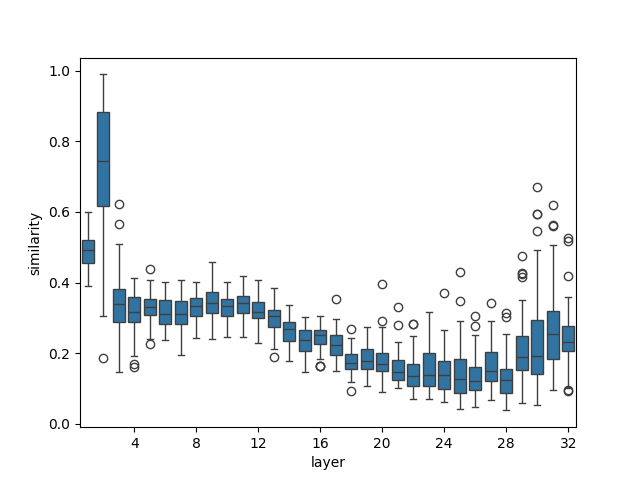}
\caption{wikitext2-gsm8k}
  \end{subfigure}
  \hfill
\begin{subfigure}{0.15\textwidth}
    \includegraphics[width=\textwidth]{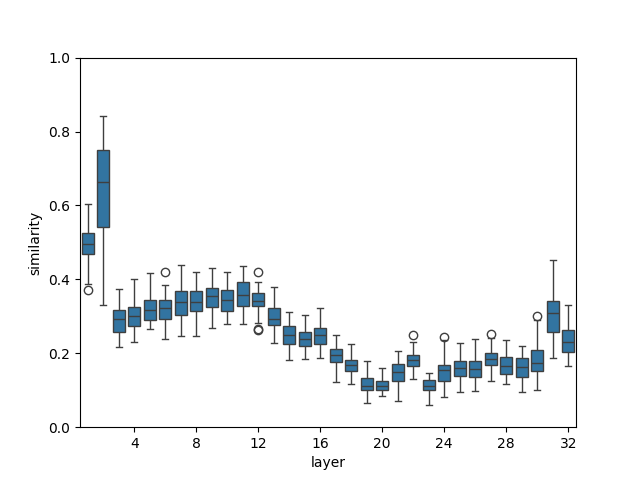}
\caption{boolq-humaneval}
  \end{subfigure}
  \hfill
\begin{subfigure}{0.15\textwidth}
    \includegraphics[width=\textwidth]{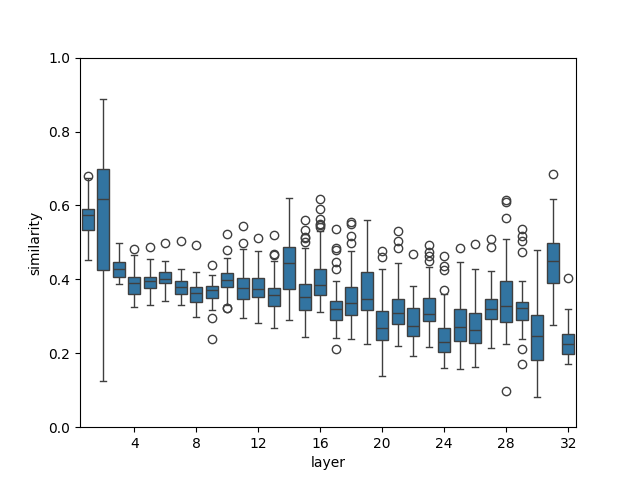}
\caption{piqa-siqa}
  \end{subfigure}
  \hfill
\begin{subfigure}{0.15\textwidth}
    \includegraphics[width=\textwidth]{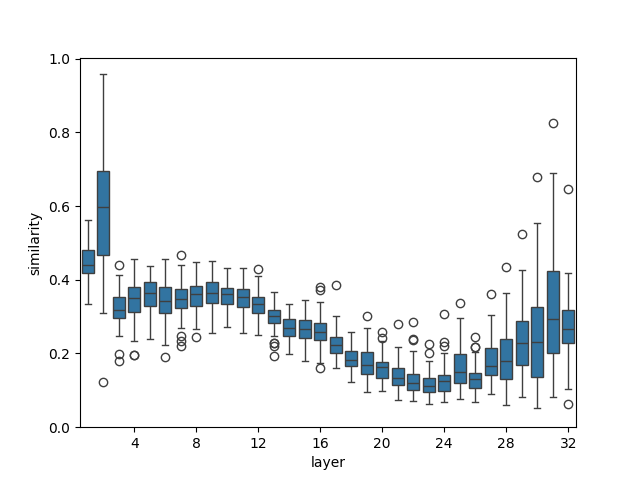}
\caption{wikitext2-hellaswag}
  \end{subfigure}
  \hfill
\begin{subfigure}{0.15\textwidth}
    \includegraphics[width=\textwidth]{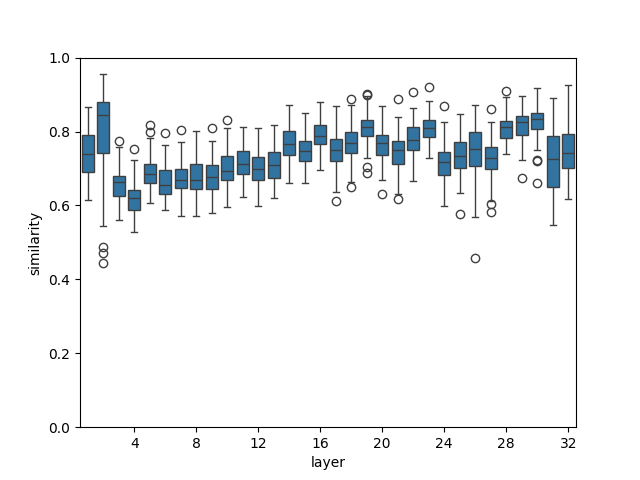}
\caption{siqa-siqa}
  \end{subfigure}
  \hfill
\begin{subfigure}{0.15\textwidth}
    \includegraphics[width=\textwidth]{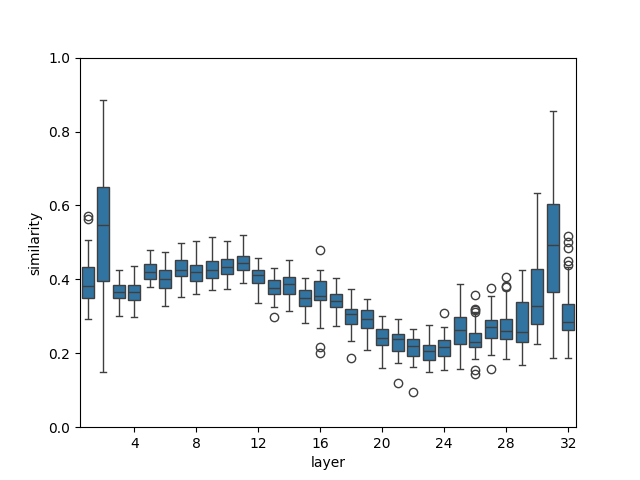}
\caption{gsm8k-hellaswag}
  \end{subfigure}
  \hfill
\begin{subfigure}{0.15\textwidth}
    \includegraphics[width=\textwidth]{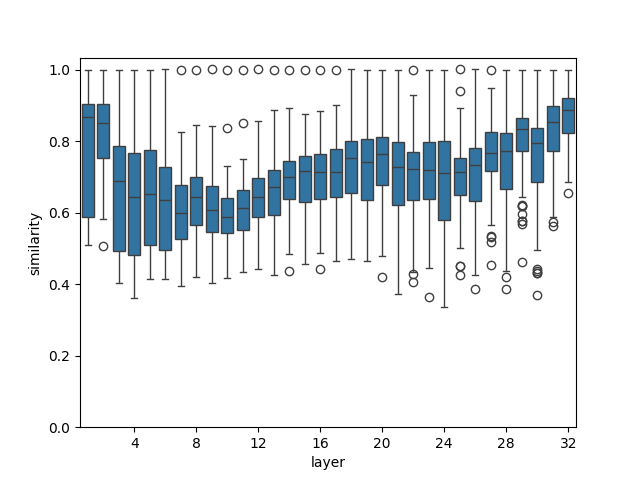}
\caption{humaneval-humaneval}
  \end{subfigure}
  \hfill
\begin{subfigure}{0.15\textwidth}
    \includegraphics[width=\textwidth]{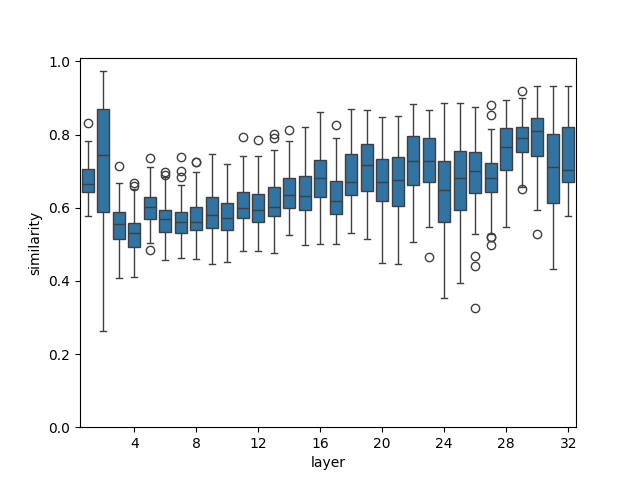}
\caption{mmlu-mmlu}
  \end{subfigure}
  \hfill
\begin{subfigure}{0.15\textwidth}
    \includegraphics[width=\textwidth]{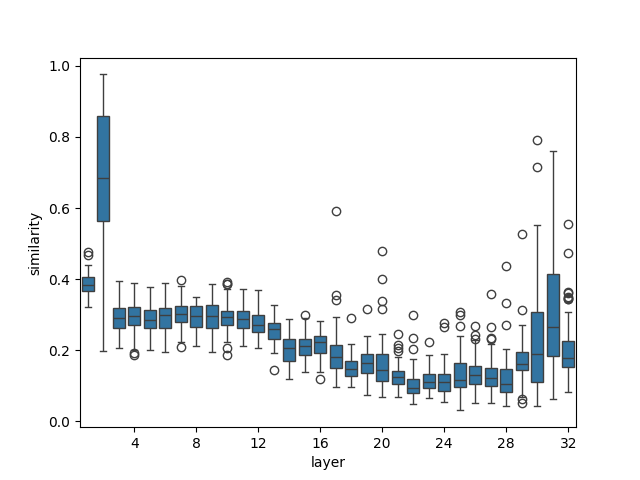}
\caption{wikitext2-piqa}
  \end{subfigure}
  \hfill
\begin{subfigure}{0.15\textwidth}
    \includegraphics[width=\textwidth]{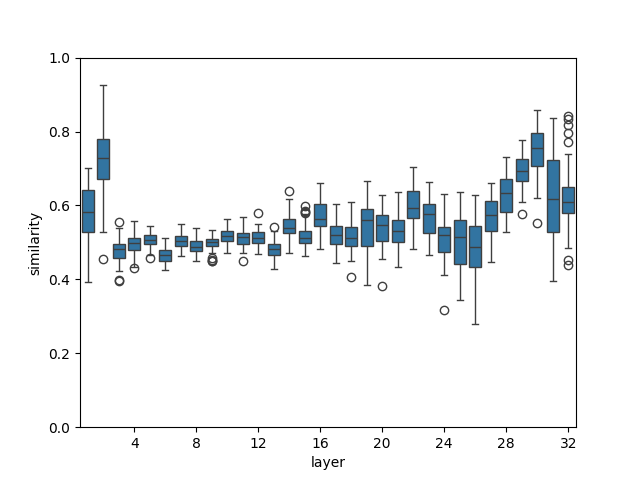}
\caption{siqa-hellaswag}
  \end{subfigure}
  \hfill
\begin{subfigure}{0.15\textwidth}
    \includegraphics[width=\textwidth]{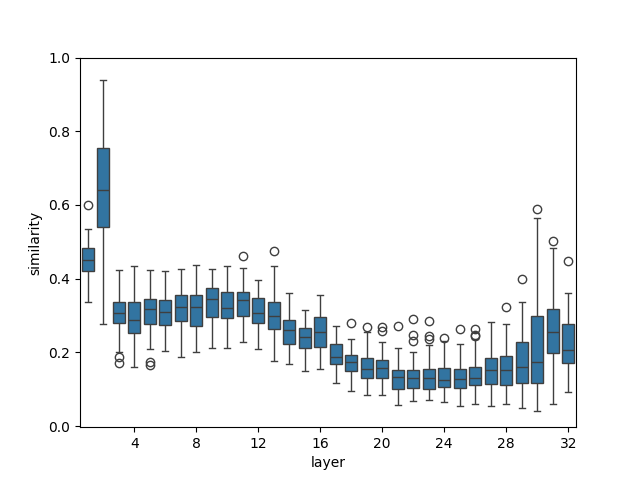}
\caption{c4-boolq}
  \end{subfigure}
  \hfill
\begin{subfigure}{0.15\textwidth}
    \includegraphics[width=\textwidth]{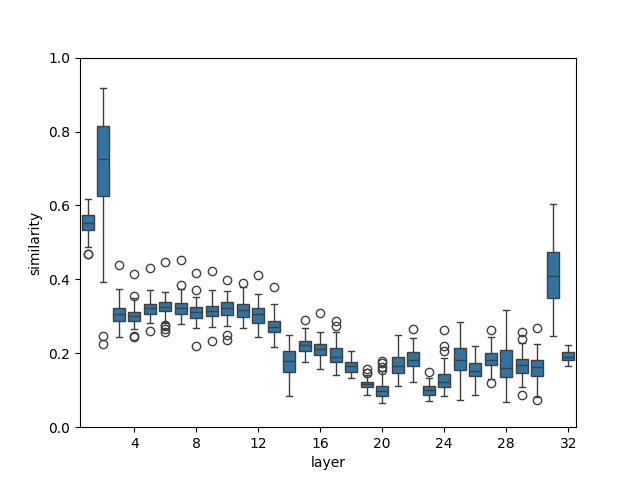}
\caption{piqa-humaneval}
  \end{subfigure}
  \hfill
\begin{subfigure}{0.15\textwidth}
    \includegraphics[width=\textwidth]{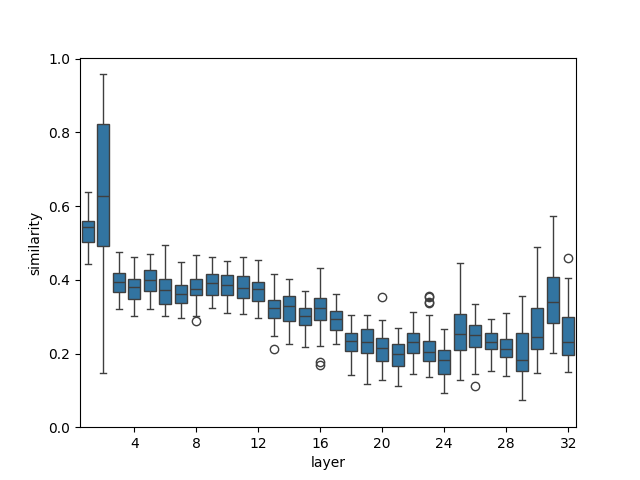}
\caption{boolq-gsm8k}
  \end{subfigure}
  \hfill
\begin{subfigure}{0.15\textwidth}
    \includegraphics[width=\textwidth]{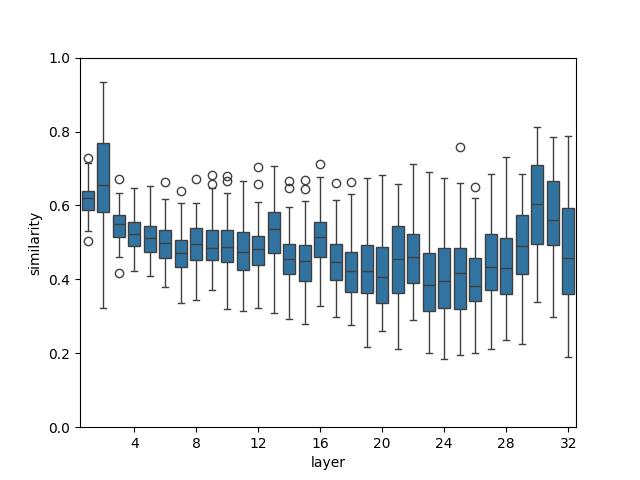}
\caption{boolq-boolq}
  \end{subfigure}
  \hfill
\begin{subfigure}{0.15\textwidth}
    \includegraphics[width=\textwidth]{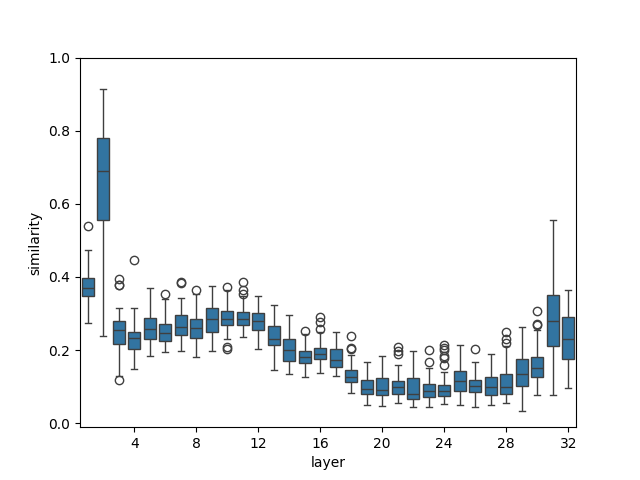}
\caption{c4-humaneval}
  \end{subfigure}
  \hfill
\begin{subfigure}{0.15\textwidth}
    \includegraphics[width=\textwidth]{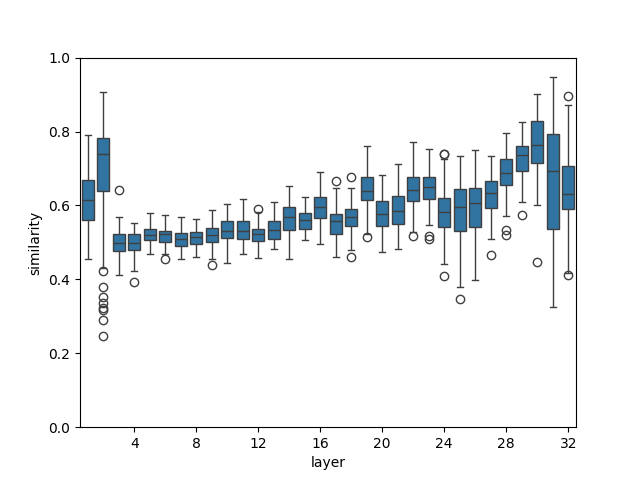}
\caption{hellaswag-mmlu}
  \end{subfigure}
  \hfill
\begin{subfigure}{0.15\textwidth}
    \includegraphics[width=\textwidth]{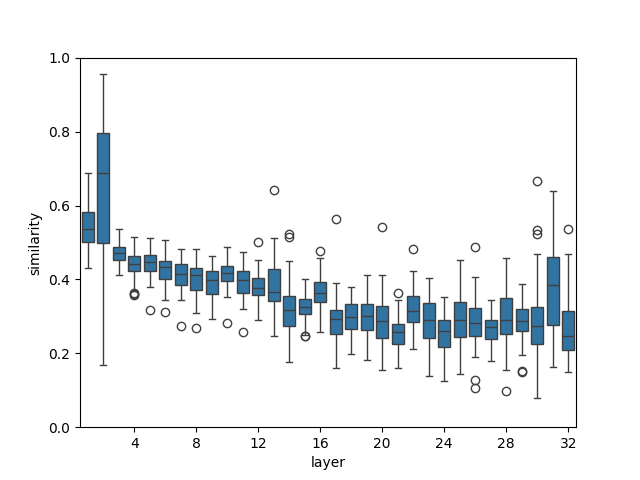}
\caption{piqa-boolq}
  \end{subfigure}
  \hfill
\begin{subfigure}{0.15\textwidth}
    \includegraphics[width=\textwidth]{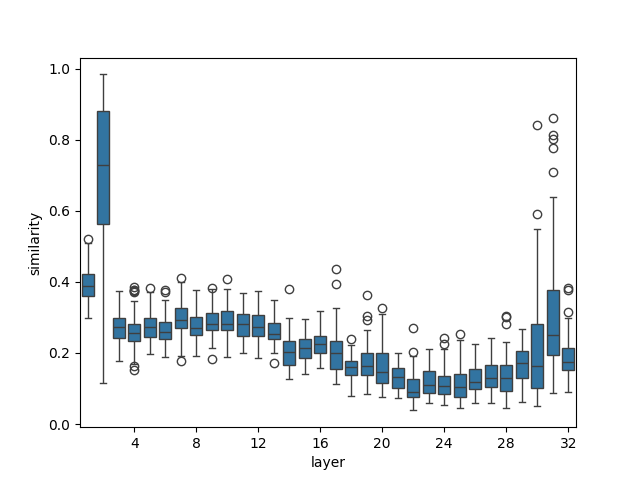}
\caption{c4-piqa}
  \end{subfigure}
  \hfill

  \caption{Statistics of LLMDcos values for different layers. The x-axis represents the layer number, and the y-axis represents the LLMDcos values of 64 samples. The two data sources are indicated below the image. All images are results from Llama2-7b. }
\end{figure}

%%%%%%%%%%%%%%%%%%%%%%%%%%%%%%%%%%%%%%%%%%%%%%%%%%%%%%%%%%%%

\newpage
\section*{NeurIPS Paper Checklist}

\begin{enumerate}

\item {\bf Claims}
    \item[] Question: Do the main claims made in the abstract and introduction accurately reflect the paper's contributions and scope?
    \item[] Answer: \answerYes{} % Replace by \answerYes{}, \answerNo{}, or \answerNA{}.
    \item[] Justification: contributions and scope are included in Section~\ref{sec:intro}.

\item {\bf Limitations}
    \item[] Question: Does the paper discuss the limitations of the work performed by the authors?
    \item[] Answer: \answerYes{} % Replace by \answerYes{}, \answerNo{}, or \answerNA{}.
    \item[] Justification: Limitations are discussed in Section~\ref{sec:lim}

\item {\bf Theory Assumptions and Proofs}
    \item[] Question: For each theoretical result, does the paper provide the full set of assumptions and a complete (and correct) proof?
    \item[] Answer: \answerNA{} % Replace by \answerYes{}, \answerNo{}, or \answerNA{}.
    \item[] Justification: There are no theoretical results. 

    \item {\bf Experimental Result Reproducibility}
    \item[] Question: Does the paper fully disclose all the information needed to reproduce the main experimental results of the paper to the extent that it affects the main claims and/or conclusions of the paper (regardless of whether the code and data are provided or not)?
    \item[] Answer: \answerYes{} % Replace by \answerYes{}, \answerNo{}, or \answerNA{}.
    \item[] Justification: We will provide code in supplementary material and release it after review.

\item {\bf Open access to data and code}
    \item[] Question: Does the paper provide open access to the data and code, with sufficient instructions to faithfully reproduce the main experimental results, as described in supplemental material?
    \item[] Answer: \answerYes{} % Replace by \answerYes{}, \answerNo{}, or \answerNA{}.
    \item[] Justification: All the data are from public datasets and code will be provided.

\item {\bf Experimental Setting/Details}
    \item[] Question: Does the paper specify all the training and test details (e.g., data splits, hyperparameters, how they were chosen, type of optimizer, etc.) necessary to understand the results?
    \item[] Answer: \answerYes{} % Replace by \answerYes{}, \answerNo{}, or \answerNA{}.
    \item[] Justification: in the appendix

\item {\bf Experiment Statistical Significance}
    \item[] Question: Does the paper report error bars suitably and correctly defined or other appropriate information about the statistical significance of the experiments?
    \item[] Answer: \answerYes{} % Replace by \answerYes{}, \answerNo{}, or \answerNA{}.
    \item[] Justification: shown in statistics figures.

\item {\bf Experiments Compute Resources}
    \item[] Question: For each experiment, does the paper provide sufficient information on the computer resources (type of compute workers, memory, time of execution) needed to reproduce the experiments?
    \item[] Answer: \answerYes{} % Replace by \answerYes{}, \answerNo{}, or \answerNA{}.
    \item[] Justification: in the appendix
    
\item {\bf Code Of Ethics}
    \item[] Question: Does the research conducted in the paper conform, in every respect, with the NeurIPS Code of Ethics \url{https://neurips.cc/public/EthicsGuidelines}?
    \item[] Answer: \answerYes{} % Replace by \answerYes{}, \answerNo{}, or \answerNA{}.
    \item[] Justification: \answerNA{}

\item {\bf Broader Impacts}
    \item[] Question: Does the paper discuss both potential positive societal impacts and negative societal impacts of the work performed?
    \item[] Answer: \answerNA{} % Replace by \answerYes{}, \answerNo{}, or \answerNA{}.
    \item[] Justification: This work only try to explain the model and verify the explaination.
    
\item {\bf Safeguards}
    \item[] Question: Does the paper describe safeguards that have been put in place for responsible release of data or models that have a high risk for misuse (e.g., pretrained language models, image generators, or scraped datasets)?
    \item[] Answer: \answerNA{} % Replace by \answerYes{}, \answerNo{}, or \answerNA{}.
    \item[] Justification: This work try to explain the model and do not release any new model or datasets.

\item {\bf Licenses for existing assets}
    \item[] Question: Are the creators or original owners of assets (e.g., code, data, models), used in the paper, properly credited and are the license and terms of use explicitly mentioned and properly respected?
    \item[] Answer: \answerYes{} % Replace by \answerYes{}, \answerNo{}, or \answerNA{}.
    \item[] Justification: All the codes, models and datasets are cited

\item {\bf New Assets}
    \item[] Question: Are new assets introduced in the paper well documented and is the documentation provided alongside the assets?
    \item[] Answer: \answerNA{} % Replace by \answerYes{}, \answerNo{}, or \answerNA{}.
    \item[] Justification: All the models/codes/datasets are public.

\item {\bf Crowdsourcing and Research with Human Subjects}
    \item[] Question: For crowdsourcing experiments and research with human subjects, does the paper include the full text of instructions given to participants and screenshots, if applicable, as well as details about compensation (if any)? 
    \item[] Answer: \answerNA{} % Replace by \answerYes{}, \answerNo{}, or \answerNA{}.
    \item[] Justification: \answerNA{}

\item {\bf Institutional Review Board (IRB) Approvals or Equivalent for Research with Human Subjects}
    \item[] Question: Does the paper describe potential risks incurred by study participants, whether such risks were disclosed to the subjects, and whether Institutional Review Board (IRB) approvals (or an equivalent approval/review based on the requirements of your country or institution) were obtained?
    \item[] Answer: \answerNA{} % Replace by \answerYes{}, \answerNo{}, or \answerNA{}.
    \item[] Justification: \answerNA{}

\end{enumerate}

\end{document}